\documentclass[lettersize,journal]{IEEEtran}

\usepackage{amsmath,amsfonts,amsthm}
\usepackage{algorithmic}
\usepackage{algorithm}
\usepackage{array}
\usepackage{textcomp}
\usepackage{stfloats}
\usepackage{url}
\usepackage{verbatim}
\usepackage{graphicx}
\usepackage{cite}
\usepackage{autobreak}
\usepackage{subfigure}
\usepackage{enumerate}
\usepackage{balance}
\usepackage[framemethod=tikz]{mdframed}

\definecolor{mygray}{RGB}{243,243,244}
\newmdenv[innertopmargin=0pt, backgroundcolor=mygray, linecolor=none, innerleftmargin=0pt, innerrightmargin=0pt, leftmargin=0pt]{mymath}

\theoremstyle{plain}
\newtheorem{theorem}{Theorem}[section]

\newtheorem{lemma}[theorem]{Lemma}

\theoremstyle{definition}
\newtheorem{definition}[theorem]{Definition}

\theoremstyle{remark}

\begin{document}

% =================================================================
% PART 1: Main Paper
% =================================================================

% --- CRITICAL FIX 1: Disable writing to the Table of Contents ---
\let\oldaddcontentsline\addcontentsline
\renewcommand{\addcontentsline}[3]{} 

\title{\textbf{\texttt{ASIL}}: Augmented Structural Information Learning for Deep Graph Clustering in Hyperbolic Space}
\author{
Li Sun\thanks{Li Sun and Yujie Wang are with Beijing University of Posts and Telecommunications, Beijing, China.}, 
Zhenhao Huang\thanks{Zhenhao Huang and Hongbo Lv  are with North China Electric Power University, Beijing, China.}, 
Yujie Wang, 
Hongbo Lv, 
Chunyang Liu\thanks{Chunyang Liu is with DiDi Chuxing, China.}, 
Hao Peng\thanks{Hao Peng is with Beihang University, Beijing, China.}, 
Philip S. Yu,~\IEEEmembership{Life Fellow,~IEEE}\thanks{Philip S. Yu is with University of Illinois at Chicago, IL, USA.}
}

\markboth{IEEE Transactions on Pattern Analysis and Machine Intelligence,~Vol.~X, No.~X, Feb~2025}%
{Shell \MakeLowercase{\textit{et al.}}: A Sample Article Using IEEEtran.cls for IEEE Journals}

\maketitle

\begin{abstract}
Graph clustering is a longstanding topic in machine learning. 
In recent years, deep learning methods have achieved encouraging  results, but they still require predefined cluster numbers $K$, and typically struggle with imbalanced graphs, especially in identifying minority clusters.
The limitations motivate us to study a challenging yet practical problem: deep graph clustering without $K$ considering the imbalance in reality.
We approach this problem from a fresh perspective of information theory (i.e., structural information).
In the literature, structural information has rarely been touched in deep clustering, and the classic definition falls short in its discrete formulation, neglecting node attributes and exhibiting prohibitive complexity.
In  this paper, we first establish a \emph{\underline{differentiable structural information}}, generalizing the discrete formalism to continuous realm,
so that we design a hyperbolic deep model (\texttt{LSEnet}) to learn the neural partitioning tree in the Lorentz model of hyperbolic space.
Theoretically, we demonstrate its capability in clustering without requiring $K$ and identifying minority clusters in imbalanced graphs.
Second, we refine hyperbolic representations of the partitioning tree, enhancing graph semantics, for better clustering.
Contrastive learning for  tree structures is non-trivial and costs quadratic complexity. 
Instead, we further advance our theory by discovering an interesting fact that  structural entropy indeed bounds the tree contrastive loss.
Finally, with an efficient reformulation, 
we approach graph clustering  through a novel \emph{\underline{augmented structural information learning}} (\texttt{ASIL}), 
which offers a simple yet effective objective of augmented structural entropy to seamlessly integrates hyperbolic partitioning tree construction and contrastive learning.
With a provable improvement in graph conductance, \texttt{ASIL} achieves effective debiased graph clustering in linear complexity with respect to the graph size.
Extensive experiments show the \texttt{ASIL} outperforms  $20$ strong baselines by an average of $+12.42\%$  in NMI on Citeseer dataset. 
\end{abstract}


\emph{\textbf{Codes}}-- \url{https://github.com/RiemannGraph/DSE_clustering}

%!TEX root = ./main.tex

\section{Introduction}

\IEEEPARstart{G}{raph} clustering shows fundamental importance for understanding graph data and has a wide range of applications, 
including biochemical analysis, recommender system and community detection \cite{DBLP:conf/www/JiaZ0W19,liu2023survey}.
In recent years, deep learning has been introduced to graph clustering, 
where graph neural networks \cite{kipf2017semisupervised,velickovic2018graph} learn informative embeddings with a clustering objective performed in the representation space \cite{devvrit2022s3gc,wang2023gc}. 
While achieving remarkable success, deep graph clustering still faces two major challenges:

\emph{Challenge 1: The requirement of predefined cluster numbers $K$.} So far, deep graph clustering typically relies on predefined cluster numbers K, and thus one is required to provide an accurate K prior to clustering, which is often impractical in real cases.
On the one hand, it is challenging to estimate the number of node clusters.
The methods, e.g., Bayesian information criterion \cite{DBLP:journals/sigkdd/Schubert23} and recent RGC \cite{liu2023reinforcement}, repeatedly train and test the deep model to search for the optimal $K$,
and the searching process typically results in high computation overhead. 
On the other hand, existing solutions without graph structures (e.g., Bayesian non-parametric methods \cite{gershman2011tutorial}, DBSCAN \cite{esterdensitybased}, and hierarchical clustering \cite{cohen-addad2019hierarchical}) cannot be directly applied  owing to the inter-correlation among nodes.
Thus, deep graph clustering without $K$ remains largely underexplored.

\emph{Challenge 2: Clustering the imbalanced graphs.}
The imbalance is prevalent in real-world graphs, where a large proportion of vertices belong to the majority clusters, while minority clusters contain significantly fewer vertices \cite{ImbalanceSurvey,icml24imbalance}. 
Take the popular Cora dataset \cite{sen2008collective} as an example:  $26.8\%$  of the papers belong to Neural Network cluster, whereas only $4.8\%$ fall under Rule Learning cluster.
Deep graph clustering models typically rely on the representation learning, and minority clusters tend to be underrepresented due to their limited sample sizes compared to the majority clusters \cite{shi2020multi}.
This imbalance increases the risk of generating biased clustering boundaries that favor majority  clusters, as evidenced by empirical observations in our Case Study.
Also, the challenge becomes even greater when the number of clusters is unknown.

\begin{figure*}[t]
\centering
 \vspace{-0.2in}
\includegraphics[width=0.95\linewidth]{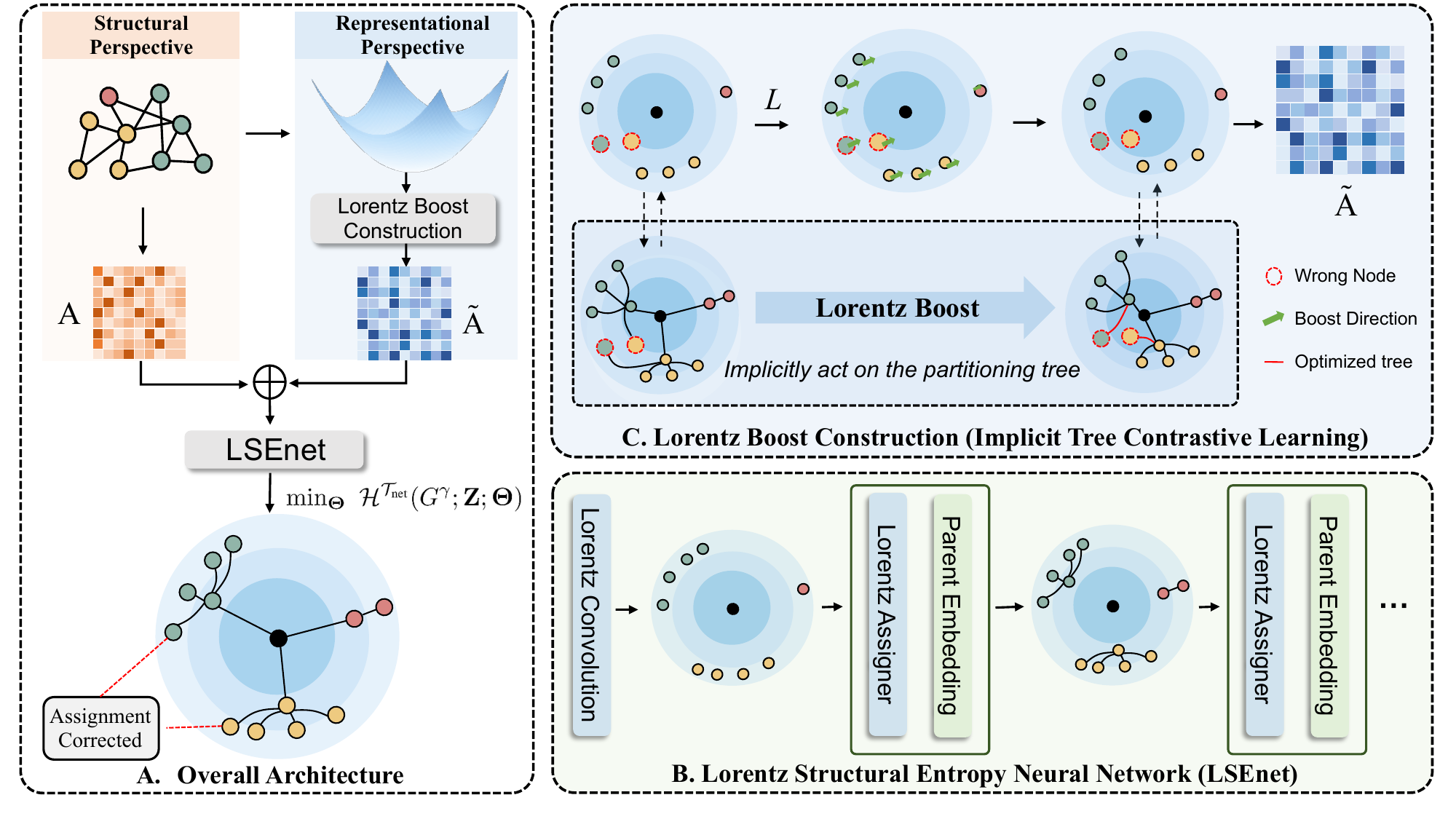}
 \vspace{-0.15in}
\caption{\textbf{Illustration of \texttt{ASIL}}. With a novel objective of augmented structural entropy, \texttt{ASIL} seamlessly integrates  partitioning tree construction and contrastive learning in the Lorentz model of hyperbolic space. It achieves debiased node  clustering in linear complexity without requiring the number of clusters.}
 \vspace{-0.2in}
\label{fig:overall}
\end{figure*}

In this paper, we revisit deep graph clustering with an unknown $K$, considering the imbalance in real-world graphs, and address this problem from a fresh perspective based on information theory.
Stemming from Shannon entropy, \textbf{structural information} \cite{li2016structural} has been proposed to measure the uncertainty in graphs.
According to structural information minimization, the optimal partitioning tree can be constructed to describe a graph's self-organization without requiring $K$.
While this sheds light on the targeted problem, it presents several significant challenges to deep clustering:
\emph{(1) Connection to Clustering.} Structural information has recently garnered increasing research attention \cite{liu2019rem,wu2023sega,zou2023segslb}, but its theoretical connection to clustering remains unclear. As a result,  structural information has not yet been introduced to graph clustering. 
\emph{(2) Discrete Formulation. } 
The discrete nature of structural information prevents gradient backpropagation, posing a major challenge in training deep graph models.
\emph{(3) Attribute Ignorance.} Structural information operates solely on the graph structure overlooking node attributes, which  are often equally important to graph clustering. 
\emph{(4) Computational Complexity.} Constructing a partitioning tree of height $3$ yields the complexity of $O(\lvert N\rvert^3 \log^2 \lvert N \rvert)$, where $N$ is the number of nodes. Note that, the complexity scales exponentially with tree height \cite{li2016structural}, making it infeasible for real-world graphs.

To tackle these challenges, 
we establish a new \emph{differentiable structural information} (\texttt{DSI}), generalizing the discrete formalism to the continuous realm. 
In \texttt{DSI},  we reformulate the structural information  with level-wise assignment matrices.
We prove that \texttt{DSI} upper bounds the graph conductance, a well-defined clustering objective, and thus \texttt{DSI} emerges as a new differentiable graph clustering objective without requiring the number of clusters  $K$.
Thus we create a \emph{deep partitioning tree} $\mathcal T_{\operatorname{net}}$ parametrized by a neural network.
Different from the partitioning tree of \cite{li2016structural} constructed by heuristic algorithm, 
the proposed deep tree can be learned by gradient backpropagation according to \texttt{DSI} minimization.
Theoretically, we show that the structural entropy of the optimal $\mathcal T_{\operatorname{net}}$ well approximates that of \cite{li2016structural} under slight constraint.
it is able to identify the minority clusters from the majority ones, which is suitable for clustering real graphs that often exhibit imbalanced cluster distributions.
Furthermore, we uncover another advantage of the deep partitioning tree—its capability to distinguish minority clusters from majority ones, making it suitable for clustering real-world graphs that often exhibit imbalanced distributions.

Grounded on the established theory, we approach graph clustering through a novel \emph{Augmented Structural Information Learning} (\textbf{\texttt{ASIL}}), as sketched in Fig. \ref{fig:overall}A, which learns and refines the deep partitioning tree in the \textbf{hyperbolic space} for node clustering without $K$.
Specifically, in the conference version\footnote{The preliminary results have been published in the Proceedings of the 41st International Conference on Machine Learning (ICML’24), Oral Paper \cite{icml24sun}.}, 
we design Lorentz Structural Entropy net  (\texttt{LSEnet}) to parameterize the deep partitioning tree, as shown in Fig. \ref{fig:overall}C. 
Given the alignment between hyperbolic space and hierarchical/tree-like structures, we propose to operate in the Lorentz model of hyperbolic space. 
In particular, we first embed the leaf nodes of the partitioning tree using Lorentz convolution over the graph.
Then, we recursively learn the parent nodes from the bottom (leaf nodes) to the top (root node) using the proposed Lorentz assigner.
Note that the structural information itself primarily operates on graph structure, presenting the complexity of $\mathcal O(|\mathcal E||\mathcal V|)$.\footnote{$\mathcal V$ and $\mathcal E$ denote the node set and edge set of the graph, respectively.}

Going beyond, in this full version, we aim to enhance both effectiveness and efficiency of structural entropy clustering.
For effectiveness, we refine the partitioning tree representation in the Lorentz model of hyperbolic space via contrastive learning.
Different from graph contrastive learning, we derive a neural Lorentz boost (Theorem VI.1, VI.2) as a special projection head of consistency insurance—pushing leaf nodes together with the parent hierarchies as a unity—and formulate a tree contrastive loss considering partitioning tree branches for clustering assignment.
However, its quadratic complexity prevents the clustering model scaling up.
For efficiency, we further advance our established theory by \textbf{an interesting new finding} (Theorem VII.1) that  structural entropy indeed bounds the contrastive loss.
Accordingly, we connect the entropy (of structural perspective) and contrastive learning (of representational perspective) through a novel virtual graph constructed from the representations of the proposed Lorentz boost, maintaining consistency insurance, 
so that optimizing the virtual graph implicitly performs the tree contrastive learning, as shown in Fig. \ref{fig:overall}B.
Meanwhile, we equivalently reformulate structural entropy by exploring level-wise assignments, achieving linear time complexity.
Finally, we propose an expressive objective of \textbf{augmented structural entropy}, fusing the virtual graph described above, which seamlessly integrates partitioning tree construction and contrastive learning.
With a provable improvement in graph conductance (Theorem VII.2), \texttt{ASIL} achieves effective debiased graph clustering in linear complexity with respect to the graph size.

\noindent \textbf{Contribution Highlights.} 
The key contributions are five-fold:
\begin{itemize}
\item \emph{Problem.} We revisit deep graph clustering without a predefined cluster number, considering the imbalance of real graphs and, to the best of our knowledge, bridge graph clustering and structural entropy for the first time.
\item \emph{Theory.} We establish the differentiable structural information in continuous realm, so that a deep partitioning tree with linear computational complexity is created by gradient backpropagation for graph clustering without $K$. 
We further connect the theory of structural information to constrastive learning by proving a new bound.
\item  \emph{Neural Architecture.} We present a hyperbolic clustering architecture that first fuses a parameterized virtual graph and then leverages  \texttt{LSEnet} to learn the partitioning tree in the Lorentz model of hyperbolic space.
\item  \emph{Learning Paradigm.} We propose a novel \emph{Augmented Structural Information Learning} that  integrates partitioning tree construction and contrastive learning for more effectively and efficiently debiased graph clustering without requiring $K$.
\item \emph{Experiment.} Extensive real world results show the superiority of \texttt{ASIL} for graph clustering, and we examine that  biased boundary  among imbalanced clusters is alleviated with case study and visualization.
\end{itemize}

\section{Preliminaries and Notations}

Different from the typical graph clustering, 
we study a challenging yet practical problem of graph clustering without the number of clusters, considering the imbalance of real graphs. 
This section formally reviews the basic concepts of structural information, hyperbolic space and Lorentz group, and introduce the main notations.

\subsubsection{\textbf{Graph \& Graph Clustering}}

An attribute graph $G=(\mathcal{V}, \mathcal{E}, \mathbf X)$ is defined on the node set $\mathcal{V}$,  $|\mathcal{V}|=N$.
The edge set  $\mathcal{E}  \subset \mathcal{V} \times \mathcal{V}$ is associated with a weight function $w:\mathcal{E} \rightarrow \mathbb{R}^{+}$.
Alternatively, $\mathbf A \in \mathbb R^{N \times N}$ is  the adjacency matrix collecting the edge weights.
Each node is attached to a $d$-dimension attribute vector, summarized  in $\mathbf{X} \in \mathbb R^{N \times D}$.
Graph clustering aims to group nodes into several disjoint clusters, so that nodes share higher similarity with the nodes in the same clusters than that of different clusters.
Deep learning has become the dominant solution to graph clustering in recent years.
In deep graph clustering, a self-supervised loss is carefully designed to generate informative embeddings, and clustering is conducted in the representation space.
On the one hand, the cluster number $K$ is a typical prerequisite of deep models, but offering $K$ before clustering is impractical in real cases.
On the other hand,  real graphs are often imbalanced, exhibiting skewed class distributions. 
The minority cluster of fewer nodes tends to be underrepresented in the self-supervised learning owing to limited samples. 
However, clustering on imbalanced graphs is still underexplored in the literature.
To bridge the gaps, we are interested in deep graph clustering without $K$, considering the imbalance of real graphs.

\subsubsection{\textbf{Structural Information \& Structural Entropy}}
Given a graph $G=(\mathcal{V},\mathcal{E})$ with weight function $w$, the weighted degree of node $v$ is written as $d_v=\sum\nolimits_{u\in \mathcal N(v)}w(v,u)$, the summation of the edge weights surrounding it. 
$\mathcal N(v)$ denotes the neighborhood of $v$.
For an arbitrary node set $\mathcal{U}  \subset \mathcal{V}$, its volume $\operatorname{Vol}(\mathcal{U})$ is defined by summing of weighted degrees of the nodes in $\mathcal{U}$, and $\operatorname{Vol}(G)$ denotes the volume of the graph.
The one-dimensional structural entropy is defined as follows,
\vspace{-0.03in}
    \begin{align} \label{eq.one.si}
            \mathcal{H}^{1}(G)=-\sum_{v\in \mathcal{V}}\frac{d_{v}}{\operatorname{Vol}(G)}\log_2\frac{d_{v}}{\operatorname{Vol}(G)},
            \vspace{-0.03in}
    \end{align}
yielding as a constant for certain given graph.
The $H$-dimensional structural information is accompanied with a partitioning tree $\mathcal{T}$ of height $H$.
Concretely, in $\mathcal{T}$ with root node $\lambda$, 
each tree node $\alpha$ is associated with a subset of $\mathcal V$, referred to as module $T_\alpha \subset \mathcal{V}$.\footnote{The vertex of partitioning tree is termed as ``tree node'' to distinguish from the node in graphs.}
and its immediate predecessor is written as $\alpha^-$. 
The leaf node of the tree corresponds to the node of graph $G$.
Accordingly, the module of the leaf node is a singleton, while $T_\lambda$ is the node set $\mathcal V$.
The structural information assigned to each non-root node $\alpha$ is defined as 
\vspace{-0.03in}
    \begin{align}  \label{eq.node.si}
            \mathcal{H}^{\mathcal{T}}(G;\alpha)=-\frac{g_\alpha}{\operatorname{Vol}(G)}\log_2\frac{V_\alpha}{V_{\alpha^-}},
    \vspace{-0.03in}
    \end{align}
where the scalar $g_\alpha$ is the total weights  of graph edges with exactly one endpoint in the $T_\alpha$, and $V_\alpha$ is the volume of the module $T_\alpha$.
The $H$-dimensional structural information of $G$ with respect to the partitioning tree $\mathcal{T}$ is written as follows,
   \vspace{-0.03in}
    \begin{align}\label{eq.tree.si}
            \mathcal{H}^{\mathcal{T}}(G) = \sum\nolimits_{\alpha \in \mathcal{T}, \alpha \neq \lambda }\mathcal{H}^{\mathcal{T}}(G;\alpha).
            \vspace{-0.03in}
    \end{align}
Traversing all possible partitioning trees of $G$ with height $H$,  the \emph{$H$-dimensional structural entropy} of $G$ is defined as
\vspace{-0.03in}
    \begin{align}
        \mathcal{H}^{H}(G) = \min_\mathcal{T} \mathcal{H}^{\mathcal{T}}(G), \quad
    \label{eq.opt_tree}
        \mathcal{T}^*=\arg_{\mathcal{T}} \min \mathcal{H}^{\mathcal{T}}(G),
        \vspace{-0.03in}
    \end{align}
where  $\mathcal{T}^*$ is the optimal partitioning tree of $G$ that minimizes the uncertainty of graph, 
and encodes the self-organization of node clusters in the tree structure.

\subsubsection{\textbf{Hyperbolic Space \& Lorentz Group}}

Riemannian geometry offers an elegant framework to study complex structures.
A Riemannian manifold is described as a smooth manifold $\mathbb M$ endowed with a Riemannian metric. 
Each point $\boldsymbol x$ in the manifold is associated with the tangent space $\mathbb T_{\boldsymbol x}\mathbb M$ where the Riemannian metric is defined.
Mapping between the tangent space and manifold is done via the exponential and logarithmic maps.
The notion of curvature $\kappa$ describes the extent of how a manifold derivatives from being “flat”.
Accordingly, 
there exist three types of isotropic manifold: hyperbolic space of negative curvature, hyperspherical space of positive curvature and the special case of zero-curvature Euclidean space with “flat” geometry.
In Riemannian geometry, a structure is related to certain manifold and tree structure is aligned with hyperbolic space.
Here, we visit the \emph{Lorentz model of hyperbolic space} to study the partitioning tree of structural entropy.
Without loss of generality, a $d-$dimensional Lorentz model $\mathbb L^{\kappa, d}$ with curvature $\kappa$ is defined on the smooth manifold embedded in $\mathbb R^{d+1}$ space,
\vspace{-0.07in}
\begin{align}
    \mathbb L^{\kappa, d}=\{\boldsymbol x \in \mathbb R^{d+1}: \langle \boldsymbol x, \boldsymbol x \rangle_{\mathbb L}= \frac{1}{\kappa}, x_0>0\},
    \vspace{-0.07in}
\end{align} 
 where the Minkowski inner product  is given as follows,
 \vspace{-0.07in}
 \begin{align}
\langle \boldsymbol x, \boldsymbol y \rangle_{\mathbb L}=\boldsymbol x\mathbf R\boldsymbol y, \quad \mathbf R=\operatorname{diag}(-1, 1, \cdots, 1),
\label{Minkowski-InnerProduct}
\vspace{-0.07in}
\end{align} 
$\mathbf R\in \mathbb R^{(d+1)\times(d+1)}$ describes the Minkowski metric in $\mathbb R^{d+1}$ space.
$x_0$ is the first element of point coordinates.
For arbitrary two points in the Lorentz model, $\forall \boldsymbol x, \boldsymbol y$, 
the induced distance is given as $d_{\mathbb L}(\boldsymbol x, \boldsymbol y)=\sqrt{-\kappa}\operatorname{arccosh}(-\langle \boldsymbol x, \boldsymbol y \rangle_{\mathbb L})$.
$\mathbb T_{\boldsymbol x}\mathbb L^{\kappa, d}$ denotes the tangent space at $\boldsymbol x$, and $\lVert \boldsymbol{u} \rVert_\mathbb{L} = \sqrt{\langle \boldsymbol u, \boldsymbol u \rangle_\mathbb{L}}$ is known as Lorentz norm for $\boldsymbol u \in \mathbb T_{\boldsymbol x}\mathbb L^{\kappa, d}$.

In group theory, the Lorentz group studies the Lorentz model of hyperbolic space. 
To be specific, Lorentz group is a (generalized) orthogonal transformation group that collects the Lorentz transformation, consisting of Lorentz boosts and Lorentz rotations, which preserves the isometry regarding the Minkowski metric \cite{chen2022fullyb,petersen2016riemannian}.
Both of them are given as special orthogonal matrices, posing a significant challenge to parameterization.
Further facts on hyperbolic space and Lorentz group are provided in Appendix V.

\subsubsection{\textbf{Notations}}
Throughout this paper, $\mathbb L$ and $\mathcal T$ denote the Lorentz model of hyperbolic space and the partitioning tree, respectively. The vector and matrix are written in the lowercase boldfaced $\boldsymbol x$ and uppercase $\mathbf X$, respectively. The notation table is provided in Appendix I.

\emph{In this paper, we establish the theory of differential structural information, demonstrating its capacity in graph clustering and connecting it to contrastive learning. Thus, we approach graph clustering through Augmented Structural Entropy Learning (\texttt{ASIL}), whose key innovation lies in the integration of partitioning tree construction and contrastive learning. Without requiring the number of clusters,  \texttt{ASIL} effectively achieves  debiased  node clustering in linear complexity with respect to the graph size.
}

     \vspace{-0.1in}
\section{Differentiable Structural Information}

In this section, we establish the theory of  differentiable structural information, which generalizes the classic discrete formalism to the continuous realm.
The main theoretical results are dual: the proposed formulation emerges as a new objective for deep graph clustering without $K$; 
the corresponding partitioning tree can be  created via gradient-based deep learning, different from the previous heuristic algorithm \cite{li2016structural}.

     \vspace{-0.1in}
\subsection{New Formulation: A Differentiable Equivalence}

We introduce a differentiable version of structural entropy where the hard indicator for calculating the volume is relaxed as soft level-wise assignments. It is shown to be equivalent to the classic formulation, while enabling the optimization via gradient backpropagation.
     \vspace{-0.02in}
\begin{definition}[\textbf{Level-wise Assignment}]
\label{def.C}
    For a partitioning tree $\mathcal{T}$ with height $H$, assuming that the number of tree nodes at the $h$-th level  is $N_h$, we define a \emph{level-wise parent assignment matrix $\mathbf{C}^h \in \{0, 1\}^{N_h \times N_{h-1}}$ from $h$-th to $(h-1)$-th level}, where $\mathbf{C}^h_{ij}=1$ means the $i$-th node of $\mathcal{T}$ at $h$-th level is the child node of $j$-th node at $(h-1)$-th level.
\end{definition}
     \vspace{-0.02in}
\begin{definition}[\textbf{H-dimensional Structural Information}]
\label{DSE}
    For a graph $G$ and its partitioning tree $\mathcal{T}$ in Definition \ref{def.C}, we rewrite the formula of $H$-dimensional structural information of $G$ with respect to $\mathcal{T}$ \textbf{at height $h$} as 
     \vspace{-0.05in}
    \begin{equation}
        \mathcal{H}^\mathcal{T}(G;h)=-\frac{1}{V} 
            \sum\limits_{k=1}^{N_h}(V^h_k - \sum\limits_{(i,j)\in \mathcal{E}}S^h_{ik}S^h_{jk}w_{ij})\log_2\frac{V^h_k}{V^{h-1}_{k^-}}
    \label{kse}
     \vspace{-0.03in}
    \end{equation}
    where $V=\operatorname{Vol}(G)$ is the volume of $G$. For the $k$-th node in height $h$, $V^h_k$ and $V^{h-1}_{k^-}$ are the volume of graph node sets $T_k$ and $T_{k^-}$, respectively. Thus we have
     \vspace{-0.03in}
    \begin{align}
    \label{eq.s}
        \mathbf{S}^h &= \prod\nolimits_{k=H+1}^{h+1}\mathbf{C}^k,  \quad \mathbf{C}^{H+1}=\mathbf{I}_N,\\
        V^h_k &= \sum\nolimits_{i=1}^N S^h_{ik}d_i, \quad V^{h-1}_{k^-} = \sum\nolimits_{k'=1}^{N_{h-1}}\mathbf{C}_{kk'}^h V^{h-1}_{k'} .
         \label{eq.cv}
          \vspace{-0.03in}
    \end{align}
    Then, the $H$-dimensional structural information of $G$ is
       $ \mathcal{H}^\mathcal{T}(G) = \sum\nolimits_{h=1}^H H^\mathcal{T}(G;h)$.
\end{definition}
     \vspace{-0.03in}
\begin{mymath}
\begin{theorem}[\textbf{Equivalence}]
   \label{theorem.equivalence}
    The formula $\mathcal{H}^\mathcal{T}(G)$ in Definition \ref{DSE} is equivalent to Eq. (\ref{eq.tree.si}).
\end{theorem}
\begin{proof}
Please refer to Appendix II, Theorem III.3.
\end{proof}
\end{mymath}

\subsection{Connecting Structural Information and Graph Clustering}
This part shows some general properties of the proposed formulation and establishes the connection between structural entropy and graph clustering. 
We first give an arithmetic property regarding Definition \ref{DSE} to support the following claim on graph clustering.
The proofs of the lemma/theorems are detailed in Appendix II.
\begin{lemma}[\textbf{Additivity}]
\label{lemma.add}
    The $1$-dimensional structural entropy of $G$ can be decomposed as follows
         \vspace{-0.03in}
    \begin{align}
        \mathcal{H}^1(G) = \sum_{h=1}^H\sum_{j=1}^{N_{h-1}}\frac{V^{h-1}_j}{V}E([\frac{C^h_{kj}V^h_k}{V^{h-1}_j}]_{k=1,...,N_h}),
             \vspace{-0.03in}
    \end{align}
    where 
      $  E(p_1, ..., p_n) = -\sum_{i=1}^n p_i\log_2 p_i$
 is the entropy. 
\end{lemma}
\begin{mymath}
\begin{theorem}[\textbf{Connection to Graph Clustering}]
\label{theorem.conductance}
        Given a graph $G=(\mathcal{V}, \mathcal{E})$ with $w$, the normalized $H$-structural entropy of graph $G$ is defined as
    $\tau(G;H) = {\mathcal{H}^H(G)}/{\mathcal{H}^1(G)}$, and 
$\Phi(G)$ is the graph structural conductance.
    With the additivity \emph{(Lemma \ref{lemma.add})}, the following inequality holds,
         \vspace{-0.03in}
    \begin{align}
        \tau(G; H) \geq \Phi(G).
                 \vspace{-0.03in}
    \end{align}
\end{theorem}
\end{mymath}
\begin{proof}
    We show the key equations here, and further details are given in Appendix II Theorem III.5. Without loss of generality, we assume $\min\{ V^h_k, V-V^h_k \}=V^h_k$. From Definition \ref{DSE},
             \vspace{-0.03in}
    \begin{align}
        \mathcal{H}^\mathcal{T}(G) &= -\frac{1}{V}\sum_{h=1}^H\sum_{k=1}^{N_h} \phi_{h,k}V^h_k\log_2\frac{V^h_k}{V^{h-1}_{k^-}} \nonumber \\
        &\geq -\frac{\Phi(G)}{V}\sum_{h=1}^H \sum_{k=1}^{N_h}V^h_k\log_2\frac{V^h_k}{V^{h-1}_{k^-}},
    \end{align}
    Let the $k$-th tree node in height $h$ denoted as $\alpha$, 
    then $\phi_{h,k}$ is the  conductance of graph node subset $T_{\alpha}$, 
    and is defined as
    $\frac{\sum_{i\in T_{\alpha}, j\in \bar{T}_{\alpha}}w_{ij}}{\min\{ \operatorname{Vol}(T_{\alpha}), \operatorname{Vol}(\bar{T}_{\alpha}) \}}$.
    where $\bar{T}_{\alpha}$ is the complement set of $T_{\alpha}$. 
  The graph  conductance is given as $\Phi(G) = \min_{h,k}\{\phi_{h,k}\}$. 
    Next, we use Eq. (\ref{eq.cv}) to obtain the following inequality
    \begin{align}
        \mathcal{H}^\mathcal{T}(G) & \geq -\frac{\Phi(G)}{V}\sum_{h=1}^H \sum_{k=1}^{N_h}\sum_{j=1}^{N_{h-1}}C^h_{kj}V^h_k\log_2\frac{V^h_k}{V^{h-1}_{k^-}}
         \nonumber \\
        % &=-\frac{\Phi(G)}{V}\sum_{h=1}^H \sum_{j=1}^{N_{h-1}} V^{h-1}_j \sum_{k=1}^{N_h} \frac{C^h_{kj}V^h_k}{V^{h-1}_j}\log_2\frac{C^h_{kj}V^h_k}{V^{h-1}_j} \nonumber \\
        % &=\Phi(G)\sum_{h=1}^H\sum_{j=1}^{N_{h-1}}\frac{V^{h-1}_j}{V}E([\frac{C^h_{kj}V^h_k}{V^{h-1}_j}]_{k=1,...,N_h}) \nonumber \\
        % &=\Phi(G)\mathcal{H}^1(G).
    \end{align}
       \begin{align}
        \mathcal{H}^\mathcal{T}(G) 
        % \geq -\frac{\Phi(G)}{V}\sum_{h=1}^H \sum_{k=1}^{N_h}\sum_{j=1}^{N_{h-1}}C^h_{kj}V^h_k\log_2\frac{V^h_k}{V^{h-1}_{k^-}} \nonumber \\
      &\geq-\frac{\Phi(G)}{V}\sum_{h=1}^H \sum_{j=1}^{N_{h-1}} V^{h-1}_j \sum_{k=1}^{N_h} \frac{C^h_{kj}V^h_k}{V^{h-1}_j}\log_2\frac{C^h_{kj}V^h_k}{V^{h-1}_j} \nonumber \\
        &= \Phi(G)\sum_{h=1}^H\sum_{j=1}^{N_{h-1}}\frac{V^{h-1}_j}{V}E([\frac{C^h_{kj}V^h_k}{V^{h-1}_j}]_{k=1,...,N_h}) \nonumber \\
        &=\Phi(G)\mathcal{H}^1(G).
         \vspace{-0.03in}
    \end{align}
Since $\tau(G;\mathcal{T}) = \frac{\mathcal{H}^\mathcal{T}(G)}{\mathcal{H}^1(G)} \geq \Phi(G)$ holds for every $H$-height partitioning tree $\mathcal{T}$ of $G$, $\tau(G)\geq \Phi(G)$ holds.
\end{proof}
The conductance is a well-defined objective for graph clustering \cite{DBLP:conf/kdd/YinBLG17}.
$\forall G$, the one-dimensional structural entropy
  $\mathcal{H}^1(G)=-\frac{1}{\operatorname{Vol}(G)}\sum_{i=1}^N d_i\log_2\frac{d_i}{\operatorname{Vol}(G)} $ 
is yielded as a constant.
Minimizing Eq. (\ref{kse}) is thus equivalent to minimizing conductance, grouping nodes into clusters. 
Also, Eq. (\ref{kse}) needs no knowledge of cluster number in the graph $G$.
\emph{In short, our formulation is a differentiable objective function for  graph clustering without predefined cluster numbers.}

\textbf{\emph{Remark.}} Different from classical graph entropies measuring the uncertainty on graph, structural entropy also models the hierarchical self-organization of graphs through a partitioning tree. This enables it to serve as a clustering objective without requiring the number of clusters, a property absent in traditional graph entropy measures.

 \vspace{-0.03in}
\subsection{Deep Partitioning Tree}

Here, we elaborate on how to construct a deep partitioning tree through gradient backpropagation.
%se a deep model to conduct graph clustering with our new objective.
Recall Eq. (\ref{kse}), our objective is differentiable over the parent assignment.
If the assignment is relaxed to be the likelihood given by a neural net,
our formulation supports learning a deep model that constructs the partitioning tree.
For a graph $G$, we denote the deep partitioning tree as $\mathcal T_{\operatorname{net}}$ where the mapping from learnable embeddings $\mathbf{Z}$ to the assignment is done via a neural net.
The differentiable structural information (DSI) is given as $\mathcal{H}^{\mathcal T_{\operatorname{net}}}(G; \mathbf{Z}; \mathbf \Theta)$. It takes the same form of Eq. (\ref{kse}) where the assignment is derived by a neural net with parameters $\Theta$.
Given $G$, we consider DSI minimization as follows, 
 \vspace{-0.03in}
 \begin{align}
 \label{eq.opt_z}
    \mathbf{Z}^*, \mathbf \Theta^* = \arg_{\mathbf{Z},\mathbf \Theta} \min \mathcal{H}^{\mathcal T_{\operatorname{net}}}(G; \mathbf{Z}; \mathbf \Theta).
 \end{align}
We learn node embeddings $\mathbf{Z}^*$ and parameter $\mathbf \Theta^*$ to derive the level-wise parent assignment for $G$.
Then, the optimal partitioning tree $\mathcal{T}_{\operatorname{net}}^*$ of $G$ is constructed by the assignment,
so that \emph{densely connected nodes have a higher probability to be assigned to the same parent in $\mathcal{T}_{\operatorname{net}}^*$, minimizing the conductance and presenting the cluster structure.}
Meanwhile, node embeddings jointly learned in the optimization provide geometric notions to refine clusters in representation space.

Next, we introduce the theoretical guarantees of $\mathcal{T}_{\operatorname{net}}^*$. 
We first define the notion of equivalence relationship in partitioning tree $\mathcal{T}$ with height $H$. 
If nodes $i$ and $j$ are in the same module at the level $h$ of $\mathcal{T}$ for $h=1, \cdots, H$, 
    they are said to be in equivalence relationship, denoted as $i \overset{h}{\sim} j$.
\begin{mymath}
\begin{theorem}[\textbf{Bound}]
\label{theorem.approx}
 %For any graph $G$, $\mathcal{T}^*$ is the optimal partitioning tree given by  \citet{li2016structural},
% and $\mathcal{T}_{\operatorname{net}}^*$ with height $H$ is the partitioning tree given from Eq. (\ref{eq.opt_z}).
  For any graph $G$, $\mathcal{T}^*$ is the optimal partitioning tree  of  \cite{li2016structural},
 and $\mathcal{T}_{\operatorname{net}}^*$ with height $H$ is the partitioning tree given by Eq. (\ref{eq.opt_z}).
For any pair of leaf embeddings $\boldsymbol z_i$ and $\boldsymbol z_j$ of $\mathcal{T}_{\operatorname{net}}^*$, 
 there exists bounded real functions $\{f_h\}$ and constant $c$,
 such that for any $0 < \epsilon < 1$, $\tau \leq \mathcal{O}(1/ \ln[(1-\epsilon)/\epsilon])$ and  $\frac{1}{1 + \exp\{-(f_h(\boldsymbol z_i, \boldsymbol z_j) - c) / \tau\}} \geq 1 - \epsilon$ satisfying $i\overset{h}{\sim} j$, we have the following bound hold,
 \vspace{-0.05in}
    \begin{align}
        \lvert \mathcal{H}^{\mathcal{T}^*}(G) - \mathcal{H}^{\mathcal{T}_{\operatorname{net}}^*}(G) \rvert \leq \mathcal{O}(\epsilon).
    \end{align}
\end{theorem}
 \vspace{-0.03in}
\begin{proof}
    Refer to Appendix II, Theorem III.3.
\end{proof}
\end{mymath}
% That is, the difference between structural entropy calculated by our neural approach and \citet{li2016structural} is quite small, and bounded.
% Thus, the optimal partitioning tree $\mathcal{T}^*$ is well approximated by the tree $\mathcal{T}_{\operatorname{net}}^*$ from Eq. (\ref{eq.opt_z}).
\noindent That is, the structural entropy calculated by our $\mathcal{T}_{\operatorname{net}}^*$ is well approximated to that of \cite{li2016structural}, and the difference is bounded under slight constraints.
Thus, $\mathcal{T}_{\operatorname{net}}^*$ serves as an alternative to the optimal $\mathcal{T}^*$ for graph clustering.

\section{Theory: Structural Information \& Graph Clustering}

In this section, we derive a more efficient formulation of differential structural information, achieving linear complexity, and provide the new theoretical results on the advantages of structural information for graph clustering.

\subsection{Reducing Complexity}

Eq. (\ref{kse}) requires all edges to compute structural information at each level.
Instead of enumerating the edges, we reduce the complexity by searching all the parent nodes whose amount is much less than that of edges.
With the assignment matrices, we rewrite the differentiable structural information as follows, 
    \begin{equation}
            \resizebox{1.03\hsize}{!}{$
        \mathcal{H}^{\mathcal{T}}(G^h;G^{h-1}, h)=-\frac{1}{V}\sum_{k=1}^{N_h}(d^h_k-w^h_{kk})\log \frac{d^h_k}{\sum_{l=1}^{N_{h-1}}C_{kl}^{h-1}d_l^{h-1}},
            \label{eq.new_ksi}
                 $}
   \end{equation}
where $G^{h-1}$ is induced by the assignment matrix $\mathbf{S}$ in Eq. (\ref{eq.s}) at the $(h-1)$ level, and $d_i^{h}$ is the degree of $i$-th node in graph $G^h$.
The equivalence and variables in Eq. (\ref{eq.new_ksi}) are stated in the following theorem.
\begin{mymath}
\begin{theorem}[\textbf{Node-wise Formulation}]
    At each level in the partitioning tree, the structural information calculated in Eq. (\ref{eq.new_ksi}) is equal to that of Eq. (\ref{eq.s}).
\end{theorem}
\begin{proof}
    In Eq. (\ref{kse}), we analyze it term by term. First, 
    \begin{equation}
            \resizebox{0.89\hsize}{!}{$
        g^h_k = V^h_k - \sum\limits_{(i,j)\in \mathcal{E}}S^h_{ik}S^h_{jk}w_{ij} = \sum\limits_{(i,j)\in \mathcal{E}}S^h_{ik}(1-S^h_{jk})w_{ij},
                 $}
   \end{equation}
    then we set $w^h_{kl}:=\sum_{i,j}^N S^h_{ik}S^h_{jk}w_{ij}=\sum\limits_{(i,j)\in \mathcal{E}}S^h_{ik}S^h_{jk}w_{ij}$ to define $\hat{g}^h_k$ and $d^h_k$ by
        \begin{equation}
            \resizebox{1.02\hsize}{!}{$
        \hat{g}^h_k = \sum\limits_{l, l\neq k}w^h_{kl}=\sum_{l, l\neq k}\sum\limits_{(i,j)\in \mathcal{E}}S^h_{ik}S^h_{jk}w_{ij} =\sum\limits_{(i,j)\in \mathcal{E}}S^h_{ik}(1-S^h_{jk})w_{ij},
                 $}
   \end{equation}
       \begin{equation}
            \resizebox{0.68\hsize}{!}{$
         d^h_k =\sum\limits_l w^h_{kl}=\sum\limits_{(i,j)\in \mathcal{E}}S^h_{ik}w_{ij} =\hat{g^h_k} + w^h_{kk}.
                 $}
   \end{equation}
        % \hat{g}^h_k &= \sum_{l, l\neq k}w^h_{kl}=\sum_{l, l\neq k}\sum\limits_{(i,j)\in \mathcal{E}}S^h_{ik}S^h_{jk}w_{ij} =\sum\limits_{(i,j)\in \mathcal{E}}S^h_{ik}(1-S^h_{jk})w_{ij},\\
        % d^h_k& =\sum_l w^h_{kl}=\sum\limits_{(i,j)\in \mathcal{E}}S^h_{ik}w_{ij} =\hat{g^h_k} + w^h_{kk}.
    We observe that $g^h_k=\hat{g}^h_k$ and $d^h_k=V^h_k$, substituting them into Eq. (\ref{kse}), the proof is completed.
\end{proof}
\end{mymath}
\noindent While keeping the space complexity unchanged, we  significantly reduce the time complexity of calculating structural information. 
Note that, the formulation of Eq. (\ref{eq.new_ksi}) achieves the linear complexity with respect to the number of nodes.

\subsection{Flexiblity of Partitioning Tree}
We show that the relaxation of the partitioning tree with redundant nodes does not affect the value of structural information under certain conditions, thanks to the logarithmic in the definition.
The flexibility of partitioning tree is claimed in the following theorem.
\begin{mymath}
\begin{theorem}[\textbf{Flexiblity}]
\label{theorem.flex}
    $\forall G$, given a partitioning tree $\mathcal{T}$ and adding a node $\beta$ to get a relaxed $\mathcal{T}'$ , the structural information remains unchanged, $H^{\mathcal{T}}(G)=H^{\mathcal{T}'}(G)$, if  one of the following conditions holds:
    \begin{enumerate}
    %  \item $\beta$ as a leaf node, and the corresponding node subset $T_\beta$ is an empty set.
        \item $\beta$ as a leaf node, and its module $T_\beta$ is an empty set.
      \item $\beta$ is inserted between node $\alpha$ and its children nodes so that the modules $T_\beta$ and $T_\alpha$ are equal.
\end{enumerate}
\end{theorem}
\vspace{-0.02in}
\begin{proof}
    Refer to Appendix II, Theorem IV.2.
\end{proof}
\end{mymath}
\vspace{-0.02in}
\noindent The theorem above will guide the neural architecture design for constructing the deep partitioning tree $\mathcal{T}_{\operatorname{net}}^*$.

\vspace{-0.07in}
\subsection{Handling Imbalanced Clusters}

We real another interesting fact that the structural entropy objective is preferable for detecting minority clusters.
To formally describe this, we consider a scenario where an imbalanced graph $G$ contains two majority clusters $V_1$, $V_2$ and one minority \( V_\varepsilon \).
Partitioning $V_\varepsilon$ further reduces the structural information of  the partitioning tree, thereby enhancing  the identifiability of the minority cluster.
% We demonstrate , given the imbalance of the graph.
% In other words,  our approach is easier to detect minority clusters.
\begin{mymath}
\begin{theorem}[\textbf{Identifiability}] \label{thm:imbalanced}
Given an imbalanced graph \( G \) of \( N \) nodes consisting of the majority clusters $V_1$, $V_2$ and minority \( V_\varepsilon \),
such that \( |V_1| \gg |V_\varepsilon| \), \( |V_2| \gg |V_\epsilon| \) and \( N \gg 3 \),
 structural information minimization makes \( V_\varepsilon \) distinguishable from the majority clusters given the following inequality 
    \begin{align}
        \mathcal{H}(G, V_2) + \mathcal{H}(G, V_1 \cup V_\varepsilon) > \sum\nolimits_{i\in\{1,2,\varepsilon\}}\mathcal{H}(G, V_i)
    \end{align}
 holds for the slight condition that $|V_1|\ge g_\alpha+1$ where $g_\alpha$ is the cut between clusters $V_1$ and $V_2$.
\end{theorem}
\vspace{-0.02in}
\begin{proof}
    Refer to Appendix II, Theorem IV.3.
\end{proof}
\end{mymath}
\vspace{-0.02in}
\noindent 
Nodes within the cluster are typically densely connected while there are fewer links connecting the nodes of different clusters.
The condition is easy to satisfy in real graphs.

\emph{That is, we obtain the graph clustering objective with linear time complexity, the differential structural information in Eq. (\ref{eq.new_ksi}), which no longer requires the knowledge of the number of clusters and is suitable to identify minority clusters in real-world graphs often exhibiting skewed cluster distributions.}

\section{\texttt{LSEnet}: Learning Partitioning Tree in Hyperbolic Space}
\label{sec:lsenet}

We design a novel Lorentz Structural Entropy Neural Network (\texttt{LSEnet}), 
which aims to learn the optimal partitioning tree $\mathcal{T}_{\operatorname{net}}^*$ in the Lorentz model of \textbf{hyperbolic space}.
\texttt{LSEnet} constructs the partitioning tree in a bottom-up fashion where we first embed leaf nodes of the tree, and then recursively learn the parent nodes, leveraging the clustering objective of differential structural information.
In particular, we learn the partitioning trees from both the original and augmented structures, which serve as the different views in Lorentz tree contrastive learning (Sec. \ref{sec:TCL}).

\subsection{Why Hyperbolic Space?}
First, we demonstrate the reason why we opt for the hyperbolic space, rather than the traditional Euclidean space.
We introduce a quantitative metric of embedding distortion which is defined as $\frac{1}{|\mathcal V|^2} \sum_{ij}\left| \frac{d_G(v_i, v_j)}{d(\mathbf x_i, \mathbf x_j)}-1\right|$, where each node $v_i \in \mathcal V$ is embedded as $\mathbf x_i$ in representation space. $d_G$ and $d$ denote the distance in the graph and the space, respectively.
\begin{theorem}[Tree $\mathcal T$  and Hyperbolic Space \cite{sarkar2012low}]
\label{thm:tree}
% Any tree whose edges length bounded below a constant about $\frac{1+\epsilon}{\epsilon}$ for arbitrary $\epsilon$, can be embedded into hyperbolic space with distortion bounded by $1+\epsilon$.
% For any 
$\forall \mathcal T$, scaling all edges by a constant so that the edge length is bounded below $\nu\frac{1+\epsilon}{\epsilon}$, there exists an embedding in hyperbolic space that the distortion
overall node pairs are bounded by $1+\epsilon$. \emph{($\nu$ is a constant detailed in Appendix V)}
\end{theorem}
\noindent That is, hyperbolic space is well suited to embed the partitioning tree. Note that, Theorem \ref{thm:tree} does not hold for Euclidean space or other Riemannian manifold.

% \textbf{Overall architecture} of \texttt{LSEnet} is sketched in Figure \ref{fig:lsenet}.
% %In hyperbolic space, we learn the partitioning tree in the bottom-up manner. \texttt{LSEnet} is built up by stacking Lorentz assigners, where each assigner is responsible for a tree layer, \emph{self-supervised by the objective of our DSE (Eq. \ref{kse})}.
% In hyperbolic space, 

%%%%%%%%%%%%%%%%%%%%%%%%%%%%%%%%%%%%%%%%%%%%%%%%%%%%%%%%%%%
%Huang
% From Proposition \ref{DSE} we notice that the $H$-dimensional structural entropy of graph $G$ can be calculated layer by layer from down to top, which means that we can construct the optimal partitioning tree of $G$ layer-wisely in a down-top way.

% The First thing is to get leaf node embeddings, not only need we consider the graph topology and graph node attributes, but also the tree embedding in hyperbolic space. Therefore, we should adopt some transforms in hyperbolic space meanwhile handling them with graph topology.
\vspace{-0.03in}
\subsection{Embedding Leaf Nodes}
\label{leaf}
%Here, we study the 
We design a Lorentz convolution layer to learn leaf embeddings in hyperbolic space, 
$\operatorname{LConv}: \boldsymbol x_i  \to  \boldsymbol z^H_i, \forall v_i$, 
where $\boldsymbol x_i\in \mathbb L^{\kappa, d_0}$  and $\boldsymbol z^H_i \in \mathbb L^{\kappa, d_{\mathcal T}}$ are the node feature and leaf embedding, respectively.
In the partitioning tree of height $H$, 
the level of nodes is denoted by superscript of $h$
%the nodes of $h$-th level node embedding $\boldsymbol z^h_{\cdot}$ is denoted as  $d_h$, 
and we have $h=H, \cdots, 1, 0$ from leaf nodes (bottom) to root (top), correspondingly.

We adopt attentional aggregation in $\operatorname{LConv}$ and specify the operations as follows.
First, dimension transform is done via a \emph{Lorentz linear operator} \cite{chen2022fullyb}.
For node $\boldsymbol x_i\in \mathbb L^{\kappa, d_0}$, the linear operator is given as 
\vspace{-0.03in}
\begin{align}
    \operatorname{LLinear}(\boldsymbol{x}) = 
    \begin{bmatrix}
        \sqrt{\lVert h(\mathbf{W}\boldsymbol{x},\boldsymbol{v}) \rVert^2 - \frac{1}{\kappa}} \ \\
        h(\mathbf{W}\boldsymbol{x},\boldsymbol{v}) \
    \end{bmatrix}
    \in \mathbb L^{\kappa, d_{\mathcal T}},
    \vspace{-0.03in}
\end{align}
where $h$ is a neural network, and $\mathbf{W}$ and $\boldsymbol{v}$ are parameters.
Second, we derive attentional weights from the \emph{self-attention mechanism}. 
% With $\mathbf{Q}, \mathbf{K}, \mathbf{V}\in \mathbb L^{\kappa, d_{\mathcal T}}$ given from $\mathbf{X} \in \mathbb L^{\kappa, d_0}$ via $\operatorname{LLinear}$, the attentional weight is calculated as 
% \begin{align}
%     \Omega_{ij} = \operatorname{LAtt}(\mathbf{Q}, \mathbf{K})=
%      \frac{\exp(-  \gamma   d^2_{\mathbb{L}}(\boldsymbol{q}_i, \boldsymbol{k}_j)  )}
%      {\sum_{l=1}^{N}\exp( -  \gamma   d^2_{\mathbb{L}}(\boldsymbol{q}_i, \boldsymbol{k}_l)  )},
%      \label{LAtt}
%          \vspace{-0.07in}
% \end{align}
% where $\gamma$ is $1/\sqrt{N}$, and $\mathbf{X} \in \mathbb L^{\kappa, d}$ denotes a matrix whose row vectors represent the points in hyperbolic space $\mathbb L^{\kappa, d}$.
Concretely, the attentional weight $\omega_{ij}$ between nodes $i$ and $j$ is calculated as 
\vspace{-0.03in}
\begin{align}
    a_{ij}&=\operatorname{LeakyReLU}(\mathbf{W}[\boldsymbol{q}_i||\boldsymbol{k}_j)])
\\
    \omega_{ij} &= \operatorname{LAtt}(\mathbf{Q}, \mathbf{K})=
     \frac{\exp(a_{ij})}{\sum_{(i,l)\in \mathcal{E}}\exp(a_{il})},
     \label{LAtt}
     \vspace{-0.03in}
\end{align}
where $\mathbf{W}\in\mathbb{R}^{1\times2d}$ is learnable parameters, $[\cdot,\cdot]$ is concatenate operation, $\boldsymbol{q}$ and $\boldsymbol{k}$ are the query and key vector collected in row vector of $\mathbf Q$ and $\mathbf K$, respectively. The queries and keys are derived from node feature $\boldsymbol x$ via $\operatorname{LLinear}$ with different parameters.
Third, weighted aggregation is considered as the \emph{arithmetic mean} among manifold-valued vectors.
Given a set of points $\{\boldsymbol{x}_i\}_{i=1, \cdots, N}$ in the Lorentz model $\mathbb L^{\kappa, d_{\mathcal T}}$, 
\vspace{-0.03in}
\begin{align}
    \operatorname{LAgg}(\boldsymbol{\omega}, \mathbf{X}) = \frac{1}{\sqrt{-\kappa}} \sum_{i=1}^N \frac{ \omega_i}{\lvert \lVert \sum_{j=1}^N \omega_j\boldsymbol{x}_j \rVert_\mathbb{L} \rvert} \boldsymbol{x}_i,
     \label{LAgg}
     \vspace{-0.03in}
\end{align}
where $\boldsymbol{\omega}$ is the weight vector, and $\boldsymbol x$ are summarized in $\mathbf{X}$.
The augmented form is that for $\mathbf{\Omega}=[\boldsymbol{\omega}_1, \cdots, \boldsymbol{\omega}_N]^\top$, 
we have $\operatorname{LAgg}(\mathbf{\Omega}, \mathbf{X}) = [\boldsymbol \mu_1, \cdots, \boldsymbol  \mu_N]^\top$, 
where the weighted mean is $\boldsymbol  \mu_i =\operatorname{LAgg}(\boldsymbol{\omega}_i, \mathbf{X})$.
Overall, the Lorentz convolution layer is formulated as,
\vspace{-0.03in}
\begin{align}
                \resizebox{0.89\hsize}{!}{$
\operatorname{LConv}(\mathbf{X}|\mathbf{A})=\operatorname{LAgg}(\operatorname{LAtt}(\mathbf{Q}, \mathbf{K})\odot \mathbf{A}, \operatorname{LLinear}(\mathbf{X}) ),
$}
\vspace{-0.03in}
\end{align}
where $\odot $ is the Hadamard product, masking the attentional weight if the corresponding edge does not exist in the graph.

\vspace{-0.03in}
\subsection{Learning Parent Nodes}
\label{parent}

A primary challenge is that, in the partitioning tree, \emph{the node number at each internal level is unknown.}
To address this issue, we introduce a simple yet effective method, setting a large enough node number $N_h$ at the $h$-th level.
A large $N_h$ may introduce redundant nodes and result in a relaxed partitioning tree. According to Theorem \ref{theorem.flex},  the redundant nodes in the partitioning tree do not affect the value of structural entropy, and finally present as empty leaf nodes by optimizing our objective.
Theoretically, if an internal level has insufficient nodes, the self-organization of the graph can still be described by multiple levels in the partitioning tree. 

Without loss of generality, we consider the assignment between $h$-th and $(h-1)$-th levels given node embeddings $\boldsymbol z^{h}  \in \mathbb L^{\kappa, d_{\mathcal T}}$  at $h$-th level.
Recalling Definition \ref{def.C}, the $i$-th row of assignment $\mathbf C^h \in \mathbb R^{N_h \times N_{h-1}}$ describes the belonging of $i$-th node at $h$-th to the parent nodes at $(h-1)$-th.
We design a Lorentz assigner following the intuition that neighborhood nodes in the graph tend to have similar assignments.
Concretely, we leverage Multilayer Perceptron $\operatorname{MLP}$ to learn the assignment from embeddings, 
and meanwhile the similarity is parameterized by $\operatorname{LAtt}$ defined in Eq. \ref{LAtt}. 
Thus,  the Lorentz assigner is formulated as follows,
\vspace{-0.03in}
\begin{align}
   \mathbf{C}^h =\operatorname{LAtt}(\mathbf{Q}, \mathbf{K})\sigma(\operatorname{MLP}(\mathbf Z^h)),
    \label{assignment}
    \vspace{-0.03in}
\end{align}
where $\mathbf{A}^h$ is the graph structure at $h$-th level of the tree.
$\sigma$ denotes the row-wise $Softmax$ function.

The remaining task is to infer the node embeddings $\boldsymbol z^{h-1} \in \mathbb L^{\kappa, d_{\mathcal T}}$ at $(h-1)$-th level. 
As reported in \cite{chami2020trees}, a parent node is located at the point that has the shortest path to all the child nodes at the $h$-th level, correspondingly, the parent node is the  Fr\'{e}chet mean of the child nodes in the manifold.
The challenge here is that Fr\'{e}chet mean regarding the canonical distance has no closed-form solution \cite{DBLP:conf/icml/LouKJBLS20}.
Alternatively, we consider the geometric centroid for the squared Lorentz distance, where the weights are given by the soft assignment $\mathbf{C}^h$,
\vspace{-0.03in}
\begin{align}
    \boldsymbol z^{h-1}_j=\arg_{   \boldsymbol z^{h-1}_j} \min \sum\nolimits_{i=1}^N c_{ij}d^2_{\mathbb L}(\boldsymbol z^{h-1}_j, \boldsymbol z^h_i).
    \label{centroid}
    \vspace{-0.03in}
\end{align}

Solving the optimization constrained in the manifold $\mathbb L^{\kappa, d_{\mathcal T}}$, we derive the closed-form solution of parent node embeddings,
\vspace{-0.1in}
\begin{align}
    \mathbf{Z}^{h-1}=\operatorname{LAgg}(\mathbf{C}^h, \mathbf{Z}^h) \in \mathbb L^{\kappa, d_{\mathcal T}},
    \label{parentNode}
\end{align}
which is the augmented form of Eq. (\ref{LAgg}).
\begin{mymath}
\begin{theorem}[Geometric Centroid]
\label{theorem.centroid}
In Lorentz model $\mathbb L^{\kappa, d_{\mathcal T}}$ of hyperbolic space,
for any set of points $\{\boldsymbol z^{h}_i\}$,
the arithmetic mean is given as follows,
\vspace{-0.03in}
\begin{align}
    \boldsymbol z^{h-1}_j =
\frac{1}{\sqrt{-\kappa}} \sum_{i=1}^N \frac{ c_{ji}}{\lvert \lVert \sum_{l=1}^n c_{jl} \boldsymbol z^{h}_l  \rVert_\mathbb{L} \rvert} \boldsymbol z^{h}_i \in \mathbb L^{\kappa, d_{\mathcal T}},
\vspace{-0.03in}
\end{align}
and is the closed-form solution of the geometric centroid defined in the minimization of Eq. \ref{centroid}.
\end{theorem}
\end{mymath}
\vspace{-0.03in}
\noindent \textbf{Remark.}
In fact, the geometric centroid in Theorem \ref{theorem.centroid} is also equivalent to the gyro-midpoint in Poincar\'{e} ball model of hyperbolic space, detailed in Appendix II, Theorem VI.2. Note that attentive aggregation for parent node embedding is uni-directional, different from that in typical graph neural nets.

According to Eq. (\ref{assignment}), it requires the graph structure  $\mathbf{A}^{h-1}$ among the parent nodes at $(h-1)$-th level to derive the assignment. We give the adjacency matrix of the $(h-1)$-th level graph as follows,
\vspace{-0.03in}
\begin{align}
    \mathbf{A}^{h-1} &= (\mathbf{C}^h)^\top \mathbf{A}^h \mathbf{C}^h.
    \label{structure}
    \vspace{-0.03in}
\end{align}
We recursively utilize Eqs. (\ref{assignment}), (\ref{parentNode}) and (\ref{structure}) to build the partitioning tree from bottom to top in hyperbolic space.

\vspace{-0.07in}
\subsection{Hyperbolic Partitioning Tree}
In the principle of structural information minimization, the optimal partitioning tree is constructed to describe the self-organization of the graph \cite{li2016structural,wu2023sega}.
In the continuous realm, 
\emph{\texttt{LSEnet} learns the optimal partitioning tree in hyperbolic space} by minimizing the objective in Eq. (\ref{kse}).
DSI is applied on all the level-wise assignment $\mathbf C$'s, which are parameterized by neural networks in Sec. \ref{leaf} and \ref{parent}.
We suggest to place the tree root at the origin of $\mathbb L^{\kappa, d_{\mathcal T}}$, so that the learnt tree enjoys the symmetry of Lorentz model.
Consequently, the hyperbolic partitioning tree describes the graph's self-organization and clustering structure in light of Theorem \ref{theorem.conductance}.

\vspace{-0.05in}
\section{Lorentz Tree Contrastive Learning}
\label{sec:TCL}
% Although we obtain leaf embeddings by integrating the graph attributes and structure into \texttt{LSEnet}, the clustering objectives still relies on graph structure, ignoring the meaning full graph embeddings, which contains graph semantics.

% In this section, we introduce the tree contrastive learning (TCL), which implicitly contrast partitioning trees when contrasting leaf embeddings. We first construct Lorentz boost group that is key to TCL, then we give a general formula in the form of Graph InfoNCE~\cite{VelickovicFHLBH19}.

The partitioning tree above resides in the hyperbolic space, and thus we propose to refine its hyperbolic representation through Lorentz tree contrastive learning, enhancing the graph semantics from a representational perspective. 
In this contrastive learning framework, we develop a novel \emph{neural Lorentz boost} for consistency insurance  and a \emph{tree contrastive loss} considering the tree branches in light of clustering assignment.

\vspace{-0.1in}
\subsection{\textbf{Neural Lorentz Boost}}

Different from graph contrastive learning, the projection head in our case further requires the consistency insurance—maintaining the parent hierarchies of the partitioning tree as a unity. That is, the level-wise child-parent assignment cannot be disrupted by this projection head.

To achieve this, we leverage the \textit{Lorentz group} $G^L = \{L_i\}$, whose elements are isometries over Lorentz model.
For any $\boldsymbol{x} \in \mathbb{L}^{\kappa, d}$, a Lorentz transformation can be written as follows,
\vspace{-0.2in}
\begin{align}
  L(\boldsymbol{x}) = \mathbf{L} \boldsymbol{x} \in \mathbb{L}^{\kappa, d}, \quad
\mathbf{L} =
\begin{bmatrix}
w & \boldsymbol{v}^\top \\
\boldsymbol{v} & \mathbf{W}
\end{bmatrix},
\label{eq. lorentz boost}
\vspace{-0.3in}
\end{align}
where $\mathbf{W} \in \mathbb{R}^{d \times d}$ is a matrix, $w$ is a scaling scalar, and $\boldsymbol{v} \in \mathbb{R}^{d}$ is a boost vector.
Note that Lorentz rotations act tangentially to the radial direction. They rotate the partitioning tree and do not affect representation refinement due to  rotational symmetry. 
Thus, we consider the Lorentz boosts, which serve as a linear transformation to operate on the partitioning tree in the Lorentz model of hyperbolic space. 
We derive a closed-form Lorentz boost parameterized by $\boldsymbol{\beta} \in \mathbb{R}^{d}$, \textbf{\emph{maintaining the consistency insurance in partitioning tree contrasting}} (as shown in Theorems VI.1 and VI.2).

\begin{mymath}
  \begin{theorem}[\textbf{Closed-from Lorentz Boost}]
    \label{theorem. lorentz boost}
    Given the construction in Eq. (\ref{eq. lorentz boost}) and a parameter vector $\boldsymbol{\beta} \in \mathbb{R}^{d}$, Lorentz boosts have a closed form with components  given as
    \begin{align}
      w&=\frac{1}{\sqrt{1 - \| \boldsymbol{\beta} \|^2_\mathbb{L}}}  \\
      \boldsymbol{v}&=-w \cdot \boldsymbol{\beta} \\
      \mathbf{W}&= \frac{(w - 1)}{\| \boldsymbol{\beta} \|^2_\mathbb{L}} \cdot \boldsymbol{\beta} \boldsymbol{\beta}^\top + \mathbf{I}_d
    \end{align}
  \end{theorem}
  \begin{proof}
    Refer to Appendix II.G, Theorem VI.1.
  \end{proof}
\end{mymath}

\begin{mymath}
\begin{theorem}[\textbf{Manifold Preserving}]
\label{theorem. lg}
    Given the construction of $\mathbf L$ in Theorem~\ref{theorem. lorentz boost}, and scalars $a, b$, the following properties  holds for any  $\boldsymbol{x}, \boldsymbol{y} \in \mathbb{L}^{d_1}$.
    \begin{itemize}
        \item[1)] $L(a\boldsymbol{x}+b\boldsymbol{y})=aL(\boldsymbol{x})+ bL(\boldsymbol{y})$;
        \item[2)] $\langle L(\boldsymbol{x}), L(\boldsymbol{y})  \rangle_\mathbb{L}=\langle \boldsymbol{x}, \boldsymbol{y}\rangle_\mathbb{L}$,
    \end{itemize}
where $\langle \boldsymbol{x}, \boldsymbol{y}\rangle_\mathbb{L}$ denotes the Minkowski inner product.
\end{theorem}
\begin{proof}
    Refer to Appendix II.G, Theorem VI.2.
\end{proof}
\end{mymath}
\noindent Note that, the manifold-preserving property is given by claim (2) in the theorem above. It is easy to verify that $G$ is closed on itself, i.e., $L_i\circ L_j \in G^L$, $\forall L_i, L_j\in G^L$.
These properties also induce the following result.
\begin{mymath}
\begin{theorem}[\textbf{Consistence Insurance}]\label{thm:consist}
  Given a set of points $\{\boldsymbol{z}_i\}$ in the Lorentz model of hyperbolic space, $\boldsymbol{z}_i\in \mathbb L^{\kappa, d_1}$ and their midpoint $\boldsymbol{z}\in \mathbb L^{\kappa, d_1}$ is constructed by Eq. (\ref{LAgg}), then $L(\boldsymbol{z})\in \mathbb L^{\kappa, d_2}$ is the midpoint of $\{L(\boldsymbol{z}_i)\}$,  $L(\boldsymbol{z}_i)\in \mathbb L^{\kappa, d_2}$, if $L$ is a transformation in the Lorentz group as stated in Theorem \ref{theorem. lg}.
\end{theorem}
\begin{proof}
  Refer to Appendix II.H, Theorem VI.3.
\end{proof}
\end{mymath}
\vspace{-0.03in}
\noindent The theorem above establishes the operation that propagates the action on tree nodes to their ancestors.
In other words, only leaf contrast is needed in tree contrastive learning, and the consistence is maintained throughout the entire  partitioning tree thanks to the properties of the proposed transformations.

\textbf{\emph{Remark}.}
We notice that recent studies propose a series of manifold-preserving operations in the Lorentz model of hyperbolic space~\cite{chen2022fullyb,SunL23AAAI}, 
but none of them satisfies consistence insurance and thereby cannot be utilized  for the tree contrastive learning in hyperbolic space.

\vspace{-0.15in}
\subsection{Tree Contrastive Loss}
In this part, we introduce the loss function for contrasting partitioning trees in the Lorentz model of hyperbolic space. For each node, the corresponding node in the counterpart view is treated as positive sample; otherwise, negative samples. Accordingly, Tree Contrastive Loss (TCL) is formulated as
\vspace{-0.07in}
\begin{align}
  \mathcal{L}_{\text{TCL}}(\mathbf{Z}, \mathbf{Z'})=\frac{1}{N}\sum_{i=1}^{N}\frac{\exp(d_\mathbb{L}(L(\boldsymbol{z_i}), L(\boldsymbol{z'_i})))}{\sum_{j=1}^{N}w_{ij}\exp(d_\mathbb{L}(L(\boldsymbol{z_i}), L(\boldsymbol{z'_j})))},
  \label{TCL}
  \vspace{-0.07in}
\end{align}
where the reweighting factor is defined with the level-wise assignment, $w_{ij}=\sum_{k=1}^{N_{H-1}}S^H_{ik}S^H_{jk}$.
That is, TCL considers the (hierarchical) tree branches in the partitioning trees, essentially differing itself from graph contrastive learning.

\emph{\textbf{Differences to Graph Contrastive Learning (GCL).}}
GCL refines representations through exploring the similarity among all nodes in the graph.
On the contrary, we contrast the partitioning trees of the graph—nodes and their parent hierarchies—under consistency insurance, so that we propose the \emph{Neural Lorentz Boost} with linearity and isometry, maintaining  the partitioning tree as a whole, and \emph{Tree Contrastive Loss}, considering the partitioning structures.
Also, GCL is typically conducted in Euclidean space, while we operate in hyperbolic space given its alignment to tree structures.

\emph{\textbf{Remaining Issues.}}
Although tree contrastive learning refines node representations in hyperbolic space, it presents the following two limitations: 
On the one hand, view generation in this case is non-trivial. A straightforward way is to perform graph augmentation followed by LSEnet to generate a new hyperbolic view, but  it may introduce noise that undermines model expressiveness as evidenced in Ablation Study.
On the other hand, computing the loss function of  Eq. (\ref{TCL}) yields a quadratic complexity with respect to $|\mathcal V|$, the number of nodes,  preventing the model scaling up to large graphs.  
\tabularnewline

\vspace{-0.05in}
\section{Augmented Structural Information Learning} 

In this section, we first elaborate on a theoretical insight that the structural entropy bounds contrastive learning, allowing for efficient and implicit tree contrastive learning.
Integrating the theoretical establishment and neural net designs, we finally propose a novel deep graph clustering approach, \emph{Augmented Structural Information Learning} (\texttt{ASIL}).

\vspace{-0.1in}
\subsection{Theoretical Insight: Connection between Structural Entropy and Contrastive Learning}
\label{sec:virtualGraph}

We uncover an inherent connection between structural entropy and contrastive learning, while simultaneously addressing the two aforementioned limitations. 

Specifically, we demonstrate that the objective of structural entropy aligns closely with that of contrastive learning in terms of graph conductance, 
and this connection is formally established by introducing a virtual graph constructed from node representations. 
First, prior to the mathematical formulation, we provide an intuitive discussion of their relationship: 
in hyperbolic space, the tree contrastive learning pulls together node representations belonging to the same partition (i.e., tree branch), 
while pushing apart those from different partitions;
the structural entropy operates directly on the graph structure—namely, the adjacency matrix—and generates an optimal partitioning tree primarily based on the node connectivity. 
Thus, both structural entropy and contrastive learning explore the node similarity, 
but from structural and representational perspectives, respectively. 
Second, we introduce a virtual graph constructed from node representations to bridge these two viewpoints.
In particular, we perform the \emph{Lorentz Boost Construction} defined as follows,
\vspace{-0.04in}
\begin{align}
\tilde{\mathbf A}_{ij} = \exp\left(-d_\mathbb{L}\left(L(\boldsymbol{z}_i), L(\boldsymbol{z}_j)\right)/t\right),
\label{vAdj}
\vspace{-0.08in}
\end{align}
where $d_\mathbb{L}$ denotes the induced distance of Minkowski metric in the hyperbolic space, and $t$ is the temperature hyperparameter.
The transformation $L$ corresponds to the Lorentz boost within the tree contrastive learning, and $\boldsymbol{z}$ refers to the optimized node representation.
In practice, we sparsify the virtual adjacency matrix by retaining $k$-nearest neighbors regarding $d_\mathbb{L}$.
An interesting finding is formally stated in the theorem below.

\begin{mymath}
\begin{theorem}[\textbf{Connection between Structural Entropy and Tree Contrastive Learning}] \label{thm:se_tcl}
Given an induced weighted graph $\tilde{G} = (V, \tilde{A})$, where $\tilde{A}$ is the virtual adjacency matrix constructed via Eq. (\ref{vAdj}),
the tree contrastive learning with the loss function of Eq. (\ref{TCL})  minimizes the graph conductance of $\tilde{G}$, and is upper bounded by the $H$-dimensional structural information $\mathcal{H}^H(\tilde{G})$.
\end{theorem}
 \vspace{-0.05in}
\begin{proof}
Refer to Appendix II.I, Theorem VII.1.
  \end{proof}
 \vspace{-0.05in}
\end{mymath}
\noindent The theorem above claims that structural entropy indeed bounds the tree contrastive learning defined in Eq. (\ref{TCL}). The key point is Eq. (\ref{TCL}) has been proven to align with minimizing the graph conductance, which is bounded by structural entropy as formalized in Theorem \ref{theorem.conductance}.

Our theoretical insight offers a promising alternative for tree contrastive learning, addressing its limitation in view generation and computational complexity.
Concretely, it is preferable to perform structural information minimization over the virtual graph, which implicitly contrasts the partitioning trees in the hyperbolic space. 
The effect of contrasting two hyperbolic views is captured in the virtual graph of Lorentz boost construction in a parameterized fashion, and thus view generation is no longer required.
As for computational complexity, we achieve the linear complexity regarding $|\mathcal V|$ with the reformulation in Eq. (\ref{eq.new_ksi}), compared to the quadratic complexity of contrastive learning in Eq. (\ref{TCL}).

\vspace{-0.08in}
\subsection{Augmented Structural Entropy: Integrating Partitioning Tree Construction and Contrastive Learning}

This part presents our deep graph clustering approach —\emph{Augmented Structural Information Learning}, referred to as \texttt{ASIL}.
Based on the established theory of differential structural information and findings on its connection to tree contrastive learning, 
\texttt{ASIL} seamlessly integrates partitioning tree construction and (implicit) contrastive learning in the hyperbolic space, allowing for effective and efficient node clustering without the knowledge of the number of clusters.

Specifically, \texttt{ASIL} first fuses the structural and representational views, and then feed them into 
\texttt{LSEnet}  (Sec. \ref{sec:lsenet}) to  learn the hyperbolic partitioning tree for clustering.
The overall architecture is sketched in Fig. \ref{fig:overall}.
The key design is a simple yet expressive clustering objective—\emph{Augmented Structural Entropy}, which is formalized below
\begin{align}
\min\nolimits_{\mathbf{\Theta}}  \ \mathcal{H}^{\mathcal{T}_\text{net}}(G^\gamma;\mathbf{Z};\mathbf{\Theta}),
\label{ASIL}
\end{align}
where $G^\gamma$ is a fusion of the original graph and its representational view, the parameterized virtual graph  (Sec. \ref{sec:virtualGraph}) that is constructed from node representations derived by Lorentz convolution in practice. Thus, the adjacency matrix is given with a balancing factor $\gamma$ as follows,
\begin{align}
\mathbf{A}^\gamma=(1-\gamma)\mathbf{A}+\gamma\tilde{\mathbf{A}}, \quad \gamma > 0.
\label{Gfuse}
\end{align}
Note that the objective of Eq. (\ref{ASIL}) simultaneously learns an enhanced partitioning tree with a virtual graph, in which optimizing the virtual graph refines node representation via implicitly Lorentz tree contrastive learning (Sec. \ref{sec:TCL}).

Interestingly, this enhanced partitioning tree enjoys superior clustering capability to the deep tree of  $\min\nolimits_{\mathbf{\Theta}}  \ \mathcal{H}^{\mathcal{T}_\text{net}}(G;\mathbf{Z};\mathbf{\Theta})$,
which is proved in light of graph conductance.

\begin{mymath}
\begin{theorem}[\textbf{Graph Conductance Improvement with Graph Fusion}]\label{thm:fused_conductance}
   Given a graph $G=(\mathcal{V}, \mathcal{E})$ and a virtual graph $\tilde{G}$ as in Theorem \ref{thm:se_tcl}, the fused graph $G^\gamma$ is derived as in Eq. (\ref{Gfuse}).
    For any nontrivial subset $\mathcal{S} \subset \mathcal{V}$ with $0 < \mathrm{vol}_A(\mathcal{S}) < \mathrm{vol}_A(V)$, 
    denote the conductance of $\mathcal{S}$ under a graph $G$ as
    \[
        \phi_G(\mathcal{S}) = \frac{\mathrm{cut}_G(\mathcal{S}, \mathcal{V} \setminus \mathcal{S})}{\min\{\mathrm{vol}_G(\mathcal{S}),\, \mathrm{vol}_G(\mathcal{V} \setminus \mathcal{S})\}},
    \]
    where $\mathrm{cut}_G(\mathcal{S}, \mathcal{P}) = \sum_{i \in \mathcal{S}, j \in \mathcal{P}} w_{ij}$ and $\mathrm{vol}_G(\mathcal{S}) = \sum_{i \in \mathcal{S}} \sum_j w_{ij}$.
    Suppose that for every $\mathcal{S} \subset \mathcal{V}$,
    \begin{equation}\label{eq:cond_assump}
        \phi_{\tilde{G}}(\mathcal{S}) \leq \phi_G(\mathcal{S}).
    \end{equation}
    Then, the conductance of the fused graph satisfies
    \[
        \phi_{G^\alpha}(\mathcal{S}) \leq \phi_G(\mathcal{S}), \quad \forall \mathcal{S} \subset V.
    \]
    Moreover, if there exists a subset $\mathcal{S}^\star$ such that $\phi_{\tilde{G}}(\mathcal{S}^\star) < \phi_G(\mathcal{S}^\star)$, 
    then $\phi_{G^\alpha}(\mathcal{S}^\star) < \phi_G(\mathcal{S}^\star)$ for all $\alpha \in (0,1]$.
\end{theorem}
\begin{proof}
    Refer to Appendix II.I, Theorem VII.2.
  \end{proof}
\end{mymath}

\begin{algorithm}[t]
    \caption{\textbf{ Training  \texttt{ASIL}}}
    \label{alg:lsenet}
    \renewcommand{\algorithmicrequire}{\textbf{Input:}}
    \renewcommand{\algorithmicensure}{\textbf{Output:}}
    \begin{algorithmic}[1]
        \REQUIRE A weighted graph $G=(\mathcal{V}, \mathcal{E}, \mathbf{X})$ and the height of  deep partitioning tree $H$, fusion parameter $\gamma \in (0,1]$, nearset neighbors number $k$.
        %, Training iterations $L$
         \ENSURE The partitioning tree $\mathcal{T}$, and tree node embeddings $\{\mathbf{Z}_\gamma^h\}_{h=1,...,H}$ at all the levels.
        %\FOR{$epoch=1$ to $L$}
          \WHILE{not converged}
        \STATE Obtain initial  embeddings $\mathbf{Z}^H = \operatorname{LConv}(\mathbf{X}, \mathbf{A})$;
        \STATE Infer virtual graph  $\tilde{\mathbf{A}}_{ij} = \exp(-d_{\mathbb{L}}(L(\boldsymbol{z}_i), L(\boldsymbol{z}_j))/\tau)$ with  Lorentz boost $L$ in Eq.~\ref{eq. lorentz boost};
          \STATE Sparsify the virtual graph with $k$-nearest neighbors;
        \STATE Conduct graph fusion by $\mathbf{A}^\gamma = \mathbf{A} + \gamma \tilde{\mathbf{A}}$;
        \STATE Obtain leaf node embeddings $\mathbf{Z}_\gamma^H = \operatorname{LConv}(\mathbf{X}, \mathbf{A}^\gamma)$;
        \STATE Compute the level-wise assignment and parent node embeddings $\{\mathbf{Z}_\gamma^h\}_{h=1,...,H-1}$ recursively with  Eqs. \ref{assignment}, \ref{parentNode} and \ref{structure};
        \STATE Compute the overall objective $\mathcal{H}^{\mathcal{T}_\text{net}}(G^\gamma;\mathbf{Z};\mathbf{\Theta})$;
        \STATE Update the parameters with gradient-descent optimizer;
        \ENDWHILE
        \STATE Construct the optimal partitioning tree of  root node $\lambda$ with a queue $\mathcal{Q}$;
        \WHILE{$\mathcal{Q}$ is not empty}
            \STATE Get first item $\alpha$ in $\mathcal{Q}$;
            \STATE Let $h=\alpha.h+1$ and search subsets $P$ from $\mathbf{S}^h$ defined with level-wise assignments in Eq. \ref{eq.s};
            \STATE Create nodes from $P$ and put into the queue $\mathcal{Q}$;
            \STATE Add theses nodes into $\alpha$'s children list;
        \ENDWHILE
        \STATE Return the partitioning tree $\mathcal{T}:=\lambda$.
    \end{algorithmic}
\end{algorithm}

\vspace{-0.2in}
\subsection{Algorithm: ASIL for Graph Clustering}
\label{sec:objective}

The overall procedure of \texttt{ASIL} is summarized in Algorithm~\ref{alg:lsenet}. 
The first phase (Line 1-10) is to learn hyperbolic partitioning tree representation in a bottom-up manner.
Specifically, in Line 2-4, a parameterized virtual graph first is constructed from node representations with a neural Lorentz boost, implicitly capturing tree contrastive learning.
In Line 2-4, we forward a fused graph into \texttt{LSEnet}, recursively inducing parent node embeddings.
In Line 8-9, the partitioning tree is optimized by a novel objective of augmented structural entropy.
The second phase (Line 11-18) is to recover the partitioning tree structure through a top-down decoding process, which prunes empty leaves and resolves redundant chains in accordance with Theorem~\ref{theorem.flex}. 
Clustering structures are naturally uncovered in the partitioning tree, whose procedure are provided in Algorithm 2 given in Appendix III.

\emph{\textbf{Computational Complexity.}}
First, we analyze the complexity of computing differentiable structural information.
Thanks to the node-wise formulation in Eq.~\eqref{eq.new_ksi}, this computation no longer requires enumerating all edges at each level. Instead, it operates solely on the coarsened graphs $\{G^h\}_{h=1}^H$, whose node counts $\{N_h\}$ satisfy $N_h \ll N_{h-1}$ and $N_H = N$.
Second, given that the induced graph $G^{h-1}$ is  sparse, $w^h_{kl}$ is computed via sparse matrix multiplication $\mathbf{W}^h = (\mathbf{S}^h)^\top \mathbf{A} \mathbf{S}^h$, yielding the computational cost of $\mathcal{O}(|\mathcal{E}|)$ at leaf level, where $\mathcal{E}$ is the edge set of graph $G$.
Third, we leverage $k$-nearest neighbors to construct the virtual adjacency matrix, which costs $\mathcal{O}(k|\mathcal{V}|)$, where $\mathcal{V}$ is the node set. 
Finally, the objective of augmented structural entropy regarding $\mathbf{A}^\gamma = (1 - \gamma) \mathbf{A} + \gamma \tilde{\mathbf{A}}$ costs  $\mathcal{O}(|\mathcal{E}| + k|\mathcal{V}|)$ in total.
That is, the proposed \texttt{ASIL}  discovers node clusters with a more preferable linear complexity in graph size, unknowing the number of clusters, while previous methods (e.g., MVGRL~\cite{HassaniA20}) typically undergo a clustering procedure costing $\mathcal{O}(|\mathcal{N}|^2)$.

\emph{In summary, we present a hyperbolic deep model  \texttt{ASIL} built on the established theory of differential structural entropy that allows for effective and efficient debiased clustering in linear complexity, without requiring the number of clusters.}

\begin{table*}[t]
  \centering
  \caption{Clustering results on Cora, Citerseer, AMAP, and Computer, Pubmed datasets in terms of NMI and ARI (\%). OOM denotes Out-of-Memory on our hardware. The best results are highlighted in \textbf{boldface}, and runner-ups are \underline{underlined}.}
    \vspace{-0.05in}
  \label{tab-main-nmi-ari}
  \renewcommand{\arraystretch}{1.08}
    \resizebox{1\linewidth}{!}{
     \begin{tabular}{c|cc|cc|cc|cc|cc}
             \hline
         & \multicolumn{2}{c|}{\textbf{Cora}}  & \multicolumn{2}{c|}{\textbf{Citeseer}}  & \multicolumn{2}{c|}{\textbf{AMAP}} & \multicolumn{2}{c|}{\textbf{Computer}} & \multicolumn{2}{c}{\textbf{Pubmed}} \\
        & NMI  & ARI  & NMI  & ARI  & NMI  & ARI  & NMI  & ARI & NMI& ARI \\
        \hline
            DEC
       & 23.54\scriptsize{$\pm$1.13}  & 15.13\scriptsize{$\pm$0.72}  & 28.34\scriptsize{$\pm$0.42}  & 28.12\scriptsize{$\pm$0.24}
       & 37.35\scriptsize{$\pm$0.84}  & 18.29\scriptsize{$\pm$0.64}  & 38.56\scriptsize{$\pm$0.24}  & 34.76\scriptsize{$\pm$0.35}
       & 22.44\scriptsize{$\pm$0.14} & 19.55\scriptsize{$\pm$0.13}\\
      \hline
       VGAE
       & 43.40\scriptsize{$\pm$1.62}  & 37.50\scriptsize{$\pm$2.13}  & 32.70\scriptsize{$\pm$0.30}  & 33.10\scriptsize{$\pm$0.55}
       & 66.01\scriptsize{$\pm$0.84}  & 56.24\scriptsize{$\pm$0.22}  & 37.62\scriptsize{$\pm$0.24}  & 22.16\scriptsize{$\pm$0.16}
       & 30.61\scriptsize{$\pm$1.71} & 30.15\scriptsize{$\pm$1.23}\\
       ARGA 
       & 48.10\scriptsize{$\pm$0.45}  & 44.10\scriptsize{$\pm$0.28}  & 35.10\scriptsize{$\pm$0.58}  & 34.60\scriptsize{$\pm$0.48}
       & 58.36\scriptsize{$\pm$1.02}  & 44.18\scriptsize{$\pm$0.85}  & 37.21\scriptsize{$\pm$0.58}  & 26.28\scriptsize{$\pm$1.38}
       & 24.80\scriptsize{$\pm$0.17} & 24.35\scriptsize{$\pm$0.17}\\
       MVGRL
       & 62.91\scriptsize{$\pm$0.42}  & 56.96\scriptsize{$\pm$0.74}  & 46.96\scriptsize{$\pm$0.25}  & 44.97\scriptsize{$\pm$0.57}
       & 36.89\scriptsize{$\pm$1.31}  & 18.77\scriptsize{$\pm$1.54}  & 10.12\scriptsize{$\pm$2.21}  & $\ \ $5.53\scriptsize{$\pm$1.78}
       & 31.59\scriptsize{$\pm$1.45} & 29.42\scriptsize{$\pm$1.06}\\
      Sublime
       & 54.20\scriptsize{$\pm$0.28}  & 50.32\scriptsize{$\pm$0.32}  & 44.10\scriptsize{$\pm$0.27}  & 43.91\scriptsize{$\pm$0.35}
       & $\ \ $6.37\scriptsize{$\pm$0.54}  &$\ \ $5.36\scriptsize{$\pm$0.24}  & 39.16\scriptsize{$\pm$1.82}  & 24.15\scriptsize{$\pm$0.63}
       & 32.01\scriptsize{$\pm$1.24} & 30.36\scriptsize{$\pm$0.85}\\
       ProGCL
       & 41.02\scriptsize{$\pm$1.64}  & 30.71\scriptsize{$\pm$1.38}  & 39.59\scriptsize{$\pm$1.58}  & 36.16\scriptsize{$\pm$2.21}
       & 39.56\scriptsize{$\pm$0.25}  & 34.18\scriptsize{$\pm$0.58}  & 35.50\scriptsize{$\pm$2.06}  & 26.08\scriptsize{$\pm$1.91} 
       & 30.24\scriptsize{$\pm$1.21} & 28.98\scriptsize{$\pm$1.06}\\
      \hline
      GRACE
       & 57.30\scriptsize{$\pm$0.86}  & 52.70\scriptsize{$\pm$1.20}  & 39.90\scriptsize{$\pm$2.26}  & 37.70\scriptsize{$\pm$1.35}
       & 53.46\scriptsize{$\pm$1.32}  & 42.74\scriptsize{$\pm$1.57}  & 47.90\scriptsize{$\pm$1.65}  & 36.40\scriptsize{$\pm$1.56}
       & 30.08\scriptsize{$\pm$0.24} & 27.60\scriptsize{$\pm$1.24}\\
      MCGC
       & 44.90\scriptsize{$\pm$1.56}  & 37.80\scriptsize{$\pm$1.24}  & 40.14\scriptsize{$\pm$1.44}  & 38.00\scriptsize{$\pm$0.85}
       & 61.54\scriptsize{$\pm$0.29}  & 52.10\scriptsize{$\pm$0.27}  & 53.17\scriptsize{$\pm$1.29}  & 39.02\scriptsize{$\pm$0.53}
       & 33.39\scriptsize{$\pm$0.02} & 29.25\scriptsize{$\pm$0.01}\\
      DCRN
       & 48.86\scriptsize{$\pm$0.85}  & 43.79\scriptsize{$\pm$0.48}  & 45.86\scriptsize{$\pm$0.35}  & 47.64\scriptsize{$\pm$0.30}
       & 73.70\scriptsize{$\pm$1.24}  & 63.69\scriptsize{$\pm$0.84}  & OOM                                       & OOM      
       & 32.20\scriptsize{$\pm$0.08} & 31.41\scriptsize{$\pm$0.12}\\
       FT-VGAE
        & 61.03\scriptsize{$\pm$0.52}   & 58.22\scriptsize{$\pm$1.27}   & 44.50\scriptsize{$\pm$0.13}  & 46.71\scriptsize{$\pm$0.75} 
        & 69.76\scriptsize{$\pm$1.06}   & 59.30\scriptsize{$\pm$0.81}   & 51.36\scriptsize{$\pm$0.92}  & 40.07\scriptsize{$\pm$2.13}
        & 33.05\scriptsize{$\pm$0.24} & 35.50\scriptsize{$\pm$0.12}
        \\
       gCooL
       & 58.33\scriptsize{$\pm$0.24}   & 56.87\scriptsize{$\pm$1.03}  & 47.29\scriptsize{$\pm$0.10}  & 46.78\scriptsize{$\pm$1.51} 
       & 63.21\scriptsize{$\pm$0.09}   & 52.40\scriptsize{$\pm$0.11}  & 47.42\scriptsize{$\pm$1.76}  & 27.71\scriptsize{$\pm$2.28}
       & 30.82\scriptsize{$\pm$0.15} & 31.48\scriptsize{$\pm$0.34}\\
       S$^3$GC
       & 58.90\scriptsize{$\pm$1.81}  & 54.40\scriptsize{$\pm$2.52}  & 44.12\scriptsize{$\pm$0.90}  & 44.80\scriptsize{$\pm$0.65}
       & 59.78\scriptsize{$\pm$0.45}  & 56.13\scriptsize{$\pm$0.58}  & 54.80\scriptsize{$\pm$1.22}  & 29.93\scriptsize{$\pm$0.22}
       & 33.32\scriptsize{$\pm$0.34} & 34.50\scriptsize{$\pm$0.57}\\
       Congregate
       & 63.16\scriptsize{$\pm$0.71}  & 59.27\scriptsize{$\pm$1.23}  & 50.92\scriptsize{$\pm$1.58}  & 47.59\scriptsize{$\pm$1.60} 
       & 70.99\scriptsize{$\pm$0.67}  & 60.55\scriptsize{$\pm$1.36}  & 46.03\scriptsize{$\pm$0.47}  & 38.57\scriptsize{$\pm$1.05} &
       35.03\scriptsize{$\pm$0.23}  & 33.42\scriptsize{$\pm$0.47}\\
      DinkNet
       & 62.28\scriptsize{$\pm$0.24}  & 61.61\scriptsize{$\pm$0.90}  & 45.87\scriptsize{$\pm$0.24}  & 46.96\scriptsize{$\pm$0.30}
       & 74.36\scriptsize{$\pm$1.24}  & \underline{68.40}\scriptsize{$\pm$1.37}  & 39.54\scriptsize{$\pm$1.52} & 33.87\scriptsize{$\pm$0.12}
       & 36.39\scriptsize{$\pm$0.34} & 35.77\scriptsize{$\pm$0.13}\\
      GC-Flow
       & 62.15\scriptsize{$\pm$1.35}  & 63.14\scriptsize{$\pm$0.80}  & 40.50\scriptsize{$\pm$1.32}  & 42.62\scriptsize{$\pm$1.44}
       & 36.45\scriptsize{$\pm$1.21}  & 37.24\scriptsize{$\pm$1.26}  & 41.10\scriptsize{$\pm$1.06} & 35.60\scriptsize{$\pm$1.82 }
       & \underline{36.82}\scriptsize{$\pm$0.28} & \underline{37.31}\scriptsize{$\pm$0.04}\\
       GCSEE
       & 53.02\scriptsize{$\pm$0.41} & 51.22\scriptsize{$\pm$0.88}
       & 35.66\scriptsize{$\pm$0.42} & 36.29\scriptsize{$\pm$0.37}
       & 64.15\scriptsize{$\pm$0.68} & 56.76\scriptsize{$\pm$1.24}
       & 39.03\scriptsize{$\pm$1.58} & 35.24\scriptsize{$\pm$1.56}
       & 26.88\scriptsize{$\pm$0.21} & 26.15\scriptsize{$\pm$0.24}\\
       FastDGC
       & 53.03\scriptsize{$\pm$0.03} & 51.25\scriptsize{$\pm$0.05}
       & 45.85\scriptsize{$\pm$0.08} & 44.12\scriptsize{$\pm$0.05}
       & 69.76\scriptsize{$\pm$0.03} & 65.71\scriptsize{$\pm$0.04}
       & 45.26\scriptsize{$\pm$0.52} & 40.23\scriptsize{$\pm$0.18}
       & 25.85\scriptsize{$\pm$0.86} & 25.08\scriptsize{$\pm$0.73}\\
       MAGI
       & 59.72\scriptsize{$\pm$1.22} & 57.31\scriptsize{$\pm$1.31}
       & 45.28\scriptsize{$\pm$0.24} & 46.86\scriptsize{$\pm$0.47}
       & 71.62\scriptsize{$\pm$0.11} & 61.55\scriptsize{$\pm$0.24}
       & 54.22\scriptsize{$\pm$1.88} & \textbf{46.27}\scriptsize{$\pm$1.64}
       & 33.28\scriptsize{$\pm$0.04} & 31.29\scriptsize{$\pm$0.08}\\
    \hline
       RGC
       & 57.60\scriptsize{$\pm$1.36}  & 50.46\scriptsize{$\pm$1.72}  & 45.70\scriptsize{$\pm$0.29}  & 45.47\scriptsize{$\pm$0.43}
       & 47.65\scriptsize{$\pm$0.91}  & 42.65\scriptsize{$\pm$1.53}  & 46.24\scriptsize{$\pm$1.05}  & 36.12\scriptsize{$\pm$0.30} 
       & 28.01\scriptsize{$\pm$1.85} & 25.56\scriptsize{$\pm$1.91}\\
       DGCluster
       & 61.47\scriptsize{$\pm$0.34} & 44.75\scriptsize{$\pm$0.28}
       & 43.75\scriptsize{$\pm$0.05} & 31.55\scriptsize{$\pm$0.09}
       & \textbf{77.12}\scriptsize{$\pm$0.25} & 67.59\scriptsize{$\pm$0.35}
       & 50.10\scriptsize{$\pm$1.87} & 38.92\scriptsize{$\pm$1.27}
       & 33.73\scriptsize{$\pm$0.68} & 21.84\scriptsize{$\pm$0.71}\\
       \texttt{LSEnet}
       & \underline{63.97}\scriptsize{$\pm$0.67}  & \underline{63.35}\scriptsize{$\pm$0.56}   & \underline{52.26}\scriptsize{$\pm$1.09}  &\underline{48.01}\scriptsize{$\pm$1.25}  
       & 71.72\scriptsize{$\pm$1.30}  & 65.08\scriptsize{$\pm$0.73}  & \underline{55.03}\scriptsize{$\pm$0.79} & 42.15\scriptsize{$\pm$1.02} &34.25\scriptsize{$\pm$0.24} & 35.16\scriptsize{$\pm$0.30}\\
      \textbf{\texttt{ASIL}}
     & \textbf{65.78}\scriptsize{$\pm$0.84}  & \textbf{64.72}\scriptsize{$\pm$0.67}   &\textbf{55.01}\scriptsize{$\pm$1.36}  &\textbf{51.77}\scriptsize{$\pm$1.23}  
       & \underline{75.81}\scriptsize{$\pm$0.33}  & \textbf{69.52}\scriptsize{$\pm$0.65}  & \textbf{56.31}\scriptsize{$\pm$0.55} & \underline{45.77}\scriptsize{$\pm$0.35} &
       \textbf{38.13}\scriptsize{$\pm$0.41}  & \textbf{39.74}\scriptsize{$\pm$0.59}\\
      \hline
      \textbf{Avg. Gain}
      & 11.55 & 15.16
      & 12.42 & 10.25
      & 18.20 & 20.52
      & 12.84 & 12.82
      & 6.95  & 10.02 \\
      \hline
    \end{tabular}
}
  \vspace{-0.1in}
\end{table*}

    \vspace{-0.05in}
\section{Experiments}
    \vspace{-0.05in}
In the experiment, we compare with $20$ strong baselines on $5$ commonly used datasets, aiming to answer the research questions as follows:
\textbf{RQ1} (Main Results): How does the proposed \texttt{ASIL} perform in terms of both effectiveness and efficiency, especially compared to vallina \texttt{LSEnet}?
\textbf{RQ2}: How does \texttt{ASIL} outperform the classic structural entropy~\cite{li2016structural} and our previous model~\cite{icml24sun}?
\textbf{RQ3} (Ablation Study): How does each proposed module contribute to the success of \texttt{ASIL}, including the choice of hyperbolic space, Neural Lorentz Boost Construction and Implicit Tree Contrastive Learning?
\textbf{RQ4} (Hyperparameter Sensitivity): How do the hyperparameters affect the performance in each experimental datasets?
\textbf{RQ5}: How do the imbalance clusters impact the clustering results? 
Besides, we examine the embedding expressiveness of \texttt{ASIL}, and include a visualization of the clusters learnt from \texttt{ASIL}.
Additional results and further details are provided in Appendix VII.
Codes and data are publicly available at \url{https://github.com/RiemannGraph/DSE_clustering}.

    \vspace{-0.1in}
\subsection{Experimental Setups}

\subsubsection{\textbf{Datasets \& Baselines}}

The experiments are conducted on a diversity of datasets including  Cora, Citeseer, Pubmed \cite{sen2008collective}, Amozon Photo, and a larger Computer dataset \cite{shchur2018pitfalls}. 
In this work, we focus on deep graph clustering and compare with $20$ strong deep baselines categorized into two groups.
The first group includes  a classic  DEC \cite{xie2016unsupervised} and $14$ state-of-the-art deep graph clustering models:
GRACE \cite{zhu2020deep}, MCGC \cite{pan2021multi}, 
 DCRN \cite{liu2021deep}, FT-VGAE \cite{DBLP:conf/ijcai/MrabahBK22}, gCooL \cite{DBLP:conf/www/LiJT22}, 
 S$^3$GC \cite{devvrit2022s3gc}, Congregate \cite{sun2023contrastive}, DinkNet \cite{liu2023dink}, 
 GC-Flow \cite{wang2023gc}, RGC \cite{liu2023reinforcement},
GCSEE \cite{ding2023graph},
FastDGC \cite{ding2024towards},
MAGI \cite{liu2024revisiting}
and DGCluster \cite{bhowmick2024dgcluster}.
The second group consists of  $5$ self-supervised graph models: 
VGAE \cite{kipf2016variational}, ARGA \cite{pan2018adversarially}, MVGRL \cite{HassaniA20}, 
Sublime \cite{liu2022towards} and ProGCL \cite{xia2021progcl}.
The embedding, learned from the self-supervised models, is then followed up with  $K$-Means algorithm to generate node clusters.
%In addition, the clustering results of $K$-Means over node attributes is listed  as a reference. 
Among the baselines, Congregate is the only Riemannian baseline that considers the curvature heterogeneity underlying the graph itself, while we explore the hyperbolic geometry of the partitioning tree.
We specify that previous deep clustering models cannot be instantiated without the cluster number $K$, except RGC and DGCluster.
% Concretely, DGCluster generates node encodings without $K$ by a carefully designed modularity maximization;
% RGC automates the training-testing process to search the optimal $K$ via reinforcement learning.
% Thus, both of them have different focus from ours, 
Distinguishing from the previous models, we are dedicated to uncovering the cluster structure in hyperbolic space from the information theory perspective.
(Detailed dataset and baseline descriptions are given in Appendix VI.)

%, but we can work with any Nh under slight constraint. For Nh not ideal, results is refined by hyperbolic distance.

\begin{table*}[t]
  \centering
  \caption{Accuracy results on Cora, Citeseer, AMAP, Computer, and Pubmed datasets in terms of percentage (\%). The best results are highlighted in \textbf{boldface}, and runner-ups are \underline{underlined}.}
    \vspace{-0.05in}
  \label{tab-main-acc}
  \renewcommand{\arraystretch}{1.08}
  \resizebox{1\linewidth}{!}{
    \begin{tabular}{c|cccccccccccc|c}
      \hline
      & VGAE & ARGA & MVGRL & Sublime & GRACE & DCRN & gCooL & S$^3$GC & DinkNet & GC-Flow & DGCluster & \textbf{ASIL} & \textbf{Avg. Gain}\\
      \hline
      \textbf{Cora} & 64.72 & 68.40 & \underline{80.03} & 71.30 & 73.87 & 66.91 & 77.03 & 74.27 & 78.11 & 78.92 & 76.45 & \textbf{82.14} & 8.50 \\
      \textbf{Citeseer} & 61.00 & 60.72 & 61.23 & 68.53 & 63.12 & 70.86 & \underline{71.35} & 68.80 & 70.36 & 70.06 & 70.14 & \textbf{72.73} & 5.81 \\
      \textbf{AMAP}   & 74.26 & 69.28 & 41.07 & 47.22 & 67.66 & 77.94 & 68.52 & 75.15 & \underline{79.23} & 72.54 & 78.92 & \textbf{79.31} & 10.97 \\
      \textbf{Computer}   & 42.44 & 45.67 & 29.53 & 45.52 & \underline{46.62} & 30.42 & 44.50 & 39.03 & 43.52 & 43.20 & 43.65 & \textbf{49.03} & 7.75 \\
      \textbf{Pubmed} & 68.31 & 64.82 & 67.01 & 66.62 & 63.72 & 69.63 & 68.24 & 71.30 & 69.42 & \underline{73.42} & 66.39 & \textbf{74.47} & 6.39 \\
      \hline
    \end{tabular}
  }
    \vspace{-0.1in}
\end{table*}

\subsubsection{\textbf{Evaluation Metrics}}
%Note that NMI is particularly robust in handling imbalanced distributions, which complements the design principles of our model.

We evaluate the node clustering results with three popular metrics: Normalized Mutual Information (NMI) \cite{HassaniA20}, Adjusted Rand Index (ARI) \cite{li2022graph} and Accuracy (ACC) \cite{HassaniA20,DBLP:conf/www/LiJT22}.
Concretely, ACC quantifies the accuracy of predicted cluster assignment. 
NMI evaluates the amount of information shared between predicted and true distribution.
ARI considers the correct and wrong assignment.
In the experiment, each case undergoes $10$ independent runs to ensure statistical robustness and fair comparison. We report the mean value along with the standard deviation.
All experiments are done on a server of 
an NVIDIA GeForce RTX 4090D GPU (24GB VRAM) utilizing CUDA 11.8, an AMD EPYC 9754 128-Core Processor (18 vCPUs allocated), 60GB of RAM.

\subsubsection{\textbf{Model Configuration \& Reproducibility}}

We instantiate the proposed \texttt{ASIL} in the Lorentz model of hyperbolic space with a standard curvature of $\kappa=-1$.
%, and the hyperbolic spaces with different curvatures are mathematically equivalent in fact.
The model configuration by default is specified as follows.
We consider the partitioning tree of height $2$ by default and hyperparameter $N_1$ of network architecture is set as $10$.
The augmentation ratio $\gamma$ is $0.01$, the $k$ nearest neighbors for sparsifying $\tilde{\mathbf{A}}$ is set to $8$.
We stack the Lorentz convolution once in \texttt{LSEnet}, 
The parameters are optimized by Adam, the further details and the hyperparameter of the baselines are provided in implementation note of Appendix VI.

\subsection{Main Results of Node Clustering}

\subsubsection{\textbf{Effectiveness}}

We report the clustering results in terms of NMI and ARI on the five real datasets in Table \ref{tab-main-nmi-ari}, and ACC in Table \ref{tab-main-acc}.
Deep clustering models require cluster number $K$ in training phase, except DGCluster, RGC and our \texttt{ASIL}.
For clustering models without $K$, 
DGCluster conducts self-supervised representation learning with modularity \cite{bhowmick2024dgcluster}, 
and RGC searches the optimal $K$ with reinforcement learning \cite{liu2023reinforcement}. 
%For self-supervised models,  we apply K-Means over node embeddings to obtain node clusters.
For Riemannian models, i.e., Congregate and \texttt{ASIL}, 
we project Euclidean attributes onto the respective manifold, and the projection is given by exponential map with the reference point of the origin.
DCRN does not scale well to the larger Computer dataset, and runs into Out-Of-Memory (OOM) error on our hardware.
In Table \ref{tab-main-nmi-ari}, the proposed \texttt{ASIL} has the best clustering results except two cases,  NMI on AMAP and ARI on Computer.
Note that, \texttt{ASIL} outperforms $20$ recent deep clustering models by an average of  $+12.42\%$ NMI on Citeseer, and generally achieves the best ACC in Table \ref{tab-main-acc}.
Also, we observe that \texttt{ASIL} shows more consistent performance over different datasets, 
e.g., DGCluster obtain the best NMI on AMAP but is unsatisfactory on Computer.

\subsubsection{\textbf{Efficiency}}

Table \ref{tab-time complexity} collects the theoretical time complexity and empirical running time on Citeseer dataset. 
The running time is given in seconds, and we are presented as the fastest deep model in the table.
As for theoretical complexity, $N$ and $|E|$ denote the number of nodes and edges, respectively. $d$ is the hidden dimension of models.
$B$ is the batch size of DinkNet.
$S$ is the average degree of nodes and thereby the complexity of S$^3$GC is related to the number of edges.
In DGCluster, $P$ is a dimensionality of auxiliary matrix, $P \ll N$.
In \texttt{ASIL}, $k$ is the number of nearest neighbors for sparsifying $\tilde{A}$,
and it yields the linear complexity thanks to the reformulation of differential structural information and adjacency sparsifying.
It is noteworthy to mention that, 
compared to recent deep clustering model such as S$^3$GC, 
we achieve superior clustering effectiveness while removing the prior knowledge of cluster number without computational overhead.
The following parts will demonstrate how the differential formulation, hyperbolic space and neural Loretnz boost contribute to the success of the proposed model.

% We can find that as the dimension grows, the variance increases as well, since constructing a partitioning tree is an NP-hard problem that requires exponential search space when the height of the tree increases, the gradient descent methods can not guarantee the optimal solution.

\begin{table*}[t]
  \centering
  \caption{Comparison to \texttt{LSEnet} and CSE on Cora, Citerseer, Computer, AMAP and PubMed datasets in terms of NMI and ARI (\%).}
    \vspace{-0.05in}
  \label{tab-se-nmi-ari}
    \resizebox{1\linewidth}{!}{
     \begin{tabular}{ c |cc|cc|cc|cc|cc}
      \hline
         & \multicolumn{2}{c|}{\textbf{Cora}}  & \multicolumn{2}{c|}{\textbf{Citeseer}}  & \multicolumn{2}{c|}{\textbf{Computer}} & \multicolumn{2}{c|}{\textbf{AMAP}} &\multicolumn{2}{c}{\textbf{PubMed}} \\
        & NMI  & ARI  & NMI  & ARI  & NMI  & ARI  & NMI  & ARI & NMI  & ARI \\
      \hline
       CSE
       & 60.82\scriptsize{$\pm$0.30}  & 59.36\scriptsize{$\pm$0.17}  & 47.12\scriptsize{$\pm$1.01}  & 45.67\scriptsize{$\pm$0.67}
       & OOM & OOM  & OOM & OOM & OOM & OOM \\
       LSEnet
       & 63.97\scriptsize{$\pm$0.67}  & 63.35\scriptsize{$\pm$0.56}  & 52.26\scriptsize{$\pm$1.09}  & 48.01\scriptsize{$\pm$1.25} 
       & 71.12\scriptsize{$\pm$1.30}  & 65.08\scriptsize{$\pm$0.73}  & 55.03\scriptsize{$\pm$0.79}  & 42.15\scriptsize{$\pm$1.02} & 34.25\scriptsize{$\pm$0.24} & 35.16\scriptsize{$\pm$0.30} \\
       \textbf{\texttt{ASIL}}
       & 65.78\scriptsize{$\pm$0.84}  & 64.72\scriptsize{$\pm$0.67}  & 55.01\scriptsize{$\pm$1.36}  & 51.77\scriptsize{$\pm$1.23}
       & 75.81\scriptsize{$\pm$0.33}  & 69.52\scriptsize{$\pm$0.65}  & 56.31\scriptsize{$\pm$0.55} & 45.77\scriptsize{$\pm$0.35} & 38.13\scriptsize{$\pm$0.41}  & 39.74\scriptsize{$\pm$0.59} \\
      \hline
      \textbf{Average Gain}
      & \textbf{3.39} & \textbf{3.37} &\textbf{5.32} &\textbf{4.93}
      & \textbf{4.69} & \textbf{4.44} &\textbf{1.28} &\textbf{3.62} &\textbf{3.88} &\textbf{4.58}\\
      \hline 
    \end{tabular}
}
  \vspace{-0.2in}
\end{table*}

\begin{table}[t]
  \centering
  \caption{Comparison of time complexity and time cost for baselines.}
    \vspace{-0.05in}
  \label{tab-time complexity}
  \renewcommand{\arraystretch}{1.08}
  \setlength{\tabcolsep}{6pt} 
  \begin{tabular}{c|c|c}
    \hline
    \textbf{Method} & \textbf{Time Complexity}& \textbf{Running Time (s)} \\
    \hline
    DGI & $\mathcal{O}(|E|d + N d^2)$ & 100.7 \\
    MVGRL & $\mathcal{O}(N^2d + N d^2)$ & 829.8 \\
    S$^3$GC & $\mathcal{O}(N S d^2)$ & 1247.2  \\
    Dink-Net & $\mathcal{O}(BKd + K^2d + Bd)$ & 217.7 \\
    DGCluster & $\mathcal{O}(K^2Nd + P^2Nd^2)$ & 79.3 \\
    \textbf{ASIL} (Ours) & $\mathcal{O}(|E|d + kNd^2)$ & 22.1 \\
    \hline
  \end{tabular}
    \vspace{-0.1in}
\end{table}

\subsection{Comparison with Classic Structural Entropy \cite{li2016structural}} \label{CompareCSE}

\begin{figure}[t]
\centering
\begin{minipage}{0.48\textwidth}
  \centering
  \includegraphics[height=0.43\textwidth]{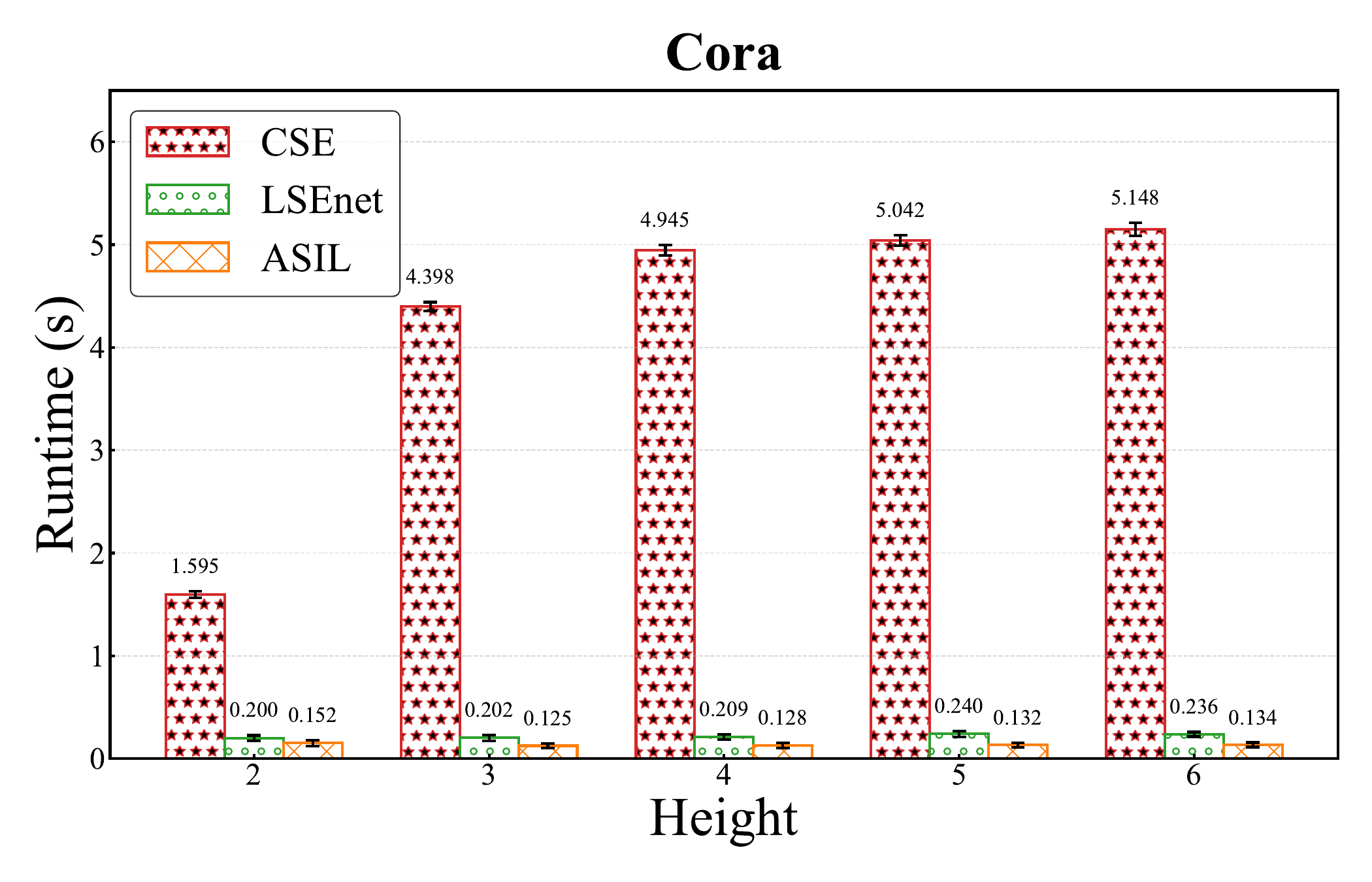}
  \includegraphics[height=0.43\textwidth]{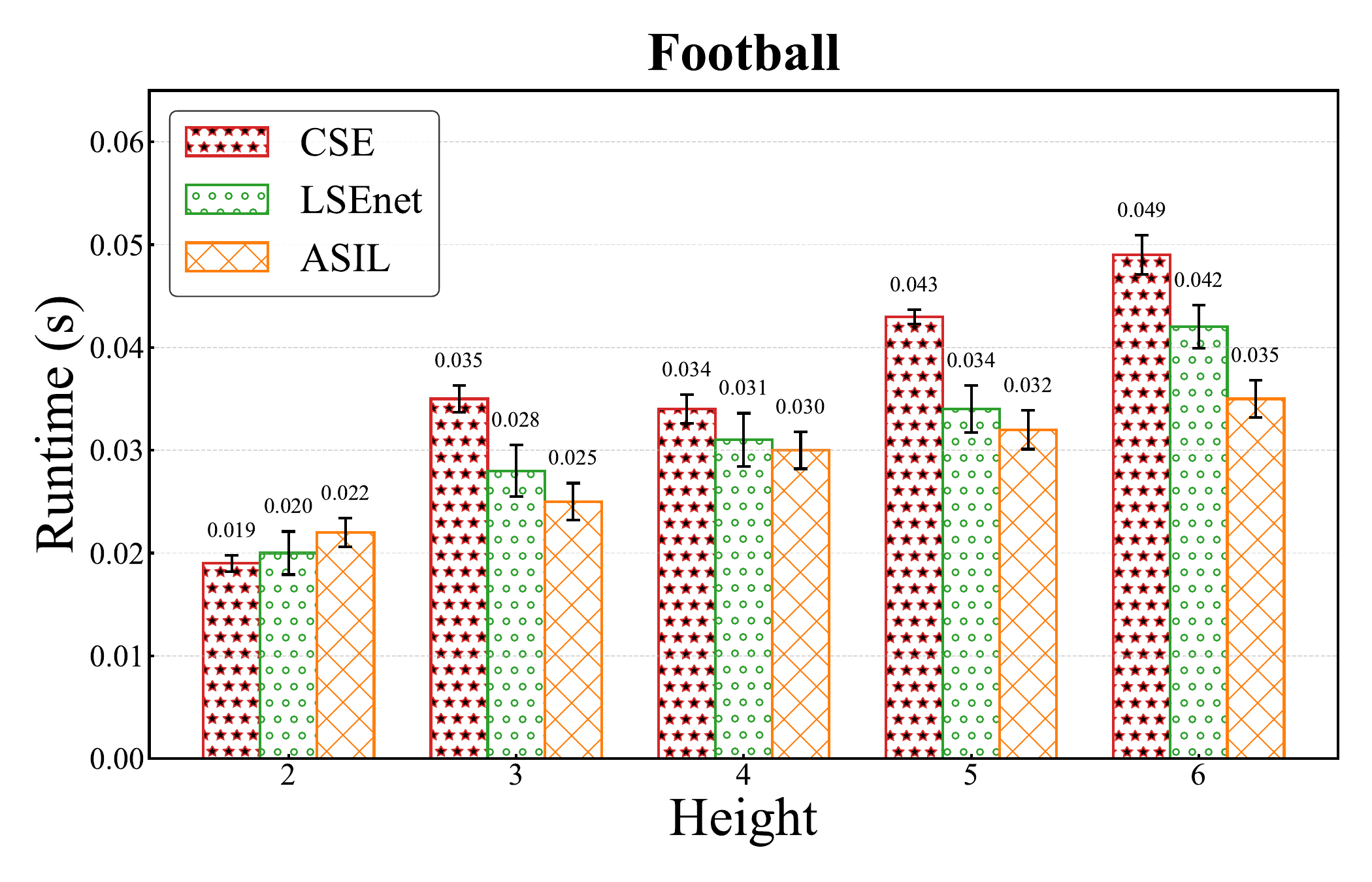}
\end{minipage}
  \vspace{-0.05in}
\caption{Running time comparison between \texttt{ASIL}, \texttt{LSEnet} and CSE.}
\label{fig-new-lse-se-K}
  \vspace{-0.2in}
\end{figure}

In this section, we analyze the strength of the proposed \texttt{ASIL} against the classic structural entropy \cite{li2016structural} (referred to as \texttt{CSE} for short), and \texttt{LSEnet} \cite{sun2024lsenet}.
%The strengths regarding time complexity and clustering effectiveness are analyzed as follows.
\textbf{(1) Time Complexity.} 
We report the running time under different heights of partitioning trees in Fig \ref{fig-new-lse-se-K}.
\texttt{ASIL} shows better scalability than \texttt{CSE}.
Concretely, \texttt{CSE} has competitive time cost to \texttt{ASIL} on the small dataset (Football, right image of Fig. \ref{fig-new-lse-se-K}), 
but performs undesirably on larger datasets, e.g., an average of $30 \times$ time cost to \texttt{ASIL} on Cora (left image of Fig. \ref{fig-new-lse-se-K}).
In addition, the linear complexity of \texttt{ASIL} is regardless of the height of partitioning tree, 
while \texttt{CSE} scales exponentially over the height, 
e.g., $\mathcal{O}(N^3 \log^2 N)$ for height $3$,
making it prohibited on large graphs.
\textbf{(2) Clustering Effectiveness.}  
Note that, \texttt{CSE} itself neither generates cluster assignment nor node representations. 
We invoke the heuristic algorithm to construct a partitioning tree of height $2$, where the leaf-parent relationship is regarded as cluster assignment, 
and thus we instantiate \texttt{ASIL} to generate a deep tree of the same height for fair comparison.
We summarize the results on Cora, Citeseer, AMAP, Computers and PubMed in Table \ref{tab-se-nmi-ari}.
\texttt{ASIL} consistently outperforms \texttt{CSE} and \texttt{LSEnet} for all the cases.
It suggests that \emph{cluster assignment is improved by the hyperbolic geometry, attribute incorporation and neural Lorentz boost construction of \texttt{ASIL}.}

\subsection{Comparison with Our Previous LSEent \cite{icml24sun}} \label{CompareLSEnet}
This subsection demonstrates the significant improvement compared to \texttt{LSEnet} in the conference version.
\texttt{LSEnet} constructs the deep partitioning tree in hyperbolic space, while \texttt{ASIL} leverages neural Lorentz boost construction to refine the deep tree of \texttt{LSEnet}.
First, we compare the clustering effectiveness and report the results on Cora, Citeseer, AMAP, Photo and Pubmed datasets in Table \ref{tab-ablation}.
\texttt{ASIL} consistently achieves better performance in both NMI and ARI.
The performance gain is attributed to hyperbolic tree representation refinement and attribute incorporation via Neural Lorentz boost construction.
Second, we examine the running time under different tree heights in Fig. \ref{fig-new-lse-se-K}.
\texttt{ASIL} outperforms \texttt{LSEnet} on both datasets owing to the reduced complexity in the proposed reformulation and the sparsifying process in the neural Lorentz boost construction.
That is, 
%\texttt{ASIL} in this paper exhibits superior clustering ability to the previous LSEnet, with minimal computational overhead. 
\textbf{\texttt{ASIL} significantly outperforms  \texttt{LSEnet} in both effectiveness and efficiency on the benchmark datasets.}
(The superiority in expressiveness and imbalanced clusters is elaborated in Sec. \ref{Expressiveness} and Sec. \ref{Imbalance}, respectively.)

\begin{figure*}[t]
\centering 
\includegraphics[width=\linewidth]{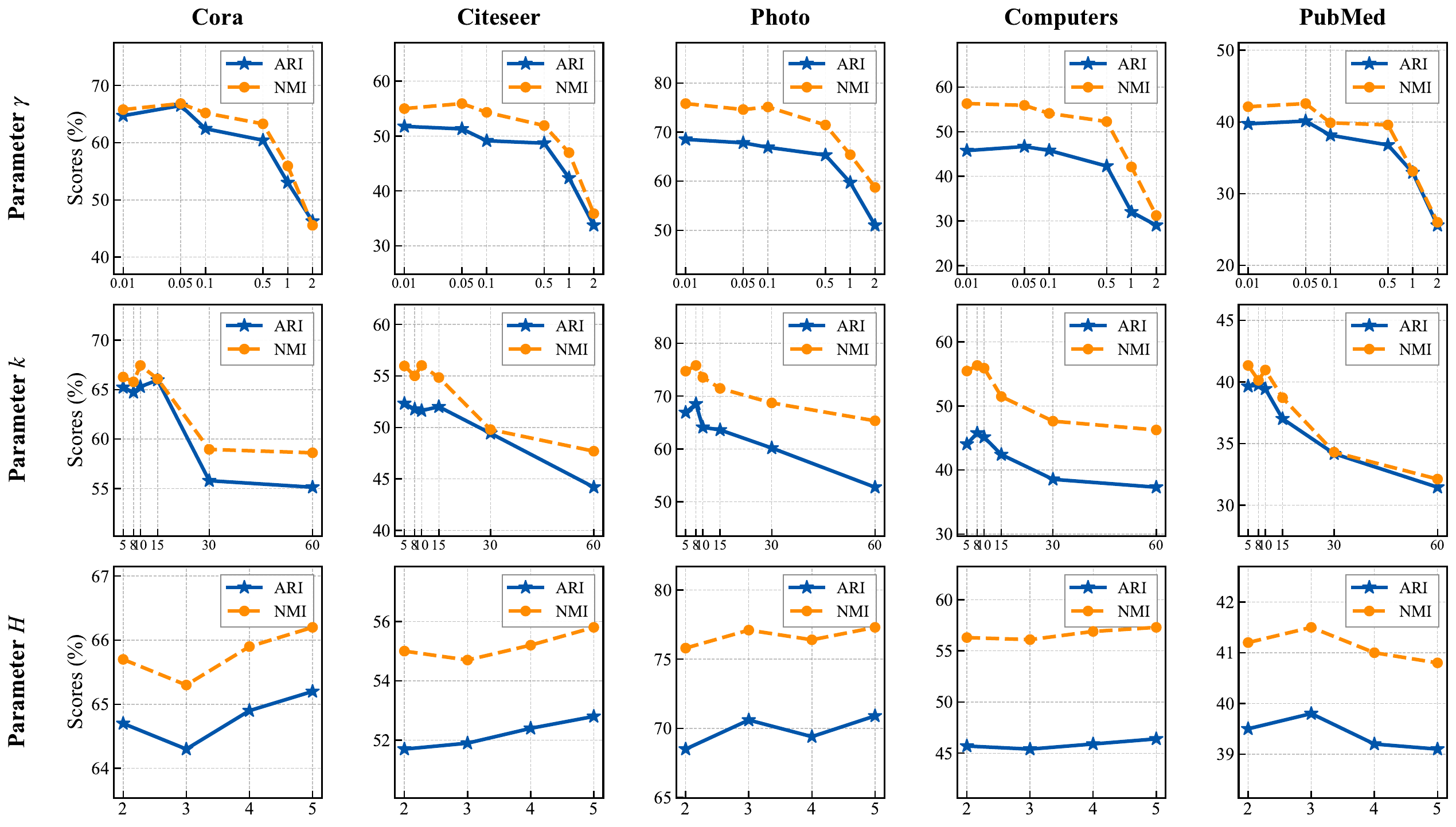}
    \vspace{-0.2in}
\caption{Sensitivity analysis on model hyperparameters $\gamma$, $k$ and $H$.}
    \vspace{-0.15in}
\label{fig:sensitivity}
\end{figure*}

\begin{table*}[t]
  \centering
  \caption{Ablation study results of different variants of \texttt{ASIL} on graph clustering tasks in terms of clustering performance (\%).}
    \vspace{-0.05in}
  \label{tab-ablation}
    \resizebox{1\linewidth}{!}{
     \begin{tabular}{ c |cc|cc|cc|cc|cc}
      \hline
         & \multicolumn{2}{c|}{\textbf{Cora}}  & \multicolumn{2}{c|}{\textbf{Citeseer}}  & \multicolumn{2}{c|}{\textbf{AMAP}} & \multicolumn{2}{c|}{\textbf{Computers}} & \multicolumn{2}{c}{\textbf{PubMed}} \\
        & NMI  & ARI  & NMI  & ARI  & NMI  & ARI  & NMI  & ARI & NMI & ARI \\
      \hline
    LSEnet
       & 63.97\scriptsize{$\pm$0.67}  & 63.35\scriptsize{$\pm$0.56}  & 52.26\scriptsize{$\pm$1.09}  & 48.01\scriptsize{$\pm$1.25} 
       & 71.12\scriptsize{$\pm$1.30}  & 65.08\scriptsize{$\pm$0.73}  & 55.03\scriptsize{$\pm$0.79}  & 42.15\scriptsize{$\pm$1.02} & 34.25\scriptsize{$\pm$0.24} & 35.16\scriptsize{$\pm$0.30} \\
    % w/o TS-Fusion
    %    & 62.77\scriptsize{$\pm$1.56}  & 61.17\scriptsize{$\pm$0.86}  & 50.45\scriptsize{$\pm$1.89}  & 47.34\scriptsize{$\pm$1.71} 
    %    & 71.31\scriptsize{$\pm$1.97}  & 63.18\scriptsize{$\pm$1.03}  & 54.64\scriptsize{$\pm$1.22}  & 42.74\scriptsize{$\pm$1.69} & 36.71\scriptsize{$\pm$0.49} & 37.01\scriptsize{$\pm$0.92} \\
    w/oLBoost
       & 61.18\scriptsize{$\pm$1.92}  & 60.73\scriptsize{$\pm$1.11}  & 51.16\scriptsize{$\pm$1.57}  & 46.92\scriptsize{$\pm$2.13} 
       & 71.47\scriptsize{$\pm$2.14}  & 63.57\scriptsize{$\pm$1.65}  & 53.76\scriptsize{$\pm$1.41}  & 40.79\scriptsize{$\pm$1.91} & 36.55\scriptsize{$\pm$0.74} & 37.12\scriptsize{$\pm$1.07} \\
    ExplicitTCL
    & 64.56\scriptsize{$\pm$0.72}  & 64.02\scriptsize{$\pm$0.70}   & 53.18\scriptsize{$\pm$1.11}  &50.45\scriptsize{$\pm$1.41}  
       & 74.54\scriptsize{$\pm$0.44}  & 69.24\scriptsize{$\pm$0.52}  & 55.93\scriptsize{$\pm$0.53} & 43.67\scriptsize{$\pm$0.35} &
       36.91\scriptsize{$\pm$0.33}  & 38.52\scriptsize{$\pm$0.23}\\
    \hline
   \textbf{\texttt{ASIL}}
     & \textbf{65.78}\scriptsize{$\pm$0.84}  & \textbf{64.72}\scriptsize{$\pm$0.67}   & \textbf{55.01}\scriptsize{$\pm$1.36}  & \textbf{51.77}\scriptsize{$\pm$1.23}
       & \textbf{75.81}\scriptsize{$\pm$0.33}  & \textbf{69.52}\scriptsize{$\pm$0.65}  & \textbf{56.31}\scriptsize{$\pm$0.55} & \textbf{45.77}\scriptsize{$\pm$0.35} &
      \textbf{38.13\scriptsize{$\pm$0.41}}  & \textbf{39.74\scriptsize{$\pm$0.59}}\\
      \hline
    \end{tabular}
}
  \vspace{-0.1in}
\end{table*}

\begin{table}[t]
  \centering
  \caption{Ablation study results of different representation space of partitioning tree of \texttt{ASIL}.}
    \vspace{-0.05in}
  \label{tab-space}
    \resizebox{\linewidth}{!}{
     \begin{tabular}{ c |cc|cc|cc}
      \hline
         & \multicolumn{2}{c|}{\textbf{Cora}}  & \multicolumn{2}{c|}{\textbf{Citeseer}}  & \multicolumn{2}{c}{\textbf{PubMed}} \\
        & NMI  & ARI  & NMI  & ARI  & NMI & ARI \\
      \hline
    \textbf{Hyperbolic}
       & \textbf{65.78}  & \textbf{64.72}  & \textbf{55.01}  & \textbf{51.77}
       & \textbf{38.13} & \textbf{39.74}\\
    \hline
    \textbf{Hyperspherical}
       & 52.39  & 51.75  & 44.93  & 40.77 
       & 28.05 & 28.17 \\
    \hline
    \textbf{Euclidean}
       & 58.39  & 55.77  & 49.72  & 47.92
       & 31.51 & 31.69 \\
    \hline
    \end{tabular}
}
  \vspace{-0.1in}
\end{table}

\subsection{Ablation Study} \label{AblationStudy}

This subsection introduces several variants to examine the effectiveness of each proposed module of \texttt{ASIL}, and comparison results are summarized in Table~\ref{tab-ablation}.

\subsubsection{\textbf{Importance of fusing a virtual graph}}
We examine the performance gain achieved by fusing a virtual graph $\mathbf{A}^\gamma$ constructed from hyperbolic node representations.
To this end, we introduce a variant that disables the construction of $\tilde{\mathbf{A}}$ and instead uses only the original topology $\mathbf{A}$, which indeed reduces to \texttt{LSEnet}.
\texttt{ASIL} significantly outperforms \texttt{LSEnet} on all datasets, particularly on AMAP that with imbalanced classes, where semantic are crucial for distinguishing topological noise.
This validates that the inclusion of a virtual enables \texttt{ASIL} to discover clusters enhanced by both structural and semantic perspectives.

\subsubsection{\textbf{Importance of consistency insurance, achieved by neural Lorentz boost}}
To investigate the role of consistency insurance in tree contrastive learning, we design a variant named \textit{w/oLBoost}, which replaces our neural Lorentz group layer with the standard Lorentz linear layer from~\cite{chen2022fullyb}.
While the baseline layer preserves the manifold structure, it lacks the isometric linearity that ensures geometric consistency between leaf and internal nodes in the partitioning tree, as formalized in Theorem~\ref{thm:consist}.
Table~\ref{tab-ablation} shows that \texttt{ASIL} achieves significantly better clustering than \textit{w/oLBoost}, demonstrating that consistency insurance are critical for high-fidelity tree contrastive learning.

\subsubsection{\textbf{Implicit or explicit tree contrastive learning}}
In this part, we demonstrate that the implicit tree contrastive learning  by a parameterized virtual graph outperforms explicit tree contrastive learning.
To this end, we give another variant named as ExplicitTCL. 
Specifically, we use an fixed isometric augmentation (details in Appendix X) to generate an augmentation view of original graph, 
and then use neural Lorentz boost to compute tree contrastive loss explicitly as in Eq.~\ref{TCL}. 
Thus, the total loss is given as $\mathcal{H}^{\mathcal{T}_{\text{net}}}(G^\gamma) + \lambda \mathcal{L}_{\text{TCL}}$, where $\lambda=0.01$ is a coefficient. 
The results are shown in the third line of Table~\ref{tab-ablation}. 
It shows that the proposed implicit approach consistently achieves better performance on the five datasets, which verifies the effectiveness of augmented structural entropy.

\subsubsection{\textbf{Representation Space of the Partitioning Tree}}
We conduct a geometric ablation to study the alignment between the underlying manifold and the hierarchical partitioning tree.
Note that our focus is on the geometry of the \textit{learned tree}, not the input graph manifold.
Theoretically, Theorem~\ref{thm:tree} establishes the low-distortion embeddability of trees in hyperbolic space.
Empirically, we instantiate \texttt{ASIL} in three constant-curvature spaces: hyperbolic ($\kappa < 0$), hyperspherical ($\kappa > 0$), and Euclidean ($\kappa = 0$), keeping all other components identical.
The clustering results on Cora, Citeseer, and Pubmed are given in Table~\ref{tab-space}.
These results confirm that hyperbolic space is uniquely suited for modeling hierarchical partitioning trees, consistently achieving the best performance across datasets.

\begin{table*}[t]
  \centering
  \caption{Embedding expressiveness regarding link prediction on Cora, Citeseer, AMAP, Computer, and Pubmed datasets. The best results are highlighted in \textbf{boldface}, and runner-ups are \underline{underlined}.}
  \vspace{-0.05in}
  \label{tab-LP}
  \renewcommand{\arraystretch}{1.2}
  \resizebox{1\linewidth}{!}{
    \begin{tabular}{ c |c c |c c |c c |c c |c c }
    \hline
                & \multicolumn{2}{c|}{\textbf{Cora}} & \multicolumn{2}{c|}{\textbf{Citeseer}} & \multicolumn{2}{c|}{\textbf{AMAP}} & \multicolumn{2}{c|}{\textbf{Computer}} & \multicolumn{2}{c}{\textbf{Pubmed}} \\
    Model      & AUC       & AP       & AUC       & AP       & AUC       & AP       & AUC       & AP       & AUC       & AP       \\
    \hline
    GCN        & 91.19\scriptsize{$\pm$0.51}  & 91.85\scriptsize{$\pm$0.43}  & 90.16\scriptsize{$\pm$0.49}  & 91.91\scriptsize{$\pm$0.24}  & 90.12\scriptsize{$\pm$0.72}  & 91.08\scriptsize{$\pm$0.08}  &  93.86\scriptsize{$\pm$0.36}  & 93.24\scriptsize{$\pm$0.32}  & 91.16\scriptsize{$\pm$0.14}  & 89.96\scriptsize{$\pm$0.21}  \\
    SAGE       & 86.02\scriptsize{$\pm$0.55}  & 88.24\scriptsize{$\pm$0.87}  & 88.18\scriptsize{$\pm$0.22}  & 89.15\scriptsize{$\pm$0.36}  & 98.02\scriptsize{$\pm$0.32}  & \textbf{97.51}\scriptsize{$\pm$0.21}  & 92.82\scriptsize{$\pm$0.20}  & 93.12\scriptsize{$\pm$0.14}  & 89.22\scriptsize{$\pm$0.87}  & 89.44\scriptsize{$\pm$0.82}  \\
    GAT        & 92.55\scriptsize{$\pm$0.49}  & 90.27\scriptsize{$\pm$0.23}  & 89.32\scriptsize{$\pm$0.36}  & 90.08\scriptsize{$\pm$0.06}  & 98.67\scriptsize{$\pm$0.08}  & 97.21\scriptsize{$\pm$0.34}  & 95.93\scriptsize{$\pm$0.19}  & 94.57\scriptsize{$\pm$0.24}  & 91.60\scriptsize{$\pm$0.37}  & 90.37\scriptsize{$\pm$0.21}  \\
    DGI        & 90.02\scriptsize{$\pm$0.80}  & 90.61\scriptsize{$\pm$1.00}  & \underline{94.53}\scriptsize{$\pm$0.40}  & \underline{95.72}\scriptsize{$\pm$0.10}  & 98.24\scriptsize{$\pm$0.13}  & 96.84\scriptsize{$\pm$0.24}  & 95.84\scriptsize{$\pm$0.05}  & 95.22\scriptsize{$\pm$0.36}  & 91.24\scriptsize{$\pm$0.60}  & 92.23\scriptsize{$\pm$0.50}  \\
    \hline 
    HGCN       & 93.60\scriptsize{$\pm$0.37}  & 93.64\scriptsize{$\pm$0.12}  & 94.39\scriptsize{$\pm$0.42}  & 94.38\scriptsize{$\pm$0.11}  & 98.06\scriptsize{$\pm$0.29}  & 96.84\scriptsize{$\pm$0.15}  & 96.88\scriptsize{$\pm$0.53}  & 96.18\scriptsize{$\pm$0.13}  & 96.30\scriptsize{$\pm$0.28}  & 94.72\scriptsize{$\pm$0.52}  \\
    LGCN       & 92.69\scriptsize{$\pm$0.26}  & 91.09\scriptsize{$\pm$0.37}  & 93.49\scriptsize{$\pm$1.11}  & 94.96\scriptsize{$\pm$0.18}  & 97.08\scriptsize{$\pm$0.08}  & 96.73\scriptsize{$\pm$0.52}  & 96.37\scriptsize{$\pm$0.70}  & 95.66\scriptsize{$\pm$0.24}  & 95.40\scriptsize{$\pm$0.23}  & 93.12\scriptsize{$\pm$0.24}  \\
    $Q$GCN     & 95.22\scriptsize{$\pm$0.29}  & 94.12\scriptsize{$\pm$0.51}  & 94.31\scriptsize{$\pm$0.73}  & 94.08\scriptsize{$\pm$0.34}  & 95.17\scriptsize{$\pm$0.45}  & 93.94\scriptsize{$\pm$0.18}  & 95.10\scriptsize{$\pm$0.03}  & 94.28\scriptsize{$\pm$0.33}  & 96.20\scriptsize{$\pm$0.34}  & 94.24\scriptsize{$\pm$0.31}  \\
    \hline
    \texttt{LSEnet}  & \underline{95.51}\scriptsize{$\pm$0.60}  & \underline{95.40}\scriptsize{$\pm$0.04}  & 94.42\scriptsize{$\pm$1.51}  & 94.69\scriptsize{$\pm$0.05}  & \underline{98.75}\scriptsize{$\pm$0.67}  & 97.04\scriptsize{$\pm$0.13}  & \underline{97.06}\scriptsize{$\pm$1.02}  & \underline{97.06}\scriptsize{$\pm$0.05}  & \underline{96.41}\scriptsize{$\pm$0.10}  & \underline{95.46}\scriptsize{$\pm$0.10} \\
    \textbf{\texttt{ASIL}}  & \textbf{97.11}\scriptsize{$\pm$0.42} & \textbf{96.74}\scriptsize{$\pm$0.34}  & \textbf{95.11}\scriptsize{$\pm$0.23}  & \textbf{96.01}\scriptsize{$\pm$0.67}  & \textbf{98.89}\scriptsize{$\pm$0.54}  & \underline{98.66}\scriptsize{$\pm$0.43}  & \textbf{97.85}\scriptsize{$\pm$0.32}  & \textbf{98.15}\scriptsize{$\pm$0.45}  & \textbf{97.81}\scriptsize{$\pm$0.09}  & \textbf{96.27}\scriptsize{$\pm$0.12} \\
    \hline
    \end{tabular}
  }
  \vspace{-0.1in}
\end{table*}

\subsection{Hyperparameter Sensitivity}
\label{sec:sensitivity}

In this subsection, we analyze the sensitivity of \texttt{ASIL} to three key hyperparameters:
(i) the number of neighbors $k$ for sparsifying $\tilde{\mathbf{A}}$,
(ii) the coefficient $\gamma$ of Lorentz boost construction $\mathbf{A}^\gamma = (1-\gamma) \mathbf{A} + \gamma \tilde{\mathbf{A}}$,
and (iii) the height $H$ of the partitioning tree (i.e., the dimension of structural entropy).

\subsubsection{\textbf{Coefficient for Lorentz Boost Construction}}
The coefficient $\gamma$ controls the importance of representational perspective in the graph fusion.
We test $\gamma \in \{0.01, 0.05, 0.1, 0.5, 1.0, 2.0\}$.
When $\gamma=0$, \texttt{ASIL} degenerates to the original \texttt{LSEnet}.
As shown in Figure~\ref{fig:sensitivity}(a--e), when $\gamma$ gets larger from $0.5$ to $1.0$, the performance of \texttt{ASIL} gets worse sharply since too much information injection may be treated as noisy structure, especially when $\gamma=2.0$ that exceeds the scope defined in Theorem~\ref{thm:fused_conductance}.

\subsubsection{\textbf{$k$-NN Neighborhood Size ($k$)}}
The virtual graph $\tilde{\mathbf{A}}$ is constructed via $k$-nearest neighbors in Lorentz space.
We vary $k \in \{5, 8, 10, 15, 30, 60\}$ and fix other hyperparameters.
As shown in Figure~\ref{fig:sensitivity}(f--j), a small $k$ (e.g., $k=5, 8, 10$) achieves well performance while a large $k$ (e.g., $k=30, 60$) introduces noisy cross-cluster edges, destroying clustering effectiveness.

\subsubsection{\textbf{Height of Partitioning Tree ($H$)}}
The hyperparameter $H$ determines the depth of the hierarchical partitioning tree.
We evaluate $H \in \{2, 3, 4, 5\}$ and report results in Figure~\ref{fig:sensitivity}(k--o).
We observe that performance is sensitive to $H$, with optimal values depending on graph complexity:
\texttt{ASIL} achieves peak NMI on Cora at $H=2$, while AMAP prefers $H=3$.
This suggests that deeper trees benefit more complex graphs with hierarchical community structures.
Notably, \texttt{ASIL} remains stable across a wide range of $H$, demonstrating robustness to model capacity. In summary, \texttt{ASIL} is generally robust to hyperparameter choices, and default values ($k=8$, $\alpha=0.01$, $H=2$) work well across datasets.
%This practical simplicity, combined with strong theoretical grounding, makes \texttt{ASIL} appealing for real-world applications.

\subsection{Embedding Expressiveness of \texttt{ASIL}}  \label{Expressiveness}
\texttt{ASIL} generates cluster assignment as well as node embedding in hyperbolic space.
Here, we evaluate the embedding  expressiveness  with link prediction. 
To this end, \texttt{ASIL} is compared with the Euclidean models
(including GCN \cite{kipf2017semisupervised}, SAGE \cite{hamilton2018inductive}, GAT \cite{velickovic2018graph} and DGI \cite{velivckovic2018deep}),
hyperbolic models (including HGCN \cite{chami2019hyperbolica} and LGCN \cite{zhang2021lorentzian}), 
a recent $Q$GCN in ultra hyperbolic space \cite{nips22QGCN}
and our previous \texttt{LSEnet}.
Popular metrics of AUC (Area Under the Curve) and AP (Average Precision) are employed to quantify the results.
The link prediction performances on the five datasets are collected in Table \ref{tab-LP}, where the  height of partitioning tree is set as $3$ for both \texttt{LSEnet} and \texttt{ASIL}.
As shown in Table~\ref{tab-LP}, Riemannian models generally outperforms the Euclidean counterparts, and \texttt{ASIL} achieves the state-of-the-art results. 
It suggests that Riemannian manifolds better model the complexity of graph structures and, more importantly, the partitioning tree encodes fruitful information for link prediction.
%preserving cluster structure contributes to the success of our model.
 Also, compared to \texttt{LSEnet}, \texttt{ASIL} consistently obtains the performance gain thanks to the implicit Lorentz tree contrastive learning.

\begin{figure*}[t]
\centering 
\subfigure[gCool: true]{
\includegraphics[width=0.18\linewidth]{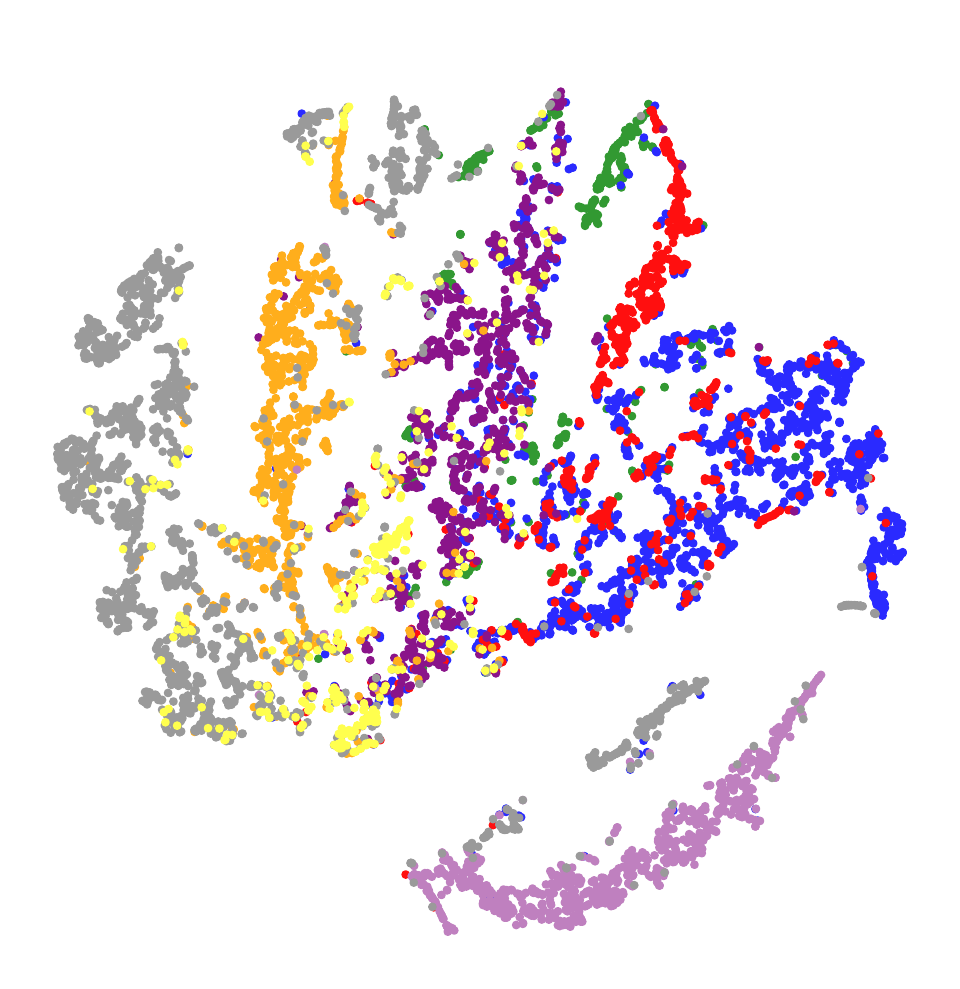}}
\subfigure[GDCluster: true]{
\includegraphics[width=0.18\linewidth]{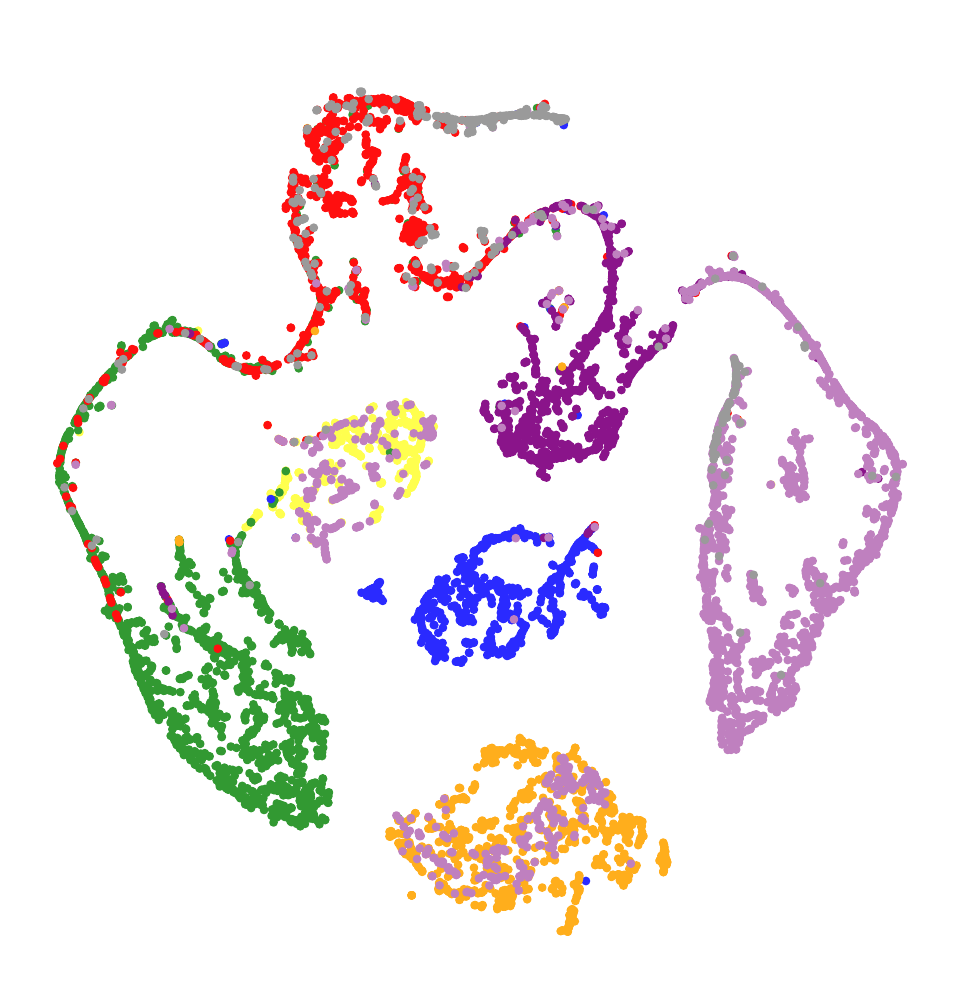}}
\subfigure[RGC: true]{
\includegraphics[width=0.18\linewidth]{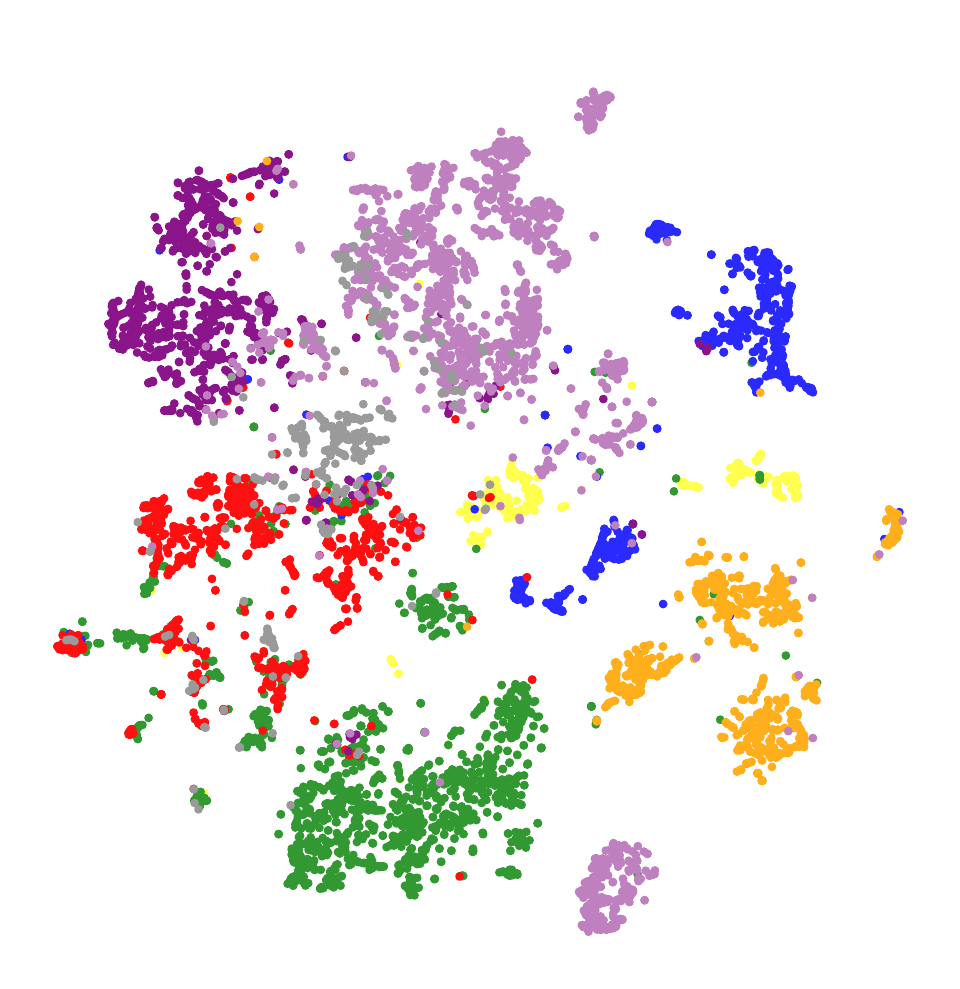}}
\subfigure[LSEnet: true]{
\includegraphics[width=0.18\linewidth]{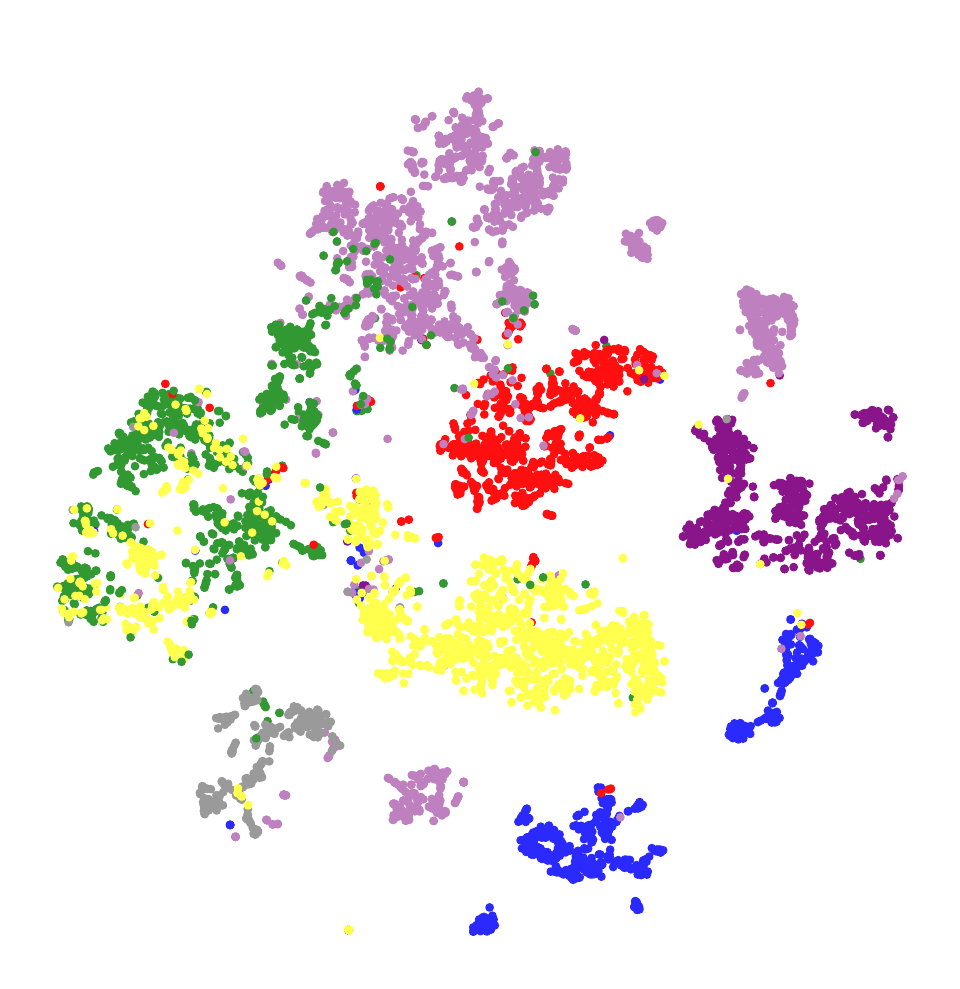}}
\subfigure[ASIL: true]{
\includegraphics[width=0.18\linewidth]{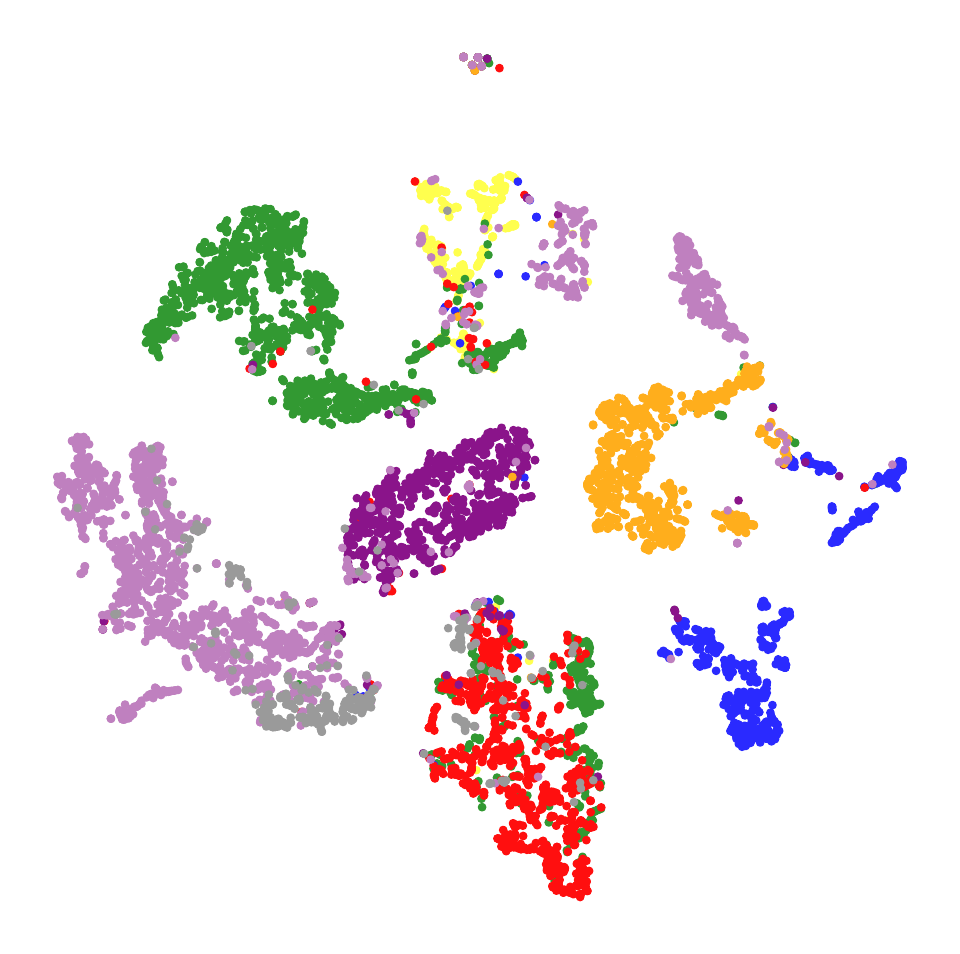}}
\subfigure[gCool: prediction]{
\includegraphics[width=0.18\linewidth]{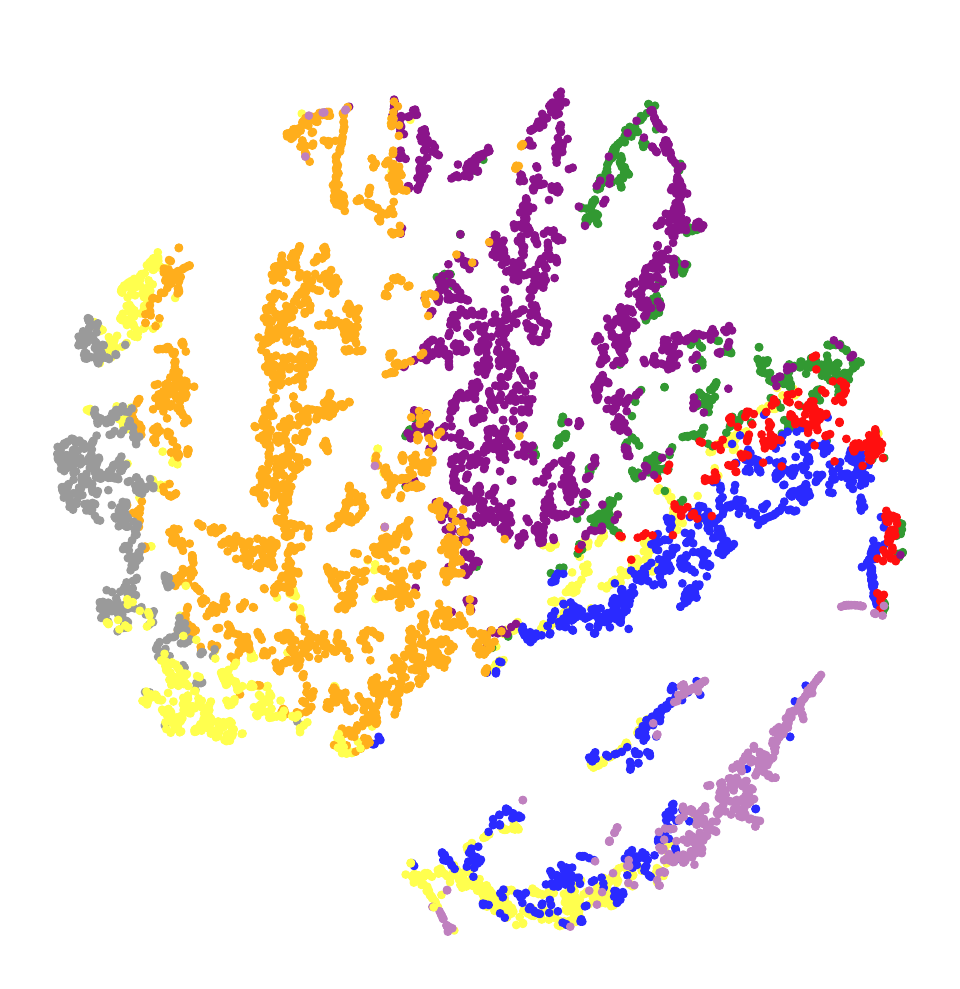}}
\subfigure[GDCluster: prediction]{
\includegraphics[width=0.18\linewidth]{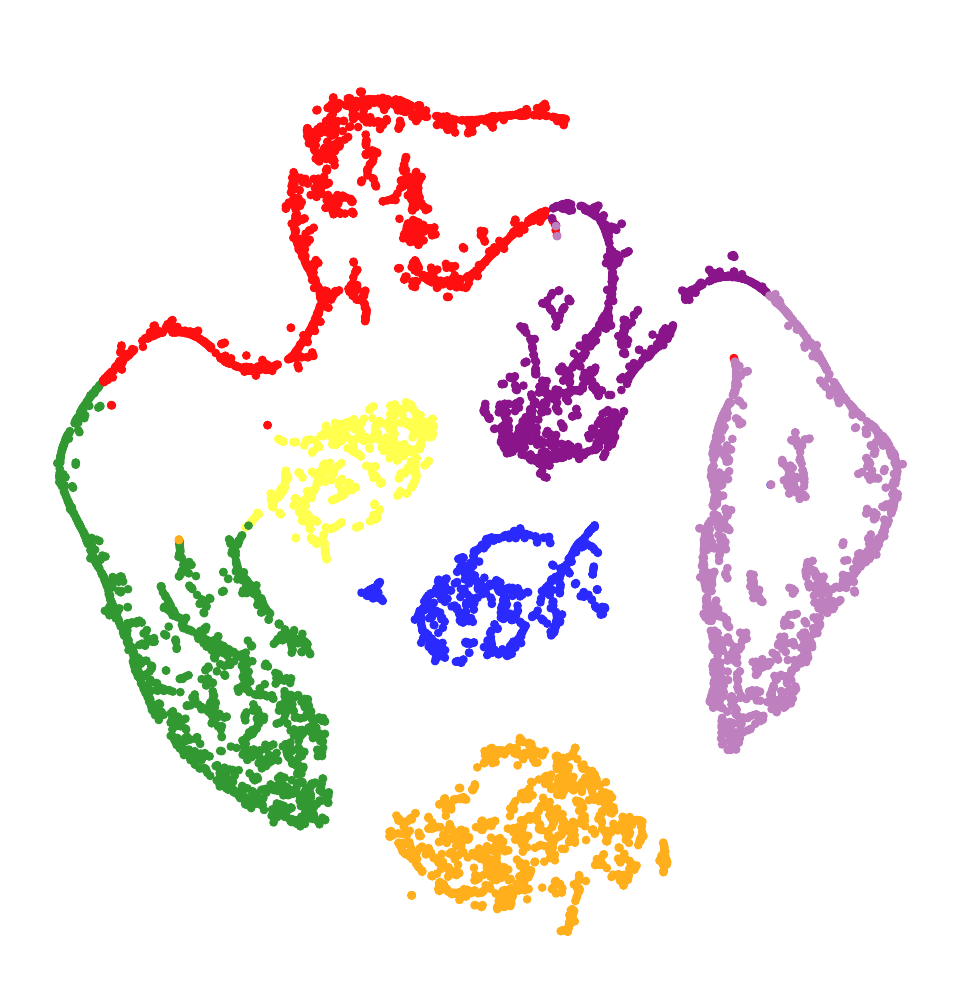}}
\subfigure[RGC: prediction]{
\includegraphics[width=0.18\linewidth]{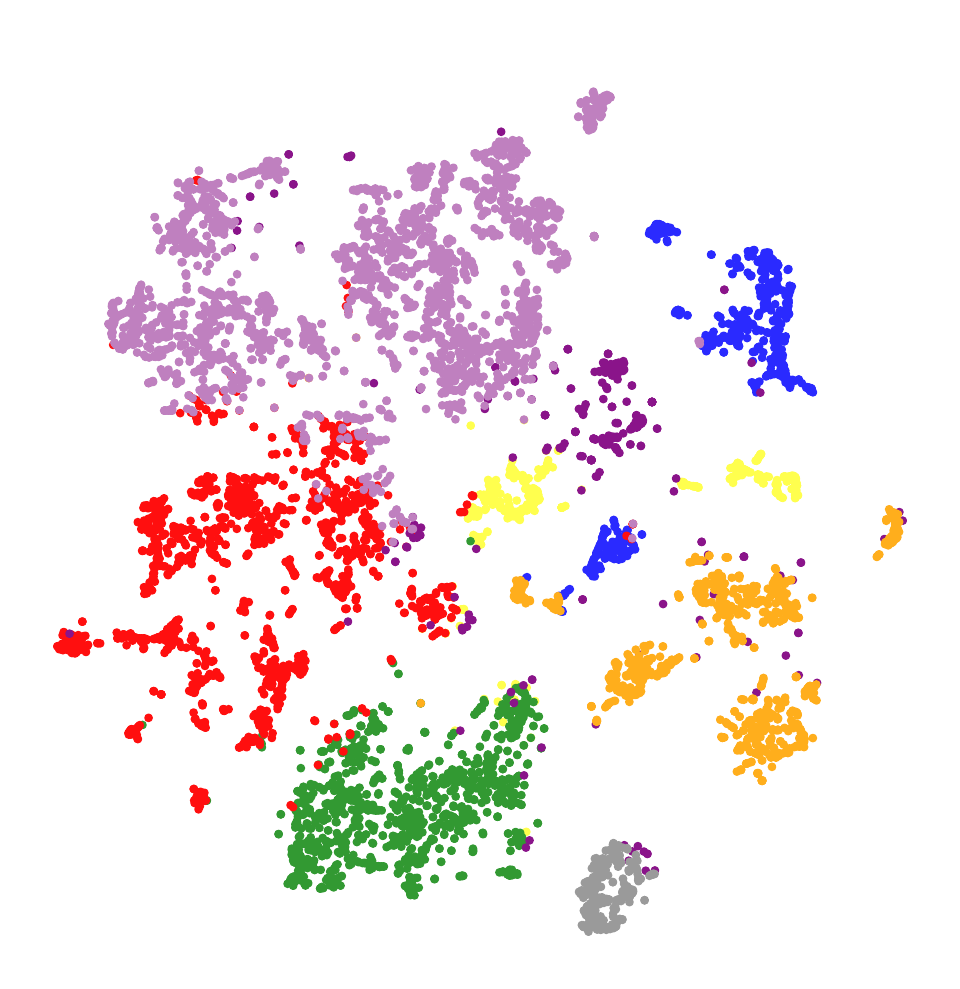}}
\subfigure[LSEnet: prediction]{
\includegraphics[width=0.18\linewidth]{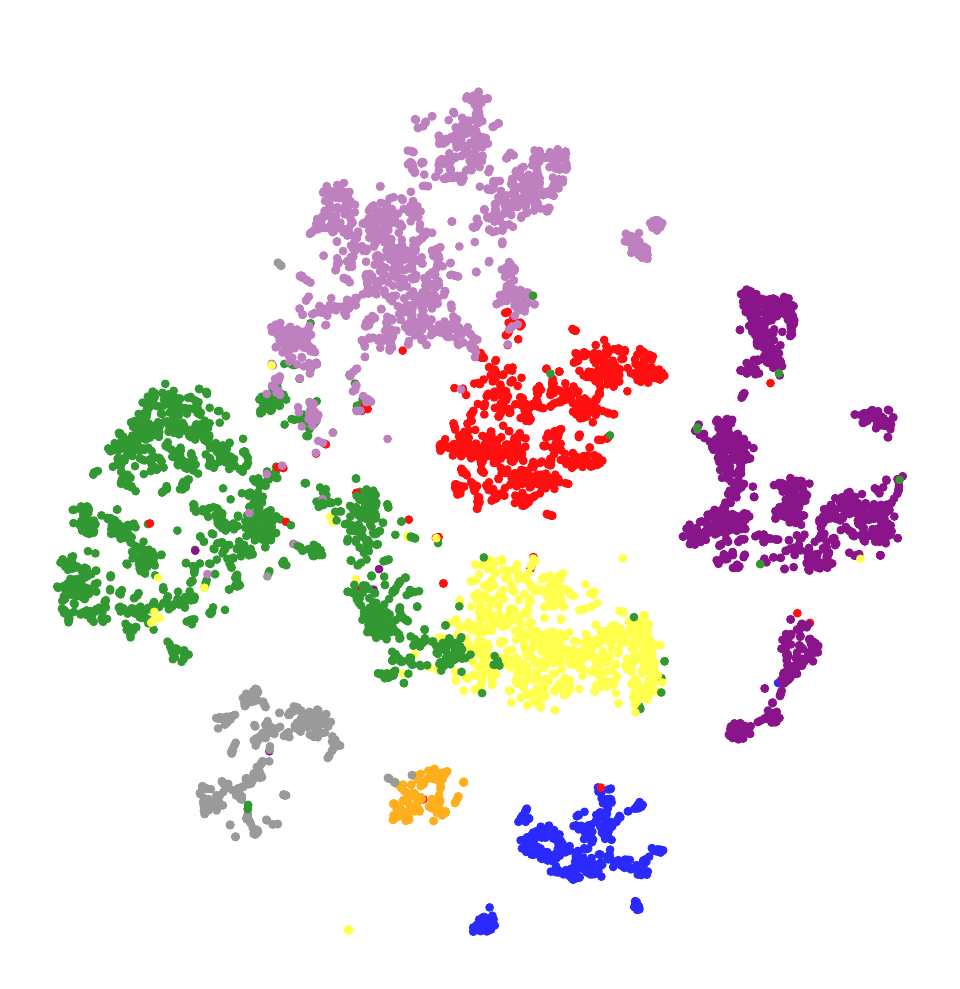}}
\subfigure[ASIL:prediction]{
\includegraphics[width=0.18\linewidth]{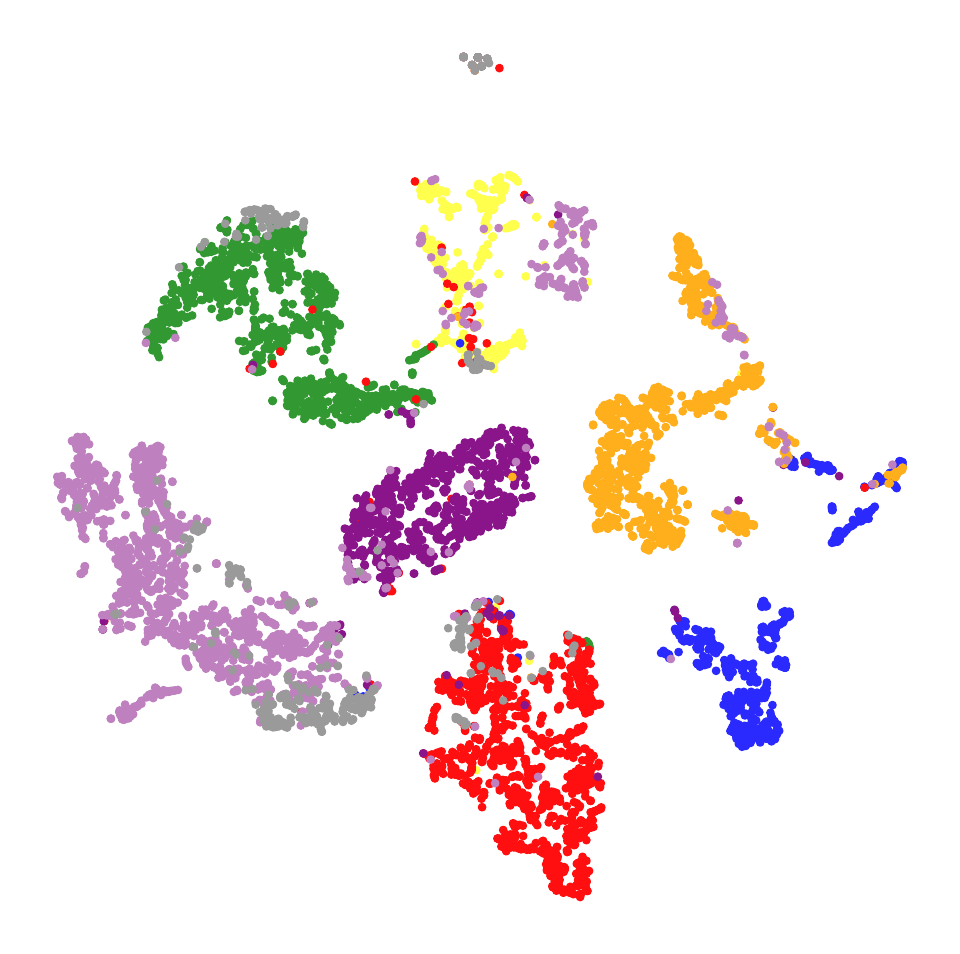}}
% \vspace{0.12in}
\caption{Visualization of real and predicted data on Photo dataset.}
\vspace{-0.1in}
\label{fig-new-clu-visual}
\end{figure*}

\subsection{Case Study on Imbalanced Clusters} \label{Imbalance}

We examine the clustering boundaries among imbalanced clusters. 
Real graphs often exhibit skewed class distributions, and for this case study, we use the Photo dataset.
%we utilize Photo dataset in this case study.
The proportion of $8$ clusters in Photo is $25\%$, $22\%$, $11\%$, $11\%$, $10\%$, $9\%$, $4\%$ and $4\%$.
We compare \texttt{ASIL} with gCool, DGCluster, RGC and \texttt{LSEnet}, and visualize the node embedding in Fig. \ref{fig-new-clu-visual} where different clusters are denoted by different colors.
% To be specific, we color the nodes according to true label in (), (), () and (), while nodes are colored by the predicted cluster in (), (), () and ().
The distributions of true labels and predicted labels are given in the top and bottom subfigures, respectively.
We invoke the normal t-sne for gCool, DGCluster, RGC,  while employing the hyperbolic metric for \texttt{LSEnet} and \texttt{ASIL}, so that all the visualizations are finally presented in 2D Euclidean space.
In Fig. \ref{fig-new-clu-visual}, we obtain three findings as follows.
(i)  Clustering model without $K$ may fail to uncover the correct cluster number, e.g., RGC outputs $7$ node clusters on Photo dataset, but  \texttt{ASIL} gives the true cluster number $8$.
(ii) 
We observe that deep clustering models tend to generate biased clustering boundary. 
% In particular, the members of smaller clusters are at risk of being assimilated by dominant clusters nearby, referred to as the ``black hole'' effect.
For example, in RGC, the members of \textcolor{violet}{\textbf{violet}} cluster and the minority \textcolor{gray}{\textbf{gray}} cluster   (at the upper left in Fig. \ref{fig-new-clu-visual}c) are assimilated into the majority cluster colored in  green, comparing  the true label in Fig. \ref{fig-new-clu-visual}h.
The reason lies in that the deep models typically perform clustering in the representation space and 
minority clusters are underrepresented in the self-supervised learning owing to limited samples.
(iii)  Different from other deep models, the minority clusters are easier to be identified and discovered in both \texttt{LSEnet} and \texttt{ASIL}.
We highlight that, the differentiable structural information objective effectively discovers node clusters without $K$ and  is suitable to identify minority clusters in  imbalanced graphs.
%owing to its inherent property as stated in Theorem X.
In addition, 
\textbf{compared to \texttt{LSEnet}, \texttt{ASIL} further mitigates the bias in clustering boundary with better hyperbolic partitioning tree representations.}

\begin{figure}[t]
\centering 
\subfigure[Karate: Results]{
\includegraphics[width=0.45\linewidth]{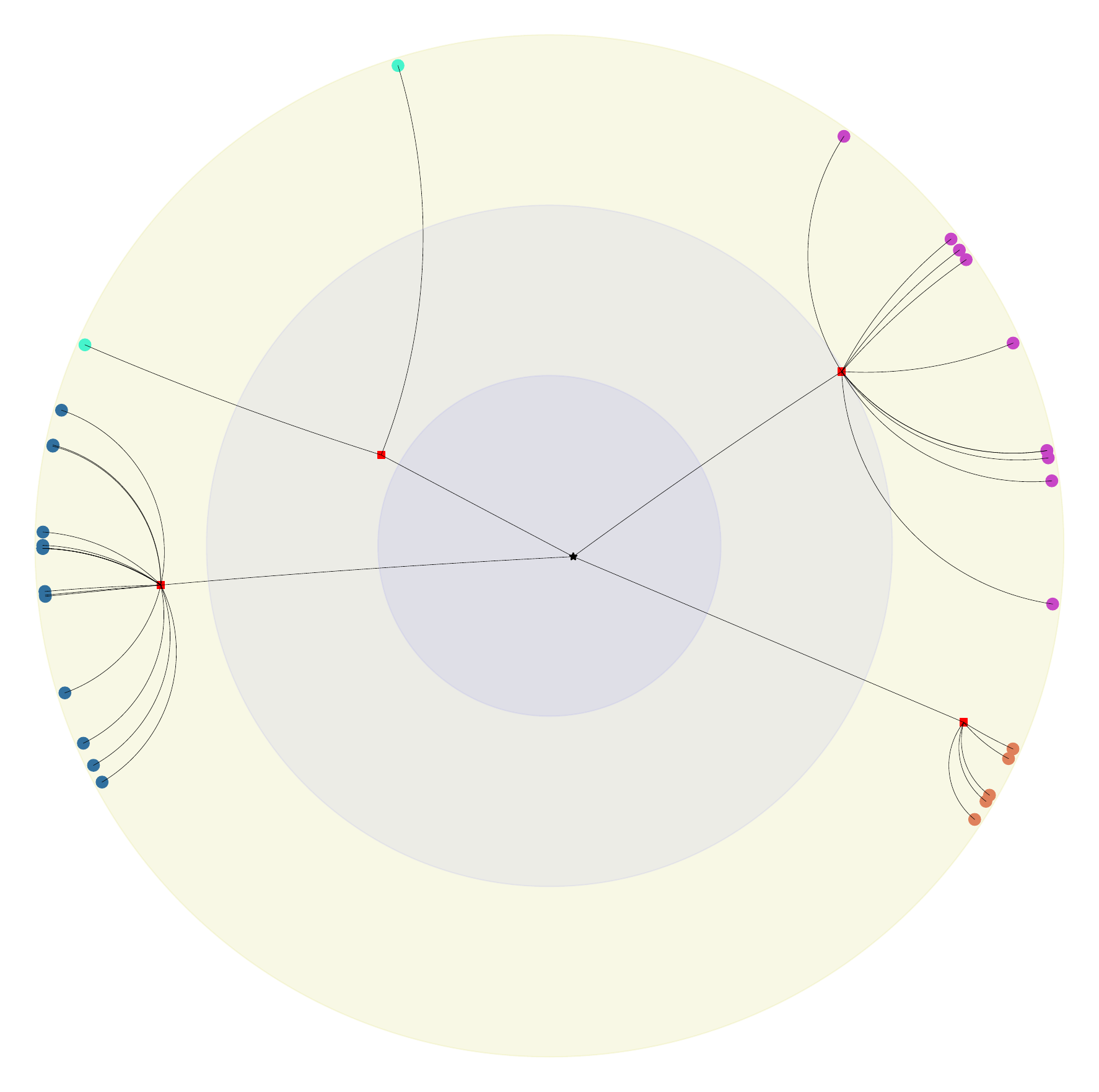}}
\subfigure[Football: Results]{
\includegraphics[width=0.45\linewidth]{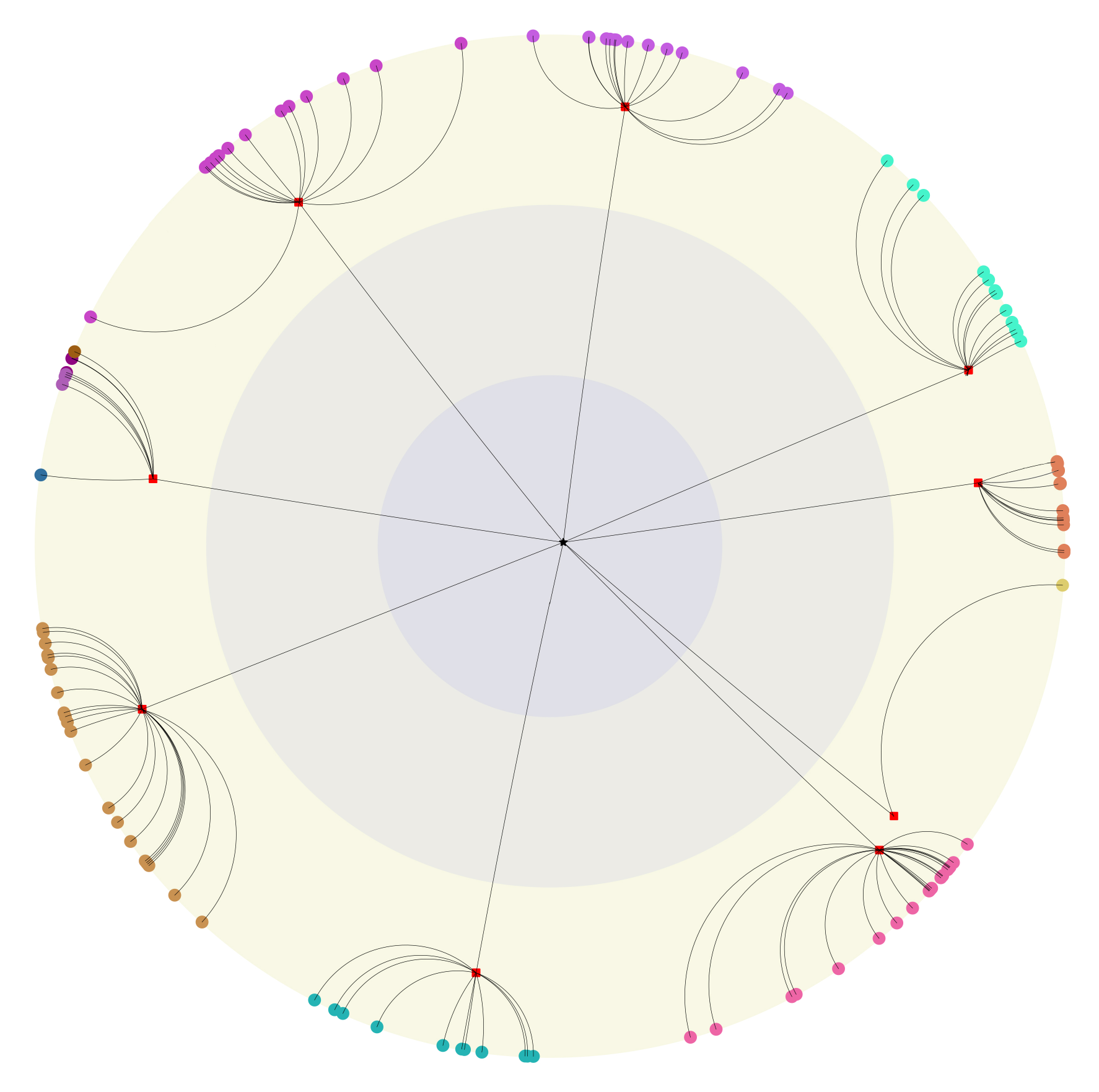}}
\subfigure[Karate: Groundtruth]{
\includegraphics[width=0.45\linewidth]{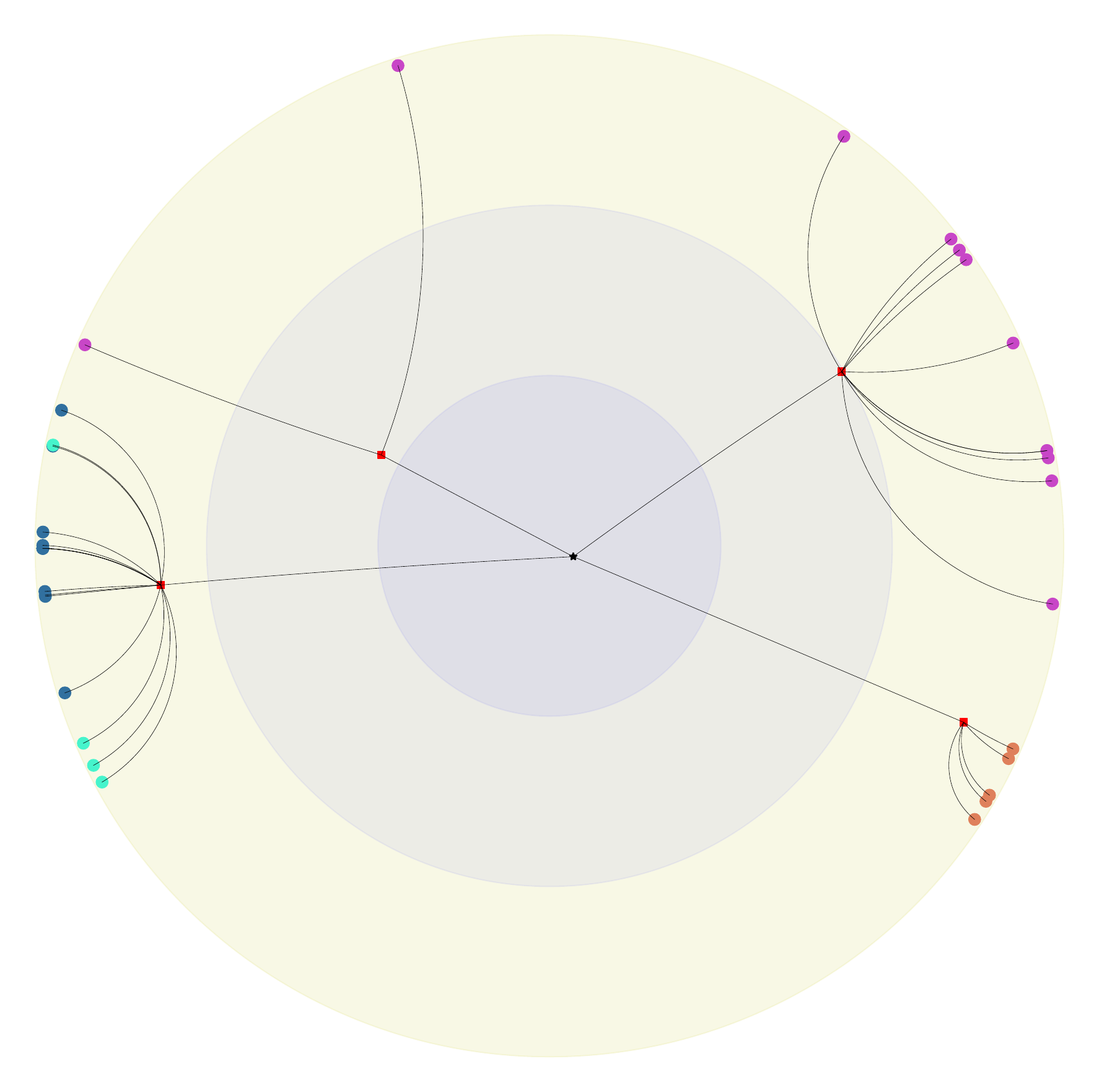}}
\subfigure[Football: Groundtruth]{
\includegraphics[width=0.45\linewidth]{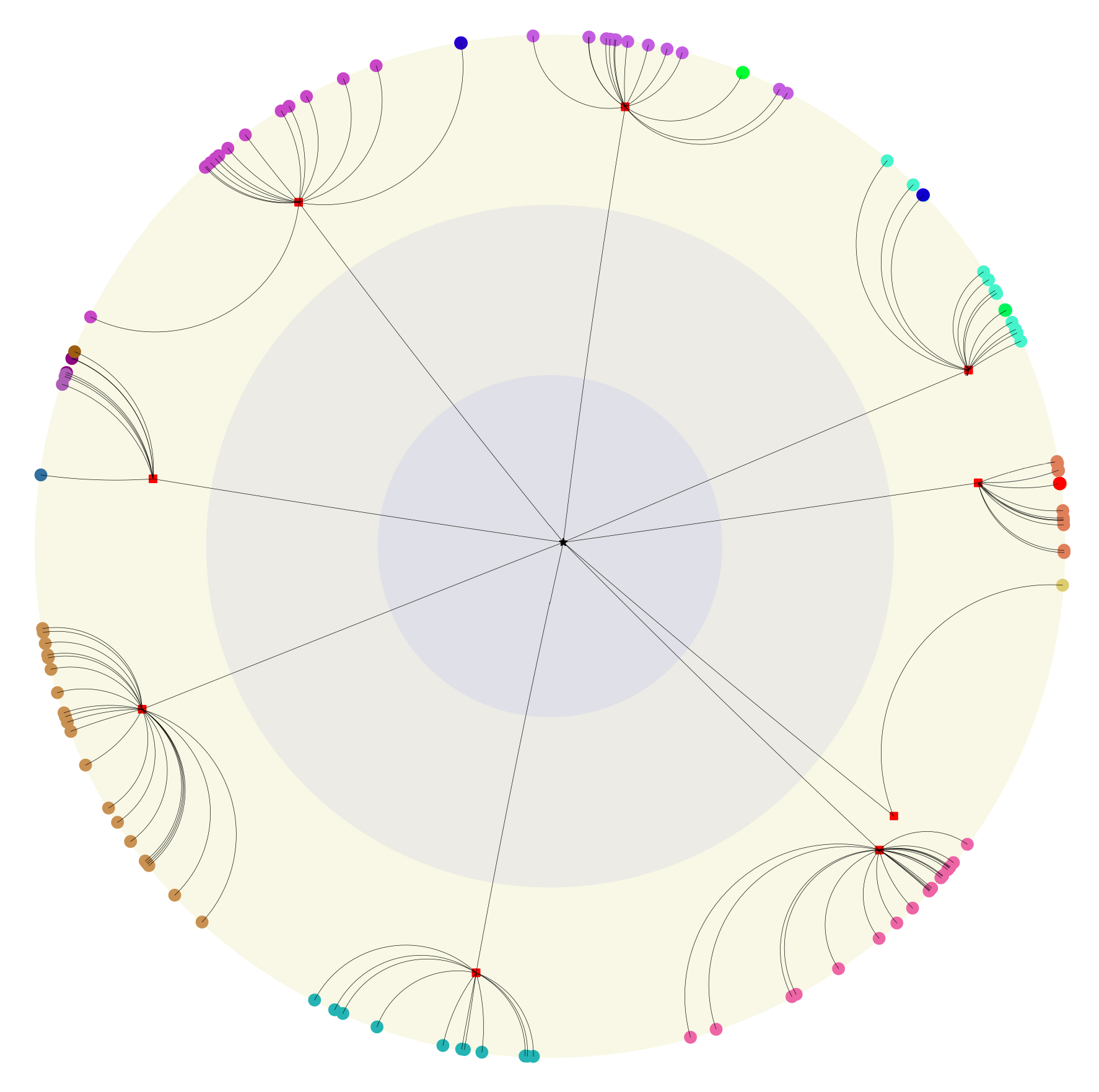}}
\caption{Visualization of hyperbolic partitioning trees.}
\label{fig-tree-visual}
  \vspace{-0.1in}
\end{figure}

% Here, we visualize hyperbolic partitioning trees of \texttt{LSEnet}, and discuss graph clustering on two toy examples of Football and Karate datasets.

\subsection{Visualizing Hyperbolic Partitioning Tree}  \label{Visualiziation}

We visualize the  hyperbolic partitioning trees of \texttt{ASIL}.
In \texttt{ASIL}, the partitioning tree is created in the Lorentz model of hyperbolic space.
Note the fact that the 2D Lorentz model is represented in 3D Euclidean space and is hard to be visualized.
To bridge this gap, we perform the stereographic projection $\Psi$ \cite{bachmann2020constant} on the Lorentz model 
and obtain the equivalent Poincar\'{e} disc $\mathbb{B}^2$, 
a 2D open disc that is more preferable for visualization.
Given the standard curvature $\kappa=-1$, 
for any point in the Lorentz model $\boldsymbol x=[x_1 \ \boldsymbol x_s]^\top \in \mathbb L^2$, 
we derive the corresponding point in Poincar\'{e} disc via $\Psi(\boldsymbol{x})=\frac{\boldsymbol x_s}{x_1 + 1} \in \mathbb{B}^2$.
With $\Psi$, we study the clustering results on two toy examples of Karate and Football datasets, and show the visualization in the 2D Poincar\'{e} disc in Fig. \ref{fig-tree-visual}.
The partitioning tree with learnt clusters are plotted in Fig. \ref{fig-tree-visual} (a) and (b), while true labels are given in Fig. \ref{fig-tree-visual} (c) and (d).
Two important findings are summarized as follows:
(i)  The clustering results of \texttt{ASIL} are generally aligned with the ground truth distribution.
(ii) The learnt embeddings of each clusters are well separated, showing the effectiveness of \texttt{ASIL}.

\section{Related Work}

\subsection{Deep Graph Clustering}

In recent years, deep graph clustering achieves remarkable success thanks to the expressive representation learning.\footnote{In this paper we study node clustering on the graph, and graph-level clustering is out of our scope.}
Typically, graph clustering requires the cluster number $K$ as the prior knowledge, and existing models can be roughly categorized into three groups.
(i) Reconstructive methods learn node representation by recovering graph information \cite{ding2023graph}, while generating node clusters via distance-based or model-based algorithms (e.g., k-means, Gaussian mixture model) \cite{xie2016unsupervised, almahairi2016dynamic, yang2017graph}.
(ii) Adversarial methods acquire  the clustering results through the minmax game between generator and discriminator \cite{DBLP:conf/ijcai/0002WGWCG20, DBLP:conf/www/JiaZ0W19, ding2024towards}. 
(iii) Contrastive methods obtain informative representation by pulling positive samples together while pushing negative samples apart \cite{devvrit2022s3gc, pan2021multi, DBLP:conf/www/LiJT22, liu2024revisiting}.
Recently, generative models, e.g., normalizing flows have  been explored \cite{wang2023gc}.
Self-supervised objective regarding Ricci curvature is designed for node clustering \cite{sun2023contrastive}.
%It is not until recently that deep graph clustering without $K$ has been investigated.
In the literature, research on deep graph clustering without $K$ is still limited.
For instance, DGCluster generates node encodings by optimizing modularity  and leverages Brich algorithm to obtain clustering results \cite{bhowmick2024dgcluster}.
RGC automates the training-testing process to search the optimal $K$ via reinforcement learning \cite{liu2023reinforcement}.
Both of them have different focus from ours, and we aim at uncovering the inherent cluster structure from the information theory perspective.
In parallel, efforts have been made to address the imbalance in graphs \cite{ImbalanceSurvey,icml24imbalance}, given  that real-world graphs often exhibit skewed class distributions.
Although resampling and reweighing techniques have been designed for imbalanced graphs \cite{shi2020multi,cui2019class,aaai23imbalanceNC}, the clustering objective for handling imbalance clusters remains underexplored. 
Our work is dedicated to bridging this gap.

\subsection{Structural Entropy}

In the information theory, 
Shannon entropy measures the amount of information for unstructured data \cite{6773024}, but the information measure over graphs remains open.
In the literature, efforts have been made to study the graph information measure.
Von Neumann entropy \cite{bra06} and Gibbs entropy \cite{bia09} fall short in modeling the graph structure,
while structural entropy addresses the absence of structural information and constructs a partitioning tree to reveal the self-organizing of graphs \cite{li2016structural}. 
Recently, it  has been successfully applied to biochemical analysis \cite{li2016three}, dimensionality estimation \cite{yang2023minimum}, graph classification \cite{wu2022structural}, adversarial attack \cite{liu2019rem}, structure augmentation \cite{zou2023segslb} and social event detection \cite{yu2024dame}.
%but has not yet been introduced to deep graph clustering.
Theoretical advances include generalization to incremental setting \cite{yang2024incremental} and incorporation with mutual information \cite{zeng2024effective}.
We observe that structural entropy is expressed as a discrete formula and its optimal partitioning tree has thus far been constructed using the heuristic algorithm \cite{wu2022structural},
leading to high computational complexity that is impractical for real graphs.
%The shortcomings prevent the application of structural entropy to deep graph clustering.
To address these challenges, this paper introduces the differential structural information,  and theoretically demonstrates its effectiveness in  clustering without requiring K, as well as in identifying minority clusters in  imbalanced graphs.
%Beyond the context of graph clustering, we open the opportunity to learn structural information with the gradient-based optimization.
Beyond graph clustering, our work paves the way for learning structural information through gradient-based optimization.

\subsection{Riemannian Deep Learning on Graphs}

Euclidean space has been the workhorse of graph learning for decades \cite{kipf2017semisupervised,velickovic2018graph},
while Riemannian manifolds have emerged as an exciting alternative in recent years.
In Riemannian geometry, a  structure is related to certain manifold with respect to its geometric property.
Among Riemannian manifolds, hyperbolic space is well recognized for its alignment to hierarchical/tree-like structures.
A series of hyperbolic GNNs (e.g., HGNN \cite{liu2019hyperbolic}, HGCN \cite{chami2019hyperbolica}) show superior performance to their Euclidean counterparts. 
In the literature, hyperbolic manifold routinely find itself in graph representation learning, recommender system, knowledge graphs, etc., 
In fact, an arbitrary tree can be embedded in hyperbolic space with bounded distortion \cite{sarkar2012low},
but the connection between hyperbolic manifold and structural entropy has not yet been established, to the best of our knowledge.
Beyond hyperbolic manifold, 
recent studies visit hyperspherical manifold \cite{icml23sphereFourier,nips19MettesPS}, gyrovector spaces \cite{bachmann2020constant,icml23GyroSpace}, product spaces \cite{iclr19Gu,aaai22SelfMix}, quotient spaces \cite{law2021ultrahyperbolic,nips22QGCN}, SPD manifolds \cite{gao2020learning}, 
as well as the notion of Ricci curvature for learning on graphs \cite{icml23revisitRicci,iclr22curvature}.
However, previous Riemannian models study the underlying manifold of the complex graph structures, while we primarily focus on the hyperbolic space to create and refine the partitioning tree for graph clustering.

\section{Conclusion}
Our work presents a principled solution for graph clustering from an information-theoretic perspective, considering the practical challenges of unknown cluster number and imbalanced node distribution.
To this end, we extend the classic structural entropy to the continuous realm, so that its partitioning tree can be learned via gradient backpropagation, uncovering clustering structures. 
Theoretically, we demonstrate its capacity in debiased clustering and connect it to contrastive learning, allowing for effective and efficient clustering via Augmented Structural Information Learning (\texttt{ASIL}). 
The key innovation is the seamless integration of partitioning tree construction and contrastive learning in hyperbolic space. Without requiring the number of clusters,  \texttt{ASIL} achieves  debiased  node clustering with linear computational complexity in graph size.
Empirically, we validate the superiority of \texttt{ASIL} against $20$ strong baselines on multiple benchmark datasets.

% Our work bridges structural information with deep learning and provides a principled solution for graph clustering without 
% without requiring a predefined number of clusters while accounting for the imbalance in real-world graphs.
% Theoretically, we establish differentiable structural information in continuous realm,
%  and prove its effectiveness in clustering without $K$ and detecting  minority clusters.
% Accordingly,
% we present  the  \textbf{\texttt{IsoSEL}} framework where  \texttt{LSEnet} is designed to learn a deep partitioning tree via gradient backpropagation in the Lorentz model of hyperbolic space. 
% Furthermore, we propose Lorentz tree contrastive learning with isometric augmentation to refine the deep tree, thereby enhancing cluster assignment.
% Extensive experiments demonstrate that \texttt{IsoSEL}  effectively generates node clusters and  mitigates the bias in the clustering boundary among imbalanced clusters.

% Furthermore, we isometrically incorporate the attribute information by Lorentz tree constrastive learning, 
% which simultaneously refines the  hyperbolic tree representation to facilitate 


% \section*{Acknowledgments}

% \section{References Section}

\bibliography{icml2024}
\bibliographystyle{IEEEtran}

\vspace{6 pt}

\begin{IEEEbiography}[{\includegraphics[width=1in,height=1.25in,clip,keepaspectratio]{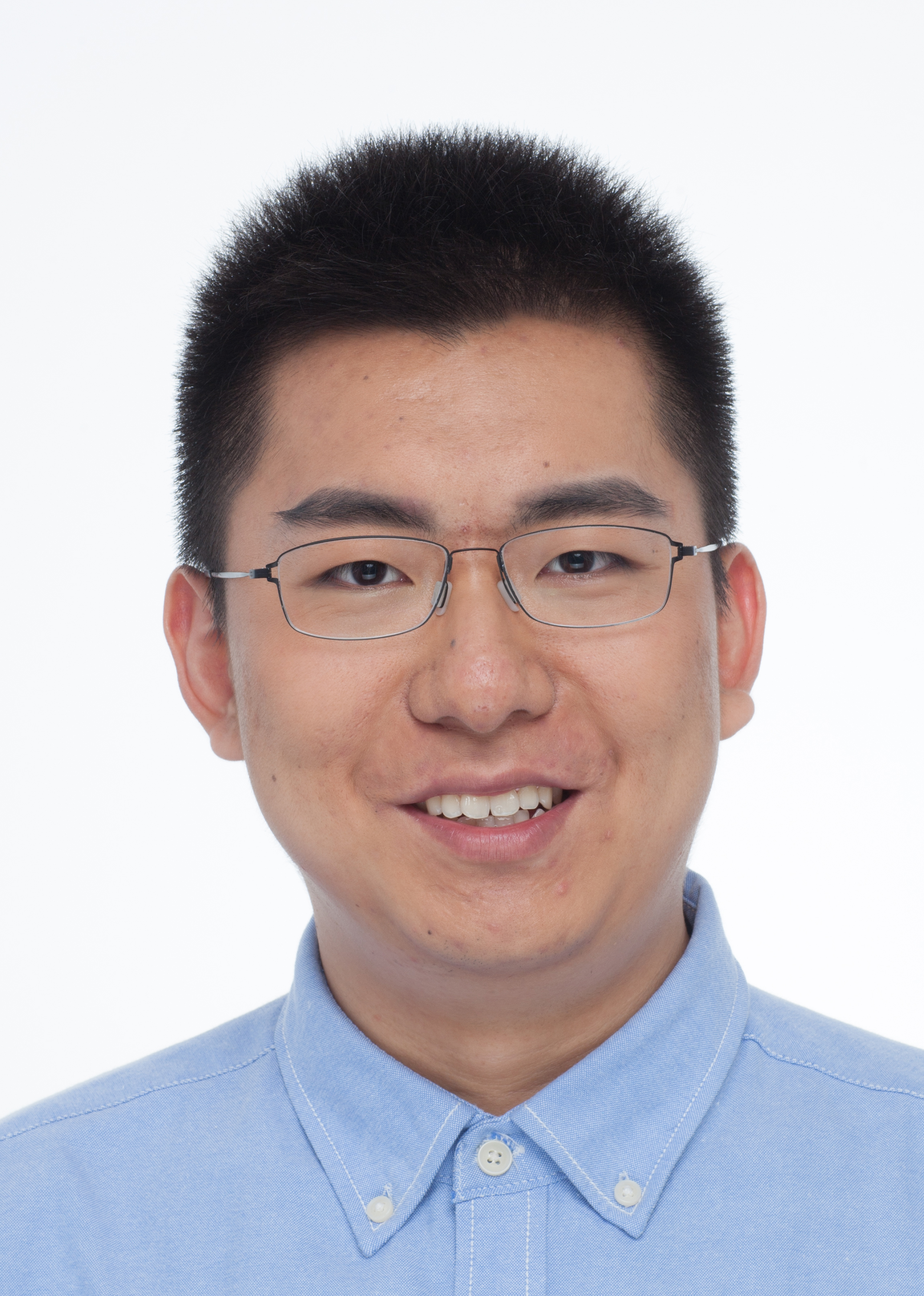}}]{Li Sun} received the Ph.D. degree in computer science from Beijing University of Posts and Telecommunications in 2021. He has been a visiting scholar at the University of Illinois, Chicago,  USA, advised by Prof. Philip S. Yu.

He is currently an Associate Professor at Beijing University of Posts and Telecommunications, Beijing, China. 
He has published over 60 research articles in top conferences and journals, e.g., ICML, NeurIPS, AAAI, IJCAI, KDD, WWW, SIGIR,  TOIS, TKDE, TIST and TWEB.
His research interests include machine learning and data mining with special attention to Riemannian geometry. 
\end{IEEEbiography}

\vspace{6 pt}

\begin{IEEEbiography} [{\includegraphics[width=1in,height=1.25in,clip,keepaspectratio]{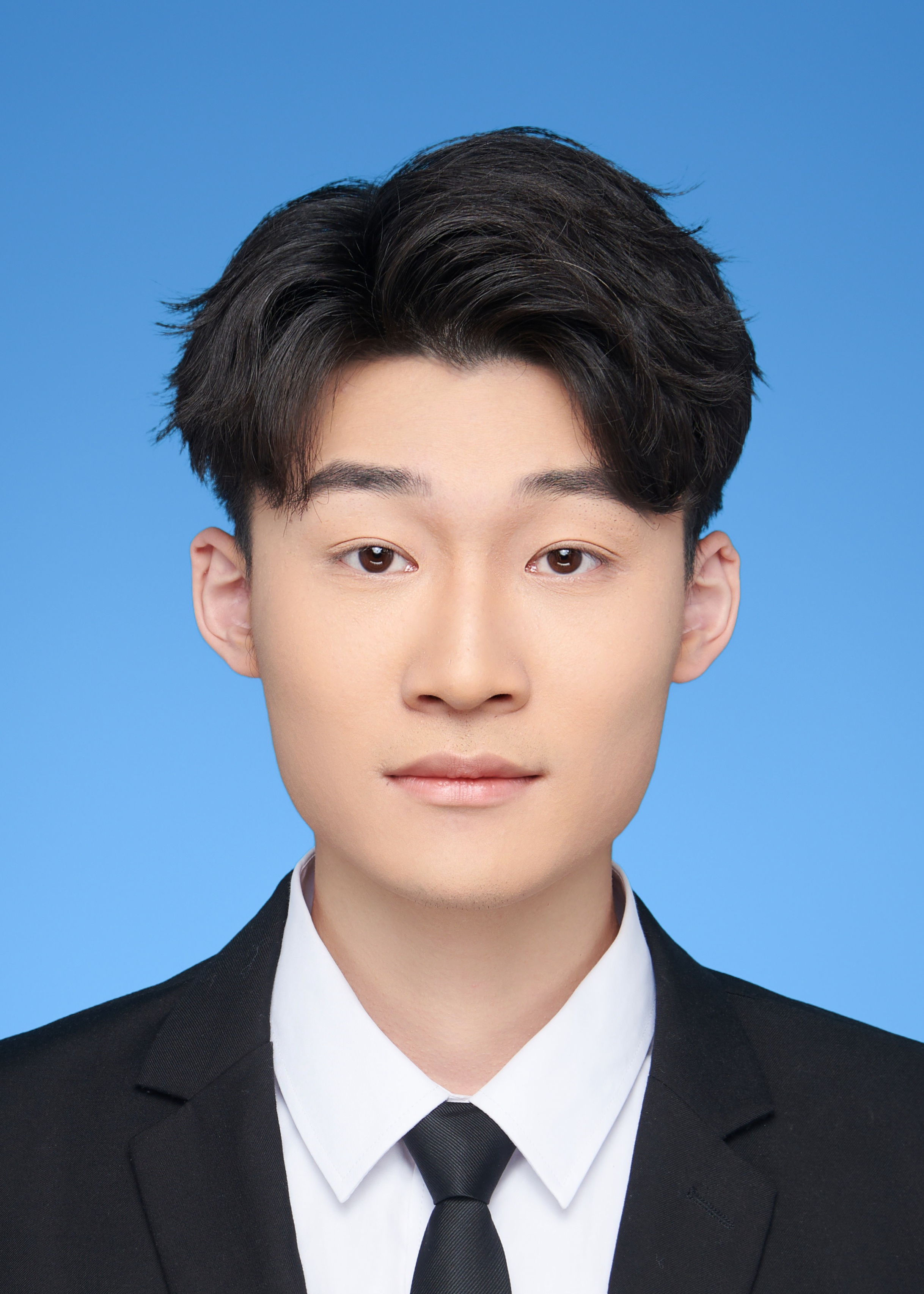}}]{Zhenhao Huang}
is now pursuing the Master degree of computer science at  North China Electric Power University under the supervision of Dr. Li Sun.
His research interests lie in machine learning on graphs. 
He has authored or co-authored in ICDM23, AAAI24, NeurIPS24 and ICML24.
\end{IEEEbiography}

\vspace{6 pt}

\begin{IEEEbiography}[{\includegraphics[width=1in,height=1.25in,clip,keepaspectratio]{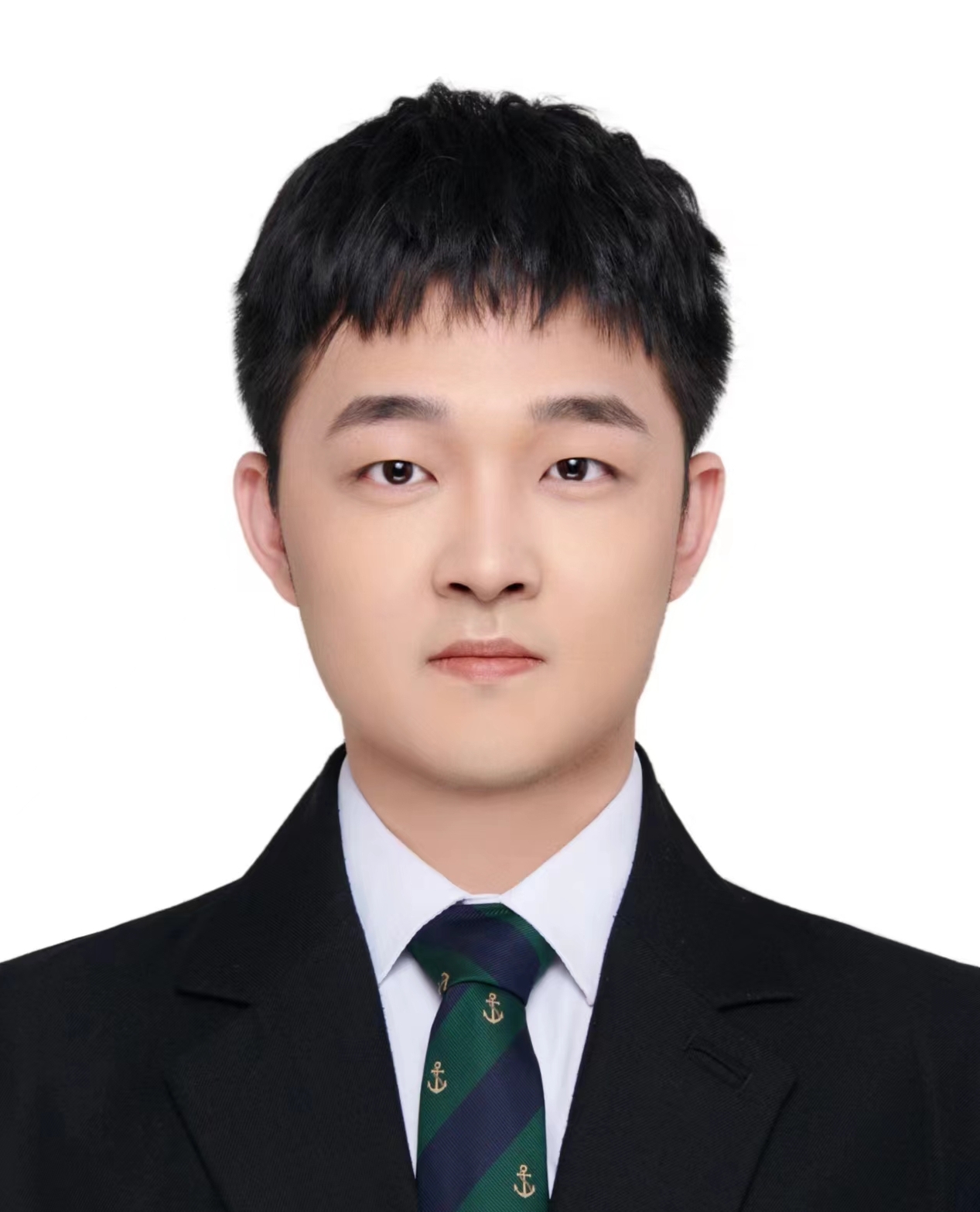}}] {Yujie Wang}
is now pursuing the Master degree of computer science at Beijing University of Posts and Telecommunications.
His research interests lie in graph mining, and he has authored in CIKM24.
\end{IEEEbiography}

\vspace{6 pt}

\begin{IEEEbiography}[{\includegraphics[width=1in,height=1.25in,clip,keepaspectratio]{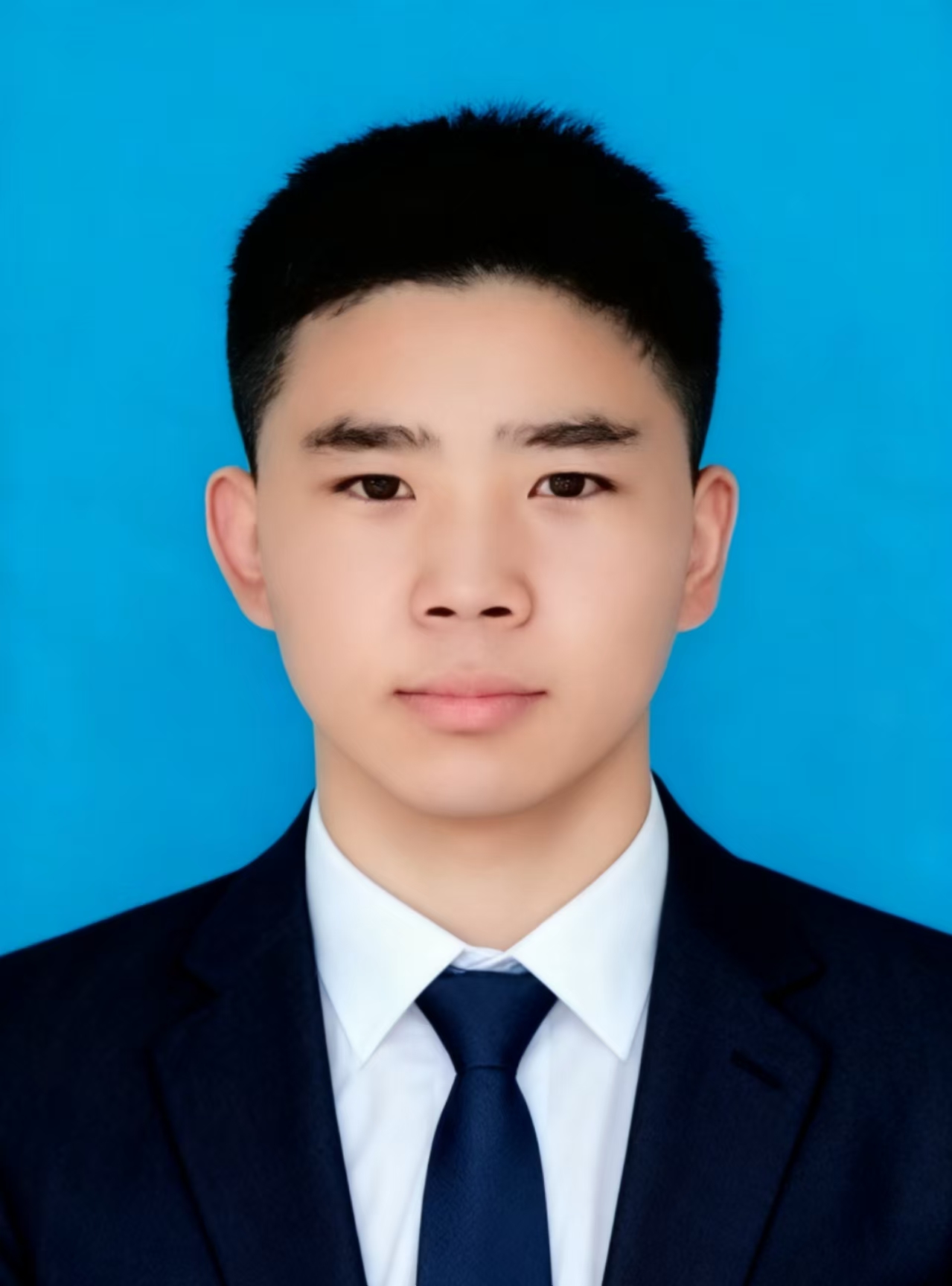}}]{Hongbo Lv}
is now pursuing the Master degree of computer science at  North China Electric Power University.
His research interests lie in function learning. 
\end{IEEEbiography}

\vspace{6 pt}

\begin{IEEEbiography}[{\includegraphics[width=1.4in,height=1.2in,clip,keepaspectratio]{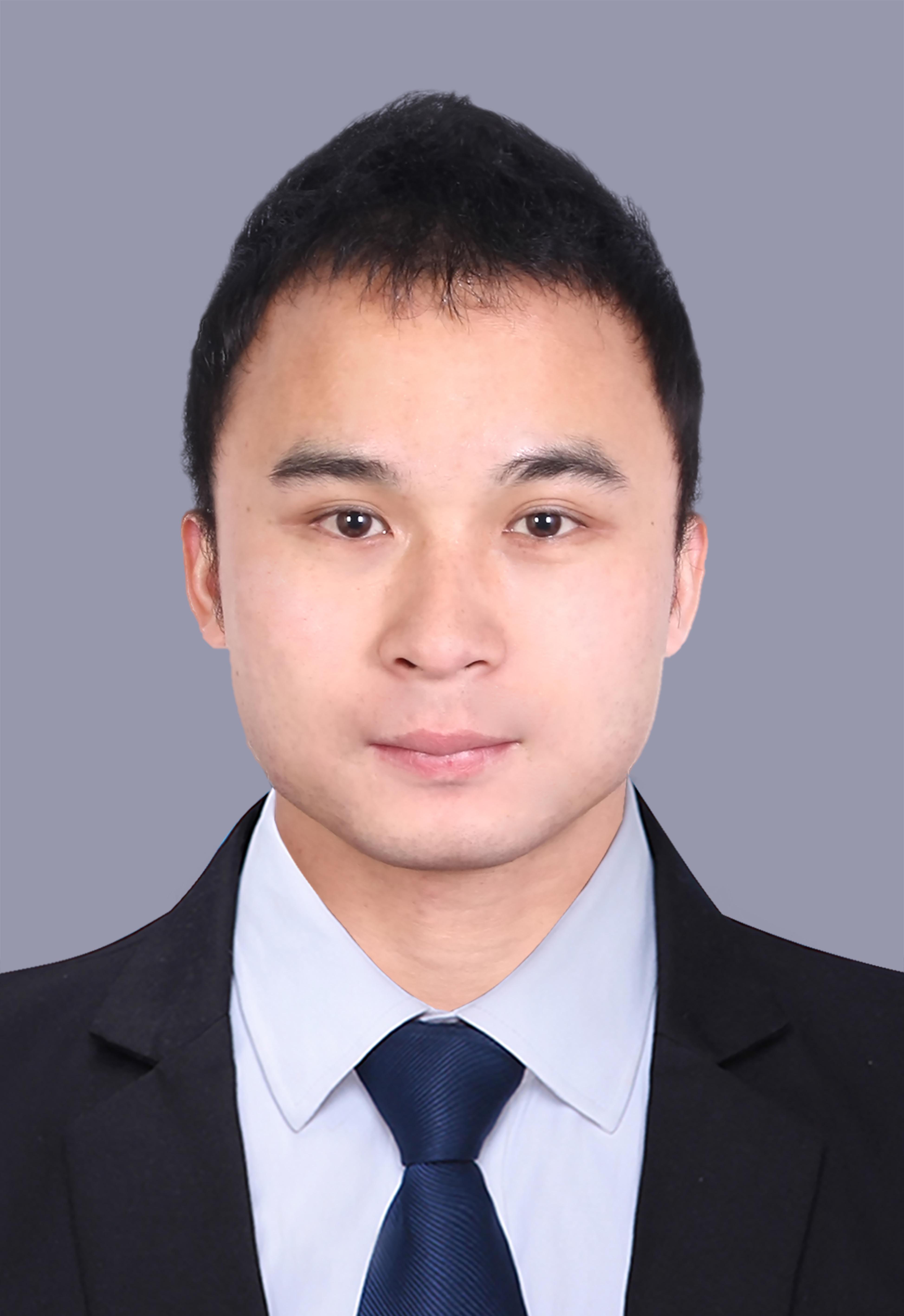}}] {Hao Peng}
is currently the full professor at School of Cyber Science and Technology, Beihang University, Beijing, China.
His research interests include data mining and machine learning.
He has published over 100 research papers in top-tier journals and conferences, e.g., TPAMI, TKDE, TC, TOIS, TKDD, WWW, SIGIR and AAAI. He is the Associate Editor of International Journal of Machine Learning and Cybernetics.
\end{IEEEbiography}

\vspace{6 pt}

\begin{IEEEbiography}[{\includegraphics[width=1in,height=1.25in,clip,keepaspectratio]{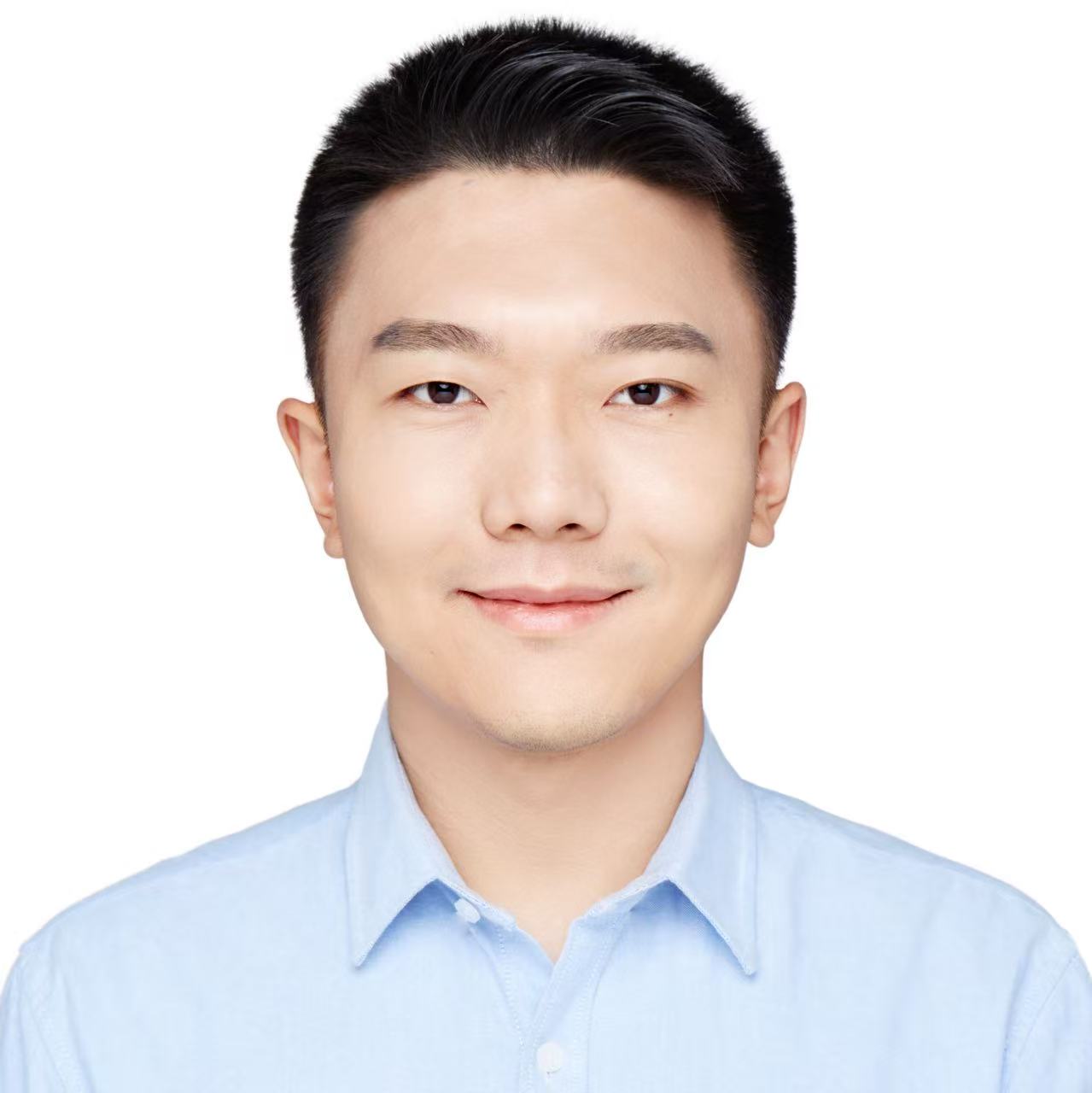}}] {Chunyang Liu}
received the Ph.D. degree from the University of Technology, Sydney, Australia. He is currently a senior algorithm engineer in DiDi Chuxing, China. His main research interests include machine learning, urban computing and operation research.
\end{IEEEbiography}

\vspace{6 pt}

\begin{IEEEbiography}[{\includegraphics[width=1in,height=1.5in,clip,keepaspectratio]{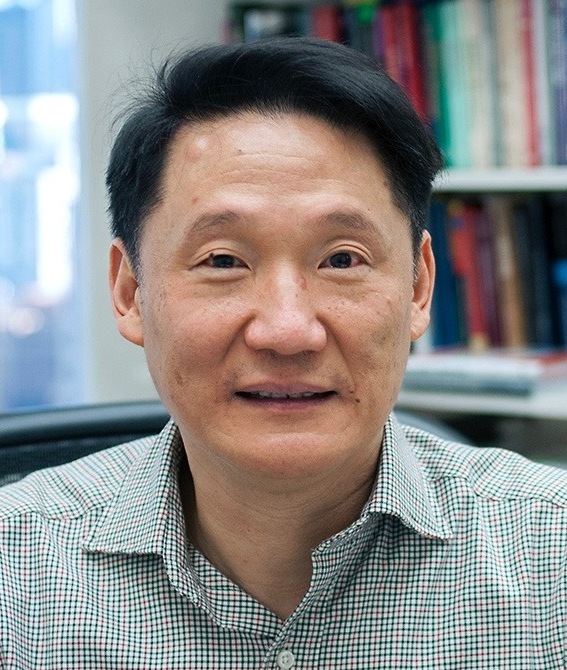}}]{Philip S. Yu (Life Fellow, IEEE)}
received the Ph.D. degree in Electrical Engineering from Stanford University, USA.

He is a Fellow of the ACM and the IEEE. He is currently a Distinguished Professor of computer science with the University of Illinois at Chicago (UIC), and also holds the Wexler Chair in information technology. 
He has published more than 1,600 papers in refereed journals and conferences cited over 218,000 times with an H-index of 204. 
He received the ACM SIGKDD's 2016 Innovation Award, the IEEE Computer Society’s 2013 Technical Achievement Award, the IEEE ICDM’s 2003 Research Contributions Award, VLDB 2022 Test of Time Award, ACM SIGSPATIAL 2021 10-year Impact Award, the EDBT 2014 Test of Time Award (2014), and the ICDM 2013 10-year Highest-Impact Paper Award. 
He was the Editor-in-Chief of ACM Transactions on Knowledge Discovery from Data (2011-2017) and IEEE Transactions on Knowledge and Data Engineering (2001-2004).
\end{IEEEbiography}

% =================================================================
% PART 2: Technical Appendix
% =================================================================

\clearpage

% --- CRITICAL FIX 2: Restore TOC writing and use \twocolumn for layout ---
\let\addcontentsline\oldaddcontentsline 

\twocolumn[
  \begin{center}
    \vspace*{2em}
    {\huge \texttt{ASIL}: Augmented Structural Information Learning for Deep Graph Clustering in Hyperbolic Space \\[0.3em] (\textbf{Technical Appendix}) \par}
    \vspace{1.5em}
    {\large Li Sun, Zhenhao Huang, Yujie Wang, Hongbo Lv, Hao Peng, Chunyang Liu, Philip S. Yu \par}
    \vspace{3em}
  \end{center}
]

% Reset counters for Appendix
\setcounter{section}{0}
\setcounter{equation}{0}
\setcounter{figure}{0}
\setcounter{table}{0}
\setcounter{footnote}{0}

% Now Table of Contents only sees what follows
\tableofcontents

\section{Notations}
To be self-contained, we first collect the important notations frequently used in this Appendix as follows.
\vspace{0.2in}
\begin{table}[h]
\caption{Notation Table.}
  \centering
  \begin{tabular}{|c|c|}
    \hline
    \textbf{Notation} & \textbf{Description} \\
    \hline
    $G$ & A weighted graph  \\
    $\mathcal{V}, \mathcal{E}$ & Graph nodes set and edges set, respectively\\
    $\mathbf{X}$ & Node attributes matrix \\
    $\mathbf{A}$ & Graph adjacency matrix \\
    $N$ & Graph node number \\
    $w$ & Edge weight function \\
    $v_i$ & A graph node \\
    $d_i$ & The degree of graph node $v_i$ \\
    $\operatorname{Vol}(\cdot)$ & The volume of graph or subset of graph \\
    $\mathcal{U}$ & A subset of $\mathcal{V}$ \\
    $K$ & Number of clusters \\
    $\mathbb{L}^{\kappa, d}$ & $d$-dimensional Lorentz model with curvature $\kappa$ \\
    $\langle \cdot, \cdot \rangle_\mathbb{L}$ & Lorentz inner product \\
    $\boldsymbol x$ & Vector or point on manifold \\
    $d_\mathbb{L}(\cdot, \cdot)$ & Lorentz distance \\
    $\lVert \cdot \rVert_\mathbb{L}$ & Lorentz norm \\
    $H$ & Dimension of structural information \\
    $\mathcal{T}$ & A partitioning tree \\
       $\mathcal{T}^\star$ & The optimal partitioning tree \\
    $\alpha$ & A Non-root node of $\mathcal{T}$ \\
    $\alpha ^ -$ & The immediate predecessor of $\alpha$ \\
    $\lambda$ & Root node of $\mathcal{T}$ \\
    $T_\alpha$ & The module of $\alpha$, a subset of $\mathcal{V}$ associated with $\alpha$ \\
    $g_\alpha$ & The total weights of graph edges \\
                &with exactly one endpoint in module $T_\alpha$ \\
    $\mathcal{H}^\mathcal{T}(G)$ & Structural information of $G$ w.r.t. $\mathcal{T}$ \\
    $\mathcal{H}^\mathcal{T}(G;\alpha)$ & Structural information of tree node $\alpha$ \\
    $\mathcal{H}^H(G)$ & $H$-dimensional structural entropy of $G$ \\
    $\mathcal{H}^\mathcal{T}(G;h)$ & Structural information of all tree node at level $h$ \\ 
    $\mathbf{C}^h$ & The assignment matrix between $h$ and $h-1$ level \\
    $N_h$ & The maximal node numbers in $h$-level of $\mathcal{T}$ \\
    $V^h_k$ & The volume of graph node sets $T_k$ \\
    $\Phi(G)$ & Graph conductance \\
    $\mathbf{Z}$ & Tree node embeddings matrix \\
    $\mathbf{S}$ & The clustering matrix \\
    $\mathcal{T}_{\operatorname{net}}$ & The partitioning tree constructed by LSEnet\\
    $\mathcal{H}^{\mathcal{T}_{\operatorname{net}}}(G;\mathbf{Z};\mathbf{\Theta})$ & The structural information of  deep partitioning tree\\
   $\mathbf{\Theta}$ & The parameters of neural model \\
   $\tilde{\mathbf{A}}$ & The virtual adjacency constructed from node
   \\               & representations  $\tilde{\mathbf{A}}_{ij} = \exp(-d_{\mathbb{L}}(L(\boldsymbol{z}_i), L(\boldsymbol{z}_j))/\tau)$ \\
   $\gamma$ & The balancing factor in $[0, 1]$ for graph fusion \\
   $\mathbf{A}^{\gamma}$ & The fused adjacency matrix $A^{\gamma}=(1-\gamma)\mathbf{A} + \gamma \tilde{\mathbf{A}}$ \\
   $k$ & The number of $k$ nearest neighbors for sparsifying $\tilde{\mathbf{A}}$ \\
    \hline
  \end{tabular}
  \label{tab:notation}
\end{table}

\newpage

\section{Theorem \& Proofs}
% Here, we detail lemmas and theorems on the proposed \textbf{Differential Structural Information}. In particular, we prove the \emph{equivalence, additivity, flexibility, bound} and \emph{the connection to graph clustering}, and show the supporting lemmas. 

This section elaborates on the theoretical results given this paper. 
We prove the theorems on the proposed differential structural information, deep partitioning tree and Lorentz tree contrastive learning. 

\vspace{-0.03in}
%, in which we give the lemmas not shown in the paper due to the limit of space.
%(Theorems in hyperbolic space are elaborated in the next section*.)

\subsection{Equivalence on Continuous Formalism}
\label{proof.equivalence}
We demonstrate the equivalence between the proposed differential formulation and the structural information proposed in the original paper \cite{li2016structural}.
\begin{mymath}
\noindent\textbf{Theorem III.3} \textbf{(Equivalence).}
\emph{The formula of level-wise assignment is equivalent to $\mathcal{H}^\mathcal{T}(G) = \sum_{\alpha \in T, \alpha \neq \lambda} \mathcal{H}^\mathcal{T}(G; \alpha)$ in \cite{li2016structural}.}
\end{mymath}
\begin{proof}
From $\mathcal{H}^\mathcal{T}(G; \alpha) = -\frac{g_{\alpha}}{\text{Vol}(G)} \log_2 \left( \frac{V_{\alpha}}{V_{\alpha^-}} \right)$, we can rewrite structural information of $G$ w.r.t all nodes of $\mathcal{T}$ at height $h$ as follows:
\begin{align}
\label{eq.si_at_k}
    \mathcal{H}^{\mathcal{T}}(G;h) &= -\frac{1}{V}\sum_{k=1}^{N_h}g^h_k \log_2 \frac{V_k^h}{V^{h-1}_{k^-}} \nonumber  \\
    &=-\frac{1}{V}[\sum_{k=1}^{N_h}g^h_k \log_2 V_k^h - \sum_{k=1}^{N_h}g^h_k \log_2 V^{h-1}_{k^-}].
\end{align}
Then, the $H$-dimensional structural information of $G$ can be written as
\begin{align}
    \mathcal{H}^{\mathcal{T}}(G) = -\frac{1}{V}[\sum_{h=1}^{H}\sum_{k=1}^{N_h}g^h_k \log_2 V^h_k - \sum_{h=1}^{H}\sum_{k=1}^{N_h}g^h_k \log_2 V^{h-1}_{k^-}].
\end{align}
At height $h$,
\begin{align}
    g^h_k &= V^h_k - \sum_{i=1}^N \sum_{j \in \mathcal{N}(i)}\mathbf{I}(i\in T_{\alpha^h_k})\mathbf{I}(j\in T_{\alpha^h_k})w_{ij} \nonumber \\
    &=V^h_k - \sum_{i=1}^N \sum_{j \in \mathcal{N}(i)}S^h_{ik}S^h_{jk}w_{ij},
\end{align}
where we use $S^h_{ik}$ to represent $\mathbf{I}(i\in T_{\alpha^h_k})$, and $T_{\alpha^h_k}$ is the subset of $\mathcal{V}$ corresponds the $k$-th node in height $h$.
\begin{align}
    V^h_k &= \sum_{i=1}^N\mathbf{I}(i\in T_{\alpha^h_k})d_i = \sum_{i=1}^N S^h_{ik} d_i.
\end{align}
Consequently, there exists
\begin{align}
    \begin{split}
    \sum_{h=1}^H\sum_{k=1}^{N_h}g^h_k \log_2 V^h_k &= \sum_{h=1}^H\sum_{k=1}^{N_h}V^h_k \log_2 V^h_k\\& - \sum_{h=1}^H\sum_{k=1}^{N_h}(\log_2 V^h_k) \sum_{i=1}^N \sum_{j \in \mathcal{N}(i)} S^h_{ik}S^h_{jk} w_{ij}.
    \end{split}
\end{align}
Similarly, we have the following equation,
\begin{align}
    \begin{split}
    \sum_{h=1}^H\sum_{k=1}^{N_h}g^h_k \log_2 V^{h-1}_{k^-}&= \sum_{h=1}^H\sum_{k=1}^{N_h}V^h_k \log_2 V^{h-1}_{k^-}\\ &-\sum_{h=1}^H\sum_{k=1}^{N_h}(\log_2 V^{h-1}_{k^-})\\
    &\cdot\sum_{i=1}^N \sum_{j \in \mathcal{N}(i)} S^h_{ik}S^h_{jk}w_{ij}.
    \end{split}
\end{align}
To give the ancestor of node $k$ in height $h-1$, we utilize the assignment matrix $C^h$ as follows:
\begin{align}
    V^{h-1}_{k^-} = \sum_{k'=1}^{N_{h-1}}\mathbf{C}^h_{k{k'}}V^{h-1}_{k'}.
\end{align}
We can  verify the following holds,
\begin{align}
\begin{split}
S^h_{ik} &= \mathbf{I}(i \in T_{\alpha^h_k}) \\
&= \sum_{\alpha^{h-1}_k}^{N_{H-1}} \sum_{\alpha^h_k}^{N_h} \cdots \sum_{\alpha^{h+1}_k}^{N_{h+1}} \mathbf{I}(\{i\} \subseteq T_{\alpha^{H-1}_k}) \\
&\cdot \mathbf{I}(T_{\alpha^{H-1}_k} \subseteq T_{\alpha^{H-2}_k}) \cdots \mathbf{I}(T_{\alpha^{h+1}_k} \subseteq T_{\alpha^h_k}) \\
&= \sum_{j_{H-1}}^{N_{H-1}} \cdots \sum_{j_{h+1}}^{N_{h+1}} \mathbf{C}^H_{ij_{H-1}} \mathbf{C}^{H-1}_{j_{H-1}j_{H-2}} \cdots \mathbf{C}^{h+1}_{j_{h+1}k}.
\end{split}
\end{align}
Since $\mathbf{C}^{H+1}=\mathbf{I}_N$, we have $\mathbf{S}^h = \prod_{h=H+1}^{h+1}\mathbf{C}^h$.
Utilizing the equations above, we have completed the proof. 
\end{proof}

\subsection{Bound on Deep Partitioning Tree}
\label{proof.theoremapprox}

We show that the proposed deep partitioning tree well approximates the optimal partitioning tree given by  \cite{li2016structural}.
\vspace{0.05in}
\begin{mymath}
\noindent\textbf{Theorem III.6} \textbf{(Bound).}
\emph{ For any graph $G$, $\mathcal{T}^*$ is the optimal partitioning tree given by  \cite{li2016structural},
 and $\mathcal{T}_{\operatorname{net}}^*$ with height $H$ is the partitioning tree given from $ Z^*, \Theta^* = \arg_{Z,\Theta}\min \mathcal{H}_{T^{net}}(G; Z; \Theta) $.
For any pair of leaf embeddings $\boldsymbol z_i$ and $\boldsymbol z_j$ of $\mathcal{T}_{\operatorname{net}}^*$, 
 there exists bounded real functions $\{f_h\}$ and constant $c$,
 such that for any $0 < \epsilon < 1$, $\tau \leq \mathcal{O}(1/ \ln[(1-\epsilon)/\epsilon])$ and  $\frac{1}{1 + \exp\{-(f_h(\boldsymbol z_i, \boldsymbol z_j) - c) / \tau\}} \geq 1 - \epsilon$ satisfying $i\overset{h}{\sim} j$, we have
 } 
    \begin{align}
        \lvert \mathcal{H}^{\mathcal{T}^*}(G) - \mathcal{H}^{\mathcal{T}_{\operatorname{net}}^*}(G) \rvert \leq \mathcal{O}(\epsilon).
    \end{align}
\end{mymath}

First, we prove the existence of $\{f_h\}$ by giving a construction process. Fix $0 < \epsilon < 1$, we have
\begin{align}
    \sigma^h_{ij}&=\frac{1}{1 + \exp\{-(f_h(\boldsymbol z_i, \boldsymbol z_j) - c) / \tau\}} \geq 1 - \epsilon \\
    &\Rightarrow f_h(\boldsymbol z_i, \boldsymbol z_j) \geq c + \tau \ln \frac{1-\epsilon}{\epsilon} \leq 1 + c ,\label{fc}
\end{align}
since $\tau \leq \mathcal{O}(1/ \ln[(1-\epsilon)/\epsilon])$.
Similarly, when $i\overset{h}{\sim} j$ fails, we have $\sigma^h_{ij}\leq \epsilon \Rightarrow f_h(\boldsymbol z_i, \boldsymbol z_j) \leq c - \tau \ln \frac{1-\epsilon}{\epsilon} \geq c - 1$. Without loss of generality, we let $c=0$ and find that if $i\overset{h}{\sim} j$ holds, $f_h \geq 1$, otherwise, $f_h \leq -1$. If we rewrite $f_h$ as a compose of scalar function $f_h = f(g_h(\boldsymbol z_i, \boldsymbol z_j)) + u(g_h(\boldsymbol z_i, \boldsymbol z_j)): \mathbb{R}\rightarrow \mathbb{R}$, where $g_h$ is a scalar function and $u$ is a step function as $u(x)=
\begin{cases}
1 & x \geq 0\\
-1 & x < 0
\end{cases}$. If $f$ has a discontinuity at $0$ then the construction is easy. At level $h$, we find a node embedding $\boldsymbol a^h$ that is closest to the root $\boldsymbol z_o$ and denote the lowest common ancestor of $\boldsymbol z_i$ and $\boldsymbol z_j$ in a level less than $h$ as $\boldsymbol a^h_{ij}$. Then set $g_h(\boldsymbol z_i, \boldsymbol z_j) = d_\mathbb{L}(\boldsymbol z_o, \boldsymbol a^h_{ij}) - d_\mathbb{L}(\boldsymbol z_o, \boldsymbol a^h)$. Clearly $g_h(\boldsymbol z_i, \boldsymbol z_j)  \geq 0$ if $i\overset{h}{\sim} j$ holds. Then we set $f(\cdot)=\sinh(\cdot)$, i.e. $f_h(\boldsymbol z_i, \boldsymbol z_j) = \sinh(g_h(\boldsymbol z_i, \boldsymbol z_j)) + u(g_h(\boldsymbol z_i, \boldsymbol z_j))$.

If $i\overset{h}{\sim} j$ holds, clearly $f_h \geq 1$, otherwise, $f_h \leq -1$. Since the distance between all pairs of nodes is bounded, $g_h$ is also bounded. All the properties are verified, so such $\{f_h\}$ exist.

Then we have the following lemma.
% \begin{lemma}
% \label{lemma.kse2}

\vspace{0.05in}
\noindent\textbf{Lemma III.7} 
    For a weighted Graph $G=(V, E)$ with a weight function $w$, and a partitioning tree $\mathcal{T}$ of $G$ with height $H$, we can rewrite the formula of $H$-dimensional structural information of $G$ w.r.t $\mathcal{T}$ as follows:
    \begin{align}
        \begin{split}
        \label{kse2}
        \mathcal{H}^\mathcal{T}(G)=&-\frac{1}{V} 
            \sum_{h=1}^H\sum_{i=1}^N (d_i-\sum_{j\in \mathcal{N}(i)}\mathbf{I}(i\overset{h}{\sim} j)w_{ij})\\
            &\cdot \log_2(\frac{\sum_l^N\mathbf{I}(i \overset{h}{\sim} l)d_l}{\sum_l^N\mathbf{I}(i \overset{h-1}{\sim} l)d_l}),
        \end{split}
    \end{align}
    where $\mathbf{I}(\cdot)$ is the indicator function. 
% \end{lemma}
\begin{proof}
    We leave the proof of Lemma III.7 in Appendix \ref{proof.kse2}.
\end{proof}
% \begin{lemma}
% \label{lemma:eq}
%     $\mathbf{I}(i \overset{h}{\sim} j)$ can be represent related to the assignment matrix $\mathbf{C}^h$ as
%     \begin{align}
%         \mathbf{I}(i \overset{h}{\sim} j) = \sum_{k=1}^{N_h}s^h_{ik}s^h_{jk},
%     \end{align}
%     where $s^h_{ik}$ is from Equation \ref{eq.s}.
% \end{lemma}
% \begin{proof}
%     At height $h$ of tree $\mathcal{T}$, $\mathbf{I}(i \overset{h}{\sim} j) = \sum_{k=1}^{N_h}\mathbf{I}(i\in T_{\alpha^h_k})\mathbf{I}(j\in T_{\alpha^h_k})=\sum_{k=1}^{N_h}s^h_{ik}s^h_{jk}$, complete the proof.
% \end{proof}
Now we start our proof of Theorem III.6.
\begin{proof}
We then focus on the absolute difference between $H^{\mathcal{T}^*}(G) $ and $\mathcal{H}^{\mathcal{T}^*_{\operatorname{net}}}(G)$
\begin{align}
\label{eq.ineq}
    \lvert \mathcal{H}^{\mathcal{T}^*}(G) - \mathcal{H}^{\mathcal{T}^*_{\operatorname{net}}}(G) \rvert &\leq \lvert \mathcal{H}^{\mathcal{T}^*}(G) - \mathcal{H}^{\mathcal{T}_{\operatorname{net}}}(G; \mathbf{Z}^*; \Theta^*) \rvert \nonumber\\ & + \lvert \mathcal{H}^{\mathcal{T}_{\operatorname{net}}}(G; \mathbf{Z}^*; \Theta^*) - \mathcal{H}^{\mathcal{T}^*_{\operatorname{net}}}(G) \rvert \nonumber \\
    &\leq 2 \sup_{\mathbf{Z}, \mathbf{\Theta}}\lvert \mathcal{H}^{\mathcal{T}}(G) - \mathcal{H}^{\mathcal{T}_{\operatorname{net}}}(G; \mathbf{Z}; \Theta) \rvert,
\end{align}
where $\mathcal{T}=\operatorname{decode}(\mathbf{Z})$.
Recall the Eq. (\ref{kse2}), we can divide them into four corresponding parts, denoted as $A$, $B$, $C$, and $D$ respectively. Then, we have
\begin{align}
    \lvert \mathcal{H}^{\mathcal{T}}(G) - \mathcal{H}^{\mathcal{T}_{\operatorname{net}}}(G; \mathbf{Z}; \Theta) \rvert &= -\frac{1}{\operatorname{Vol}(G)}[A + B + C + D] \nonumber \\
    &\leq -\frac{1}{\operatorname{Vol}(G)} [\lvert A \rvert + \lvert B \rvert + \lvert C \rvert + \lvert D \rvert].
\end{align}
For part $A$, give fixed $i$ and $k$, we assume $i \overset{h}{\sim} l$ holds for some $l_{1},...,l_{m_i}$.
\begin{align}
    \lvert A \rvert &= \lvert \sum_{h=1}^H\sum_i^N d_i \log_2(\sum_l^N \mathbf{I}(i \overset{h}{\sim} l) d_l)-\sum_{h=1}^H\sum_i^N d_i \log_2(\sum_l^N \sigma_{il}^h d_l) \rvert \nonumber \\
    &\leq \lvert \sum_{h=1}^H\sum_i^N d_i\log_2 \frac{\sum_{c=1}^{m_i} d_{l_c}}{\sum_{c=1}^{m_i} (1-\epsilon)d_{l_c} + \sum_{\overset{-}{c}}\epsilon d_{l_{\overset{-}{c}}}} \rvert \nonumber \\
    &\leq \lvert \sum_{h=1}^H\sum_i^N d_i\log_2 \frac{\sum_{c=1}^{m_i} d_{l_c}}{\sum_{c=1}^{m_i}(1-\epsilon)d_{l_c}} \rvert \nonumber \\
    &=\lvert \sum_{h=1}^H\sum_i^N d_i \log_2 \frac{1}{1-\epsilon} \rvert \nonumber \\
    &= H\operatorname{Vol}(G)\log_2 \frac{1}{1-\epsilon}.
\end{align}
For part B, we still follow the assumption above and assume $i \overset{h}{\sim} j$ holds for some $j_e,...,j_{n_i}$ in the neighborhood of $i$.
\begin{align}
    \lvert B \rvert &= \lvert-\sum_{h=1}^H\sum_i^N[\log_2(\sum_l^N\mathbf{I}(i\overset{h}{\sim} l)d_l)\cdot \sum_{j\in \mathcal{N}(i)}\mathbf{I}(i\overset{h}{\sim} j)w_{ij}] \nonumber\\ &+\sum_{h=1}^H\sum_i^N[\log_2(\sum_l^N\sigma^h_{il}d_l)\cdot \sum_{j\in \mathcal{N}(i)}\sigma^h_{ij}w_{ij} ] \rvert \nonumber \\
    &\leq \lvert\sum_{h=1}^H\sum_i^N\log_2(\sum_{c=1}^{m_i} d_{l_c})\cdot \sum_{e=1}^{n_i}w_{ij_e} -\log_2(\sum_{c=1}^{m_i}(1-\epsilon)d_{l_c} \nonumber\\ &+\sum_{\overset{-}{c}}\epsilon d_{l_{\overset{-}{c}}})\cdot [\sum_{e=1}^{n_i}(1-\epsilon)w_{ij_e} + \sum_{\overset{-}{e}}\epsilon w_{ij_{\overset{-}{e}}}] \rvert \nonumber \\
    & \leq \lvert\sum_{h=1}^H\sum_i^N\log_2(\sum_{c=1}^{m_i} d_{l_c})\cdot \sum_{e=1}^{n_i}w_{ij_e} \nonumber\\ & -\log_2(\sum_{c=1}^{m_i}(1-\epsilon)d_{l_c}) \sum_{e=1}^{n_i}(1-\epsilon)w_{ij_e} \rvert \nonumber \\
    &=\lvert \sum_{h=1}^H\sum_i^N (\sum_{e=1}^{n_i}w_{ij_e})[\log_2(\sum_{c=1}^{m_i} d_{l_c} - (1-\epsilon)\log_2(1-\epsilon) \nonumber\\ &- (1-\epsilon)\log_2(\sum_{c=1}^{m_i} d_{l_c}]  \rvert 
    \nonumber \\
    &= \lvert \sum_{h=1}^H\sum_i^N (\sum_{e=1}^{n_i}w_{ij_e})[\epsilon \log_2(\sum_{c=1}^{m_i} d_{l_c} - (1-\epsilon)\log_2(1-\epsilon)]  \rvert 
    \nonumber \\
    &\leq (1-\epsilon)\log_2 \frac{1}{1-\epsilon}\sum_{h=1}^H\sum_i^N\sum_{e=1}^{n_i}w_{ij_e}\nonumber \\
    &\leq (1-\epsilon)\log_2 \frac{1}{1-\epsilon}\sum_{h=1}^H\sum_i^N d_i\nonumber \\
    &=(1-\epsilon)H V\log_2 \frac{1}{1-\epsilon} .
\end{align}
Similarly, we can get the same results for parts $C$ and $D$ as $A$ and $B$ respectively. Then,
\begin{align}
    \lvert \mathcal{H}^{\mathcal{T}}(G) - \mathcal{H}^{\mathcal{T}_{\operatorname{net}}}(G; \mathbf{Z}; \Theta) \rvert &\leq 2H(2-\epsilon)\log_2\frac{1}{1-\epsilon}.
\end{align}
Substituting into Equation.\ref{eq.ineq}, we obtain
\begin{align}
    \lvert \mathcal{H}^{\mathcal{T}^*}(G) - \mathcal{H}^{\mathcal{T}^*_{\operatorname{net}}}(G) \rvert &\leq 4H(2-\epsilon)\log_2\frac{1}{1-\epsilon}.
\end{align}
Since the limitation is a constant number,
\begin{align}
    \lim_{\epsilon \rightarrow 0} \frac{(2-\epsilon)\log_2\frac{1}{1-\epsilon}}{\epsilon}=2,
\end{align}
we have $\lvert H^{\mathcal{T}^*}(G) - H^{\mathcal{T}^*_{\operatorname{net}}}(G) \rvert \leq \mathcal{O}(\epsilon)$, which completes the proof. 
\end{proof}

\subsection{Additivity of Structural Entropy}
\label{proof.add}

We elaborate on the additivity of structural entropy to support the theorem in the next part.
\vspace{0.05in}
\begin{mymath}
\noindent\textbf{Lemma III.8} \textbf{(Additive).}
\emph{
    The one-dimensional structural entropy of $G$ can be decomposed as follows
    \begin{align}
        \mathcal{H}^1(G) = \sum_{h=1}^H\sum_{j=1}^{N_{h-1}}\frac{V^{h-1}_j}{V}E([\frac{C^h_{kj}V^h_k}{V^{h-1}_j}]_{k=1,...,N_h}),
    \end{align}
    where 
      $  E(p_1, ..., p_n) = -\sum_{i=1}^n p_i\log_2 p_i$
 is the entropy. 
}
\end{mymath}
\begin{proof}
To verify the equivalence to the one-dimensional structural entropy, we expand the above formula.
\begin{align}
    \begin{split}
        \mathcal{H}(G)&=-\sum_{h=1}^H\sum_{j=1}^{N_{h-1}}\frac{V^{h-1}_j}{V}\sum_{k=1}^{N_h}\frac{C^h_{kj}V^h_k}{V^{h-1}_j}\log_2 \frac{C^h_{kj}V^h_k}{V^{h-1}_j} \\
    &= -\frac{1}{V}\sum_{h=1}^H \sum_{k=1}^{N_h}\sum_{j=1}^{N_{h-1}} C^h_{kj}V^h_k \log_2 \frac{C^h_{kj}V^h_k}{V^{h-1}_j}\\
    &=-\frac{1}{V}\sum_{h=1}^H \sum_{k=1}^{N_h}V^h_k \log_2\frac{V^h_k}{V^{h-1}_{k^-}}.
    \end{split}
\end{align}
We omit some $j$ that make $C^h_{kj}=0$ so that the term $V^{h-1}_j$ only exists in the terms where $C^h_{kj}=1$, meaning that $V^{h-1}_j$ exists if and only if $V^{h-1}_j=V^{h-1}_{k^-}$ since we summing over all $k$. Continue the process that
\begin{align}
    \begin{split}
        \mathcal{H}(G) &= -\frac{1}{V}\sum_{h=1}^H \sum_{k=1}^{N_h}V^h_k \log_2\frac{V^h_k}{V} + \frac{1}{V}\sum_{h=1}^H \sum_{k=1}^{N_h}V^h_k \log_2\frac{V^{h-1}_{k^-}}{V} \\
    &=-\frac{1}{V}\sum_{i=1}^N d_i\log_2\frac{d_i}{V} -\frac{1}{V}\sum_{h=1}^{H-1} \sum_{k=1}^{N_h}V^h_k \log_2\frac{V^h_k}{V} \\ &+ \frac{1}{V}\sum_{h=1}^H \sum_{k=1}^{N_h}V^h_k \log_2\frac{V^{h-1}_{k^-}}{V} \\
    &=-\frac{1}{V}\sum_{i=1}^N d_i\log_2\frac{d_i}{V} -\frac{1}{V}\sum_{h=1}^{H-1} \sum_{k=1}^{N_h}V^h_k \log_2\frac{V^h_k}{V} \\ &+ \frac{1}{V}\sum_{h=2}^H \sum_{k=1}^{N_h}V^h_k \log_2\frac{V^{h-1}_{k^-}}{V}.
    \end{split}
\end{align}
From the first line to the second line, we separate from the term when $h=H$. For the third line, we eliminate the summation term in $h=1$, since when $h=1$, in the last term, $\log_2\frac{V^{0}_{k^-}}{V}=\log_2\frac{V}{V}=0$.

Then let us respectively denote $-\sum_{h=1}^{H-1} \sum_{k=1}^{N_h}V^h_k \log_2\frac{V^h_k}{V}$ and $\sum_{h=2}^H \sum_{k=1}^{N_h}V^h_k \log_2\frac{V^{h-1}_{k^-}}{V}$ as $A$ and $B$, using the trick about $C^h_{kj}$ like above, we have
\begin{align}
    \begin{split}
        A+B &= \sum_{h=2}^{H} [(\sum_{k=1}^{N_h}V^h_k \log_2\frac{V^{h-1}_{k^-}}{V}) - (\sum_{j=1}^{N_{h-1}}V^{h-1}_j \log_2\frac{V^{h-1}_j}{V})] \\
    &= \sum_{h=2}^{H}[ \sum_{k=1}^{N_h}V^h_k \log_2\frac{V^{h-1}_{k^-}}{V} - \sum_{j=1}^{N_{h-1}}\sum_{k=1}^{N_h}C^h_{kj}V^h_k \log_2\frac{V^{h-1}_j}{V})] \\
    &=\sum_{h=2}^{H} [\sum_{k=1}^{N_h}V^h_k \log_2\frac{V^{h-1}_{k^-}}{V} - \sum_{k=1}^{N_h}V^h_k \log_2\frac{V^{h-1}_{k^-}}{V}] \\
    &=0 .
    \end{split}
\end{align}
Thus 
\begin{align}
    \mathcal{H}(G)=-\frac{1}{V}\sum_{i=1}^N d_i\log_2\frac{d_i}{V},
\end{align}
which is exactly the one-dimensional structural entropy $\mathcal{H}^1(G)$ of $G$. The proof is completed.
\end{proof}

\subsection{Connection to Graph Clustering}
\label{proof.conductance}

We show that structural entropy is the upper bound of graph conductance, a typical clustering objective.
That is, structural entropy emerges as deep graph clustering objective without the knowledge of $K$.

\vspace{0.05in}
\begin{mymath}
\noindent \textbf{Theorem III.5} \textbf{(Connection to Graph Clustering).}
\emph{Given a graph $G=(\mathcal{V}, \mathcal{E})$ with  $w$, the normalized $H$-structural entropy of graph $G$ is defined as
    $\tau(G;H) = {\mathcal{H}^H(G)}/{\mathcal{H}^1(G)}$, and 
$\Phi(G)$ is the graph conductance. 
    With the additivity of DSE, the following inequality holds,
    %Then the normalized $H$-dimensional structural information of $G$ w.r.t. a partitioning tree $\mathcal{T}$ satisfies the following property:
    \begin{align}
        \tau(G; H) \geq \Phi(G).
    \end{align}
    }
\end{mymath}
\begin{proof}
The formula for the H-dimensional structural information of graph G for a partitioning tree $\mathcal{T}$ is $\mathcal{H}^\mathcal{T}(G;h)=-\frac{1}{V}\sum_{k=1}^{N_h}(V_k^h-\sum_{(i,j)\in\varepsilon}S_{ik}^h S_{jk}^h w_{ij})\log_2{\frac{V_k^h}{V_{k^-}^{h-1}}}$.
 For the conductance $\phi_{h,k}$ of $k$-th node in $\mathcal{T}$ as height $h$, following the definition of graph conductance, we have
\begin{align}
    \phi_{h,k} = \frac{\sum_{i\in T_{h,k}, j \notin T_{h,k}}w_{ij}}{\min\{ V^h_k, V-V^h_k \}} 
    = \frac{V^h_k-\sum_{(i,j)\in\mathcal{E}}s^h_{ik}s^h_{jk}w_{ij}}{V^h_k}.
\end{align}
Without loss of generality, we assume that $V^h_k \leq \frac{1}{2}V$ such that $\min\{ V^h_k, V-V^h_k \}=V^h_{k^-}$.
From $\mathcal{H}^\mathcal{T}(G;h)=-\frac{1}{V}\sum_{k=1}^{N_h}(V_k^h-\sum_{(i,j)\in\varepsilon}S_{ik}^h S_{jk}^h w_{ij})\log_2{\frac{V_k^h}{V_{k^-}^{h-1}}}$, we have
\begin{align}
    \begin{split}
        \mathcal{H}^\mathcal{T}(G) &= -\frac{1}{V}\sum_{h=1}^H\sum_{k=1}^{N_h} \phi_{h,k}V^h_k\log_2\frac{V^h_k}{V^{h-1}_{k^-}} \\
    &\geq -\Phi(G)\sum_{h=1}^H \sum_{k=1}^{N_h}\frac{V^h_k}{V}\log_2\frac{V^h_k}{V^{h-1}_{k^-}} \\
    &=-\Phi(G)\sum_{h=1}^H \sum_{k=1}^{N_h}\frac{\sum_{j=1}^{N_{h-1}}C^h_{kj}V^h_k}{V}\log_2\frac{V^h_k}{V^{h-1}_{k^-}} \\
    &=-\Phi(G)\sum_{h=1}^H \sum_{k=1}^{N_h} \sum_{j=1}^{N_{h-1}} \frac{V^{h-1}_j}{V}\frac{C^h_{kj}V^h_k}{V^{h-1}_j}\log_2\frac{V^h_k}{V^{h-1}_{k^-}} \\
    &=-\Phi(G)\sum_{h=1}^H \sum_{j=1}^{N_{h-1}} \frac{V^{h-1}_j}{V} \sum_{k=1}^{N_h} \frac{C^h_{kj}V^h_k}{V^{h-1}_j}\\ 
    &\cdot\log_2\frac{V^h_k}{\sum_{m=1}^{N_{h-1}}C^h_{km}V^{h-1}_{m}} \\
    &=-\Phi(G)\sum_{h=1}^H \sum_{j=1}^{N_{h-1}} \frac{V^{h-1}_j}{V} \sum_{k=1}^{N_h} \frac{C^h_{kj}V^h_k}{V^{h-1}_j} \\ 
    &\cdot\log_2\frac{C^h_{kj}V^h_k}{C^h_{kj}\sum_{m=1}^{N_{h-1}}C^h_{km}V^{h-1}_{m}}\\
    &=-\Phi(G)\sum_{h=1}^H \sum_{j=1}^{N_{h-1}} \frac{V^{h-1}_j}{V} \sum_{k=1}^{N_h} \frac{C^h_{kj}V^h_k}{V^{h-1}_j} \cdot \log_2\frac{C^h_{kj}V^h_k}{V^{h-1}_j} \\
    &=\Phi(G)\sum_{h=1}^H\sum_{j=1}^{N_{h-1}}\frac{V^{h-1}_j}{V}\operatorname{Ent}([\frac{C^h_{kj}V^h_k}{V^{h-1}_j}]_{k=1,...,N_h}) \\
    &=\Phi(G)\mathcal{H}^1(G).
    \end{split}
\end{align}

To get the final results, we use the inverse trick about $C^h_{kj}$ mentioned in Appendix \ref{proof.add}.
So the normalized $H$-dimensional structural information of $G$ w.r.t. $\mathcal{T}$ satisfies
\begin{align}
    \tau(G;\mathcal{T}) = \frac{\mathcal{H}^\mathcal{T}(G)}{\mathcal{H}^1(G)} \geq \Phi(G).
\end{align}
Since this inequality holds for every $H$-height partitioning tree $\mathcal{T}$ of $G$, $\tau(G;H)\geq \Phi(G)$ holds. 
\end{proof}

\subsection{Node-wise Reformulation}
\label{proof.kse2}

We show that node-wise formula is equivalent to the formula of level-wise assignment, while reducing time complexity.

\vspace{0.05in}
\begin{mymath}
\noindent \textbf{Theorem IV.1} \textbf{(Node-wise Reformulation).}
    \emph{For a weighted graph $G=(V, E)$ with a weight function $w$, and a partitioning tree $\mathcal{T}$ of $G$ with height $H$, we can rewrite the formula of $H$-dimensional structural information of $G$ w.r.t $\mathcal{T}$ as follows:
    \begin{align}
        \begin{split}
            \mathcal{H}^\mathcal{T}(G;H)=&-\frac{1}{V} 
            \sum_{h=1}^H\sum_{i=1}^N (d_i-\sum_{j\in \mathcal{N}(i)}\mathbf{I}(i\overset{h}{\sim} j)w_{ij})
            \\ &\cdot \log_2(\frac{\sum_l^N\mathbf{I}(i \overset{h}{\sim} l)d_l}{\sum_l^N\mathbf{I}(i \overset{h-1}{\sim} l)d_l}),
        \end{split}
    \end{align}
    where $\mathbf{I}(\cdot)$ is the indicator function.}
\end{mymath}
\begin{proof}
From $\mathcal{H}^\mathcal{T}(G; \alpha) = -\frac{g_{\alpha}}{\text{Vol}(G)} \log_2 \left( \frac{V_{\alpha}}{V_{\alpha} - 1} \right)$, we can rewrite structural information of $G$ w.r.t all nodes of $\mathcal{T}$ at height $h$ as follows:
\begin{align}
\label{eq.si_at_k}
    \mathcal{H}^{\mathcal{T}}(G;h) &= -\frac{1}{V}\sum_{k=1}^{N_h}g^h_k \log_2 \frac{V^h_k}{V^{h-1}_{k^-}} \nonumber  \\
    &=-\frac{1}{V}[\sum_{k=1}^{N_h}g^h_k \log_2 V^h_k - \sum_{k=1}^{N_h}g^h_k \log_2 V^{h-1}_{k^-}].
\end{align}
Then the $H$-dimensional structural information of $G$ can be written as
\begin{align}
    \mathcal{H}^{\mathcal{T}}(G) = -\frac{1}{V}[\sum_{h=1}^H\sum_{k=1}^{N_h}g^h_k \log_2 V^h_k - \sum_{h=1}^H\sum_{k=1}^{N_h}g^h_k \log_2 V^{h-1}_{k^-}].
\end{align}
Since
\begin{align}
    g^h_k &= V^h_k - \sum_{i=1}^N \sum_{j \in \mathcal{N}(i)}\mathbf{I}(i\in T_{\alpha^h_k})\mathbf{I}(j\in T_{\alpha^h_k})w_{ij},
\end{align}
\begin{align}
    V^h_k = \sum_i^N\mathbf{I}(i\in T_{\alpha^h_k})d_i,
\end{align}
we have the following equation with the help of the equivalence relationship $i\overset{h}{\sim} j$:
\begin{align}
    \begin{split}
        \sum_{h=1}^H\sum_{k=1}^{N_h}g^h_k \log_2 V^h_k &= \sum_{h=1}^H\sum_{k=1}^{N_h}V^h_k \log_2 V^h_k \\ &-\sum_{h=1}^H\sum_{k=1}^{N_h}(\log_2 V^h_k) \sum_{i=1}^N \sum_{j \in \mathcal{N}(i)} \mathbf{I}(i\overset{h}{\sim} j)w_{ij}.
    \end{split}
\end{align}
Then we focus on the above equation term by term. If we fix the height of $h$ and the node $i$ of $G$, then the node $\alpha$ to which $i$ belongs and its ancestors $\alpha^-$ in $\mathcal{T}$ are also fixed. Let's denote $V^h_k$ as $V_\alpha$ where $h(\alpha)=h$.
\begin{align}
    \begin{split}
        & \sum_{h=1}^H\sum_{h(\alpha)=h}V_\alpha \log_2 V_\alpha \\
        =  & \sum_{h=1}^H\sum_{h(\alpha)=h}\sum_{i=1}^N\mathbf{I}(i\in T_\alpha)d_i \cdot\log_2(\sum_{l=1}^N\mathbf{I}(l\in T_\alpha)d_l)  \\
    = & \sum_{h=1}^H\sum_{i=1}^Nd_i \log_2 (\sum_{l=1}^N\mathbf{I}(i\overset{h}{\sim} l)d_l),
    \end{split}
\end{align}
\begin{align}
    \begin{split}
        &\sum_{h=1}^H\sum_{h(\alpha)=h}(\log_2 V_\alpha) \sum_{i=1}^N \sum_{j \in \mathcal{N}(i)} \mathbf{I}(i\overset{h}{\sim} j)w_{ij} \\ 
    = & \sum_{h=1}^H\sum_{i=1}^N [\log_2 (\sum_{l=1}^N\mathbf{I}(i\overset{h}{\sim} l)d_l)] \cdot \sum_{j \in \mathcal{N}(i)}\mathbf{I}(i\overset{h}{\sim} j)w_{ij}. 
    \end{split}
\end{align}
Similarly, 
\begin{align}
     \begin{split}
         \sum_{h=1}^H\sum_{h(\alpha)=h}g_\alpha \log_2 V_{\alpha^-} 
    =& \sum_{h=1}^H\sum_{h(\alpha)=h}V_\alpha \log_2 V_{\alpha^-} \\   &-\sum_{h=1}^H\sum_{h(\alpha)=h}(\log_2 V_{\alpha^-})\\
    &\cdot\sum_i^N \sum_{j \in \mathcal{N}(i)} \mathbf{I}(i\overset{h}{\sim} j)w_{ij},
     \end{split}
\end{align}
\begin{align}
    \begin{split}
        \sum_{h=1}^H\sum_{h(\alpha)=h}V_\alpha \log_2 V_\alpha &=  \sum_{h=1}^H\sum_{h(\alpha)=h}\sum_{i=1}^N\mathbf{I}(i\in T_\alpha)d_i \\ &\cdot\log_2(\sum_{l=1}^N\mathbf{I}(l\in T_{\alpha^-})d_l)  \\
    &= \sum_{h=1}^H\sum_{i=1}^Nd_i \log_2 (\sum_{l=1}^N\mathbf{I}(i\overset{h-1}{\sim} l)d_l),
    \end{split}
\end{align}
\begin{align}
    \begin{split}
        &\sum_{h=1}^H\sum_{h(\alpha)=h}(\log_2 V_{\alpha^-}) \sum_{i=1}^N \sum_{j \in \mathcal{N}(i)} \mathbf{I}(i\overset{h}{\sim} j)w_{ij}  \\ 
    &= \sum_{h=1}^H\sum_{i=1}^N [\log_2 (\sum_{l=1}^N\mathbf{I}(i\overset{h-1}{\sim} l)d_l)] \cdot \sum_{j \in \mathcal{N}(i)}\mathbf{I}(i\overset{h}{\sim} j)w_{ij}. 
    \end{split}
\end{align}
Utilizing the equations above, we have completed the proof.
\end{proof}

\subsection{Relaxation \& Flexiblity}
\label{proof.flex}

We demonstrate that the allowance of  partitioning tree  relaxation offers the flexiblity of neural architecture.

\vspace{0.05in}
\begin{mymath}
\noindent \textbf{Theorem IV.2} \textbf{(Flexiblity).}
    \emph{$\forall G$, given a partitioning tree $\mathcal{T}$ and adding a node $\beta$ to get a relaxed partitioning tree $\mathcal{T}'$, the structural information remains unchanged, $H^{\mathcal{T}}(G)=H^{\mathcal{T}'}(G)$, if  one of the following conditions holds:
    \begin{enumerate}
      \item $\beta$ as a leaf node, and the corresponding node subset $T_\beta$ is an empty set.
      \item $\beta$ is inserted between node $\alpha$ and its children nodes so that the corresponding node subsets $T_\beta = T_\alpha$.
\end{enumerate}}
\end{mymath}
\begin{proof}
    For the first case, since $\beta$ is a leaf node corresponds an empty subset of $\mathcal{V}$, $g_\beta=0$, then the structural information of $\beta$ is $\mathcal{H}^{\mathcal{T}}(G;\beta)=0$.
    For the second case, we notice the term $\log_2\frac{V_\alpha}{V_{\alpha^-}}$ in $\mathcal{H}^\mathcal{T}(G; \alpha) = -\frac{g_{\alpha}}{\text{Vol}(G)} \log_2 \left( \frac{V_{\alpha}}{V_{\alpha^-}} \right)$. Since the volume of $T_\alpha$ equals the one of $T_\beta$, the term vanishes to $0$.
\end{proof}

{
\subsection{Neural Lorentz Boost}
\label{proof.lg}

We specify the condition that a neural transformation $L$ is an element of Lorentz group, and give a construction of $L$ according to the closed-form solution.

\begin{mymath}
\noindent\textbf{Theorem VI.1}(\textbf{Closed-from Lorentz Boost})
    \emph{Given the construction in Eq. (32) and a parameter vector $\boldsymbol{\beta} \in \mathbb{R}^{d}$, Lorentz boosts have a closed form with components  given as
    \begin{align}
      w&=\frac{1}{\sqrt{1 - \| \boldsymbol{\beta} \|^2_\mathbb{L}}}  \\
      \boldsymbol{v}&=-w \cdot \boldsymbol{\beta} \\
      \mathbf{W}&= \frac{(w - 1)}{\| \boldsymbol{\beta} \|^2_\mathbb{L}} \cdot \boldsymbol{\beta} \boldsymbol{\beta}^\top + \mathbf{I}_d
    \end{align}}
\end{mymath}
\begin{proof}
We work in $(d+1)$-dimensional Minkowski spacetime $\mathbb{R}^{1,d}$ equipped with the metric tensor of signature $(-,+,+,\dots,+)$.
The Minkowski inner product of two vectors $X = (x^0, \mathbf{x})$ and $Y = (y^0, \mathbf{y})$, where $x^0, y^0 \in \mathbb{R}$ and $\mathbf{x}, \mathbf{y} \in \mathbb{R}^d$, is given by
\[
\langle X, Y \rangle_{\mathbb{L}} = -x^0 y^0 + \mathbf{x}^\top \mathbf{y}.
\]
A Lorentz boost is a linear isometry of this inner product that preserves the time orientation and has no spatial rotation component.

Let $\boldsymbol{\beta} \in \mathbb{R}^d$ be a spatial velocity vector satisfying $\|\boldsymbol{\beta}\| < 1$, where $\|\cdot\|$ denotes the standard Euclidean norm. In the $(-,+,+,\dots,+)$ convention, the relativistic gamma factor is still defined as
\[
w := \frac{1}{\sqrt{1 - \|\boldsymbol{\beta}\|^2}}.
\]

We seek a $(d+1)\times(d+1)$ Lorentz transformation $\Lambda$ of the block form
\[
\Lambda =
\begin{pmatrix}
w & \boldsymbol{v}^\top \\
\boldsymbol{v} & \mathbf{W}
\end{pmatrix},
\]
such that $\Lambda$ maps the rest frame to a frame moving with velocity $\boldsymbol{\beta}$, and satisfies the isometry condition
\[
\Lambda^\top \, \eta \, \Lambda = \eta,
\quad \text{where} \quad
\eta = \operatorname{diag}(-1, 1, \dots, 1).
\tag{1}
\]

From the physical interpretation of a boost, the worldline of an observer at rest in the new frame satisfies $\mathbf{x}(t) = \boldsymbol{\beta} t$ in the original coordinates. In the boosted frame, this observer must be at spatial rest, i.e., $\mathbf{x}' = \mathbf{0}$. Writing the transformation as
\[
\begin{pmatrix} t' \\ \mathbf{x}' \end{pmatrix}
= \Lambda \begin{pmatrix} t \\ \mathbf{x} \end{pmatrix}
= \begin{pmatrix} w t + \boldsymbol{v}^\top \mathbf{x} \\ \boldsymbol{v} t + \mathbf{W} \mathbf{x} \end{pmatrix},
\]
and substituting $\mathbf{x} = \boldsymbol{\beta} t$, we require $\mathbf{x}' = \boldsymbol{v} t + \mathbf{W} \boldsymbol{\beta} t = \mathbf{0}$ for all $t$, which implies
\[
\boldsymbol{v} + \mathbf{W} \boldsymbol{\beta} = \mathbf{0}.
\tag{2}
\]

Moreover, the time component in special relativity (in any signature convention) must satisfy $t' = w(t - \boldsymbol{\beta}^\top \mathbf{x})$. Matching with the first row of $\Lambda$, we obtain
\[
w t + \boldsymbol{v}^\top \mathbf{x} = w t - w \boldsymbol{\beta}^\top \mathbf{x}
\quad \Rightarrow \quad
\boldsymbol{v} = -w \boldsymbol{\beta}.
\tag{3}
\]

Substituting (3) into (2) yields
\[
-w \boldsymbol{\beta} + \mathbf{W} \boldsymbol{\beta} = \mathbf{0}
\quad \Rightarrow \quad
\mathbf{W} \boldsymbol{\beta} = w \boldsymbol{\beta}.
\tag{4}
\]

From Eq. (4), we have $\mathbf{W} \boldsymbol{\beta} = w \boldsymbol{\beta}$. This equation alone does not uniquely determine $\mathbf{W}$, as it allows for arbitrary rotations around $\boldsymbol{\beta}$. However, by definition, a pure Lorentz boost is free of spatial rotations. This implies that the transformation must act as the identity on the subspace orthogonal to the velocity $\boldsymbol{\beta}$.

We decompose any spatial vector $\mathbf{x} \in \mathbb{R}^d$ into a component parallel to $\boldsymbol{\beta}$, denoted $\mathbf{x}_{\parallel}$, and a component orthogonal to $\boldsymbol{\beta}$, denoted $\mathbf{x}_{\perp}$. The action of $\mathbf{W}$ is thus given by:
\[
\mathbf{W}\mathbf{x} = \mathbf{W}(\mathbf{x}_{\parallel} + \mathbf{x}_{\perp}) = w \mathbf{x}_{\parallel} + \mathbf{x}_{\perp},
\]
where we used the boost property along the motion ($\mathbf{W}\mathbf{x}_{\parallel} = w\mathbf{x}_{\parallel}$) and the invariance of transverse coordinates ($\mathbf{W}\mathbf{x}_{\perp} = \mathbf{x}_{\perp}$).
Using the projection operator $\mathbf{P}_{\parallel} = \frac{\boldsymbol{\beta}\boldsymbol{\beta}^\top}{\|\boldsymbol{\beta}\|^2}$, we write $\mathbf{x}_{\parallel} = \mathbf{P}_{\parallel}\mathbf{x}$ and $\mathbf{x}_{\perp} = (\mathbf{I} - \mathbf{P}_{\parallel})\mathbf{x}$. Substituting these back yields:
\[
\mathbf{W} = w \mathbf{P}_{\parallel} + (\mathbf{I} - \mathbf{P}_{\parallel}) = \mathbf{I} + (w - 1)\mathbf{P}_{\parallel}.
\]
Explicitly substituting $\mathbf{P}_{\parallel}$:
\[
\mathbf{W} = \mathbf{I}_d + \frac{w - 1}{\|\boldsymbol{\beta}\|^2} \boldsymbol{\beta} \boldsymbol{\beta}^\top.
\]
This uniquely determines the symmetric matrix $\mathbf{W}$.

Finally, denoting the Euclidean norm $\|\boldsymbol{\beta}\|$ as $\|\boldsymbol{\beta}\|_{\mathbb{L}}$ (as in the statement of the theorem), we obtain the desired expressions:
\begin{align}
w &= \frac{1}{\sqrt{1 - \|\boldsymbol{\beta}\|_{\mathbb{L}}^2}}, \\
\boldsymbol{v} &= -w \, \boldsymbol{\beta}, \\
\mathbf{W} &= \frac{w - 1}{\|\boldsymbol{\beta}\|_{\mathbb{L}}^2} \, \boldsymbol{\beta} \boldsymbol{\beta}^\top + \mathbf{I}_d.
\end{align}
This completes the proof.
\end{proof}

\vspace{0.05in}
\begin{mymath}
\noindent\textbf{Theorem VI.2} \textbf{(Manifold Preserving).}
    \emph{Given the construction of $\mathbf L$ in Theorem VI.1, and scalars $a, b$, the following properties holds for any  $\boldsymbol{x}, \boldsymbol{y} \in \mathbb{L}^{d_1}$.
    \begin{itemize}
        \item[1)] $L(a\boldsymbol{x}+b\boldsymbol{y})=aL(\boldsymbol{x})+ bL(\boldsymbol{y})$;
        \item[2)] $\langle L(\boldsymbol{x}), L(\boldsymbol{y})  \rangle_\mathbb{L}=\langle \boldsymbol{x}, \boldsymbol{y}\rangle_\mathbb{L}$,
    \end{itemize}
    where $\langle \boldsymbol{x}, \boldsymbol{y}\rangle_\mathbb{L}$ denotes the Minkowski inner product.}
    \end{mymath}
\begin{proof}
Let $\mathbf{L} \in \mathbb{R}^{(d+1) \times (d+1)}$ be the matrix form derived in Theorem VI.1:
\[
\mathbf{L} =
\begin{pmatrix}
w & \boldsymbol{v}^\top \\
\boldsymbol{v} & \mathbf{W}
\end{pmatrix},
\]
where $w = (1-\|\boldsymbol{\beta}\|^2)^{-1/2}$, $\boldsymbol{v} = -w\boldsymbol{\beta}$, and $\mathbf{W} = \mathbf{I}_d + \frac{w-1}{\|\boldsymbol{\beta}\|^2}\boldsymbol{\beta}\boldsymbol{\beta}^\top$.

We work with the metric tensor $\boldsymbol{\eta} = \operatorname{diag}(-1, \mathbf{I}_d)$. The inner product is $\langle \boldsymbol{x}, \boldsymbol{y} \rangle_{\mathbb{L}} = \boldsymbol{x}^\top \boldsymbol{\eta} \, \boldsymbol{y}$.

\medskip
\noindent\textbf{Part 1) Linearity.}
Linearity follows directly from matrix algebra definitions. For any scalars $a, b$ and vectors $\boldsymbol{x}, \boldsymbol{y}$:
\[
\mathbf{L}(a \boldsymbol{x} + b \boldsymbol{y}) = a \mathbf{L}\boldsymbol{x} + b \mathbf{L}\boldsymbol{y} = a L(\boldsymbol{x}) + b L(\boldsymbol{y}).
\]

\medskip
\noindent\textbf{Part 2) Inner product preservation.}
The transformation preserves the inner product if and only if $\mathbf{L}^\top \boldsymbol{\eta} \mathbf{L} = \boldsymbol{\eta}$. Computing the block product (noting $\mathbf{L}$ is symmetric):
\begin{align}
\mathbf{L}^\top \boldsymbol{\eta} \mathbf{L} &= 
\begin{pmatrix} w & \boldsymbol{v}^\top \\ \boldsymbol{v} & \mathbf{W} \end{pmatrix}
\begin{pmatrix} -1 & \mathbf{0}^\top \\ \mathbf{0} & \mathbf{I} \end{pmatrix}
\begin{pmatrix} w & \boldsymbol{v}^\top \\ \boldsymbol{v} & \mathbf{W} \end{pmatrix}\\
&=
\begin{pmatrix}
-w^2 + \|\boldsymbol{v}\|^2 & -w\boldsymbol{v}^\top + \boldsymbol{v}^\top \mathbf{W} \\
-w\boldsymbol{v} + \mathbf{W}\boldsymbol{v} & -\boldsymbol{v}\boldsymbol{v}^\top + \mathbf{W}^2
\end{pmatrix}.
\end{align}
To verify this equals $\boldsymbol{\eta}$, we must verify three conditions:
\begin{enumerate}
    \item \textbf{Time component:} $-w^2 + \|\boldsymbol{v}\|^2 = -1$.
    Substituting $\boldsymbol{v} = -w\boldsymbol{\beta}$:
    \[ -w^2 + w^2\|\boldsymbol{\beta}\|^2 = -w^2(1 - \|\boldsymbol{\beta}\|^2) = -w^2(1/w^2) = -1.  \]
    
    \item \textbf{Off-diagonal blocks:} $-w\boldsymbol{v} + \mathbf{W}\boldsymbol{v} = \mathbf{0}$.
    From Theorem VI.1 Eq.(4), we established $\mathbf{W}\boldsymbol{\beta} = w\boldsymbol{\beta}$. Multiplying by $-w$, we get $\mathbf{W}(-w\boldsymbol{\beta}) = w(-w\boldsymbol{\beta})$, which implies $\mathbf{W}\boldsymbol{v} = w\boldsymbol{v}$. Thus $-w\boldsymbol{v} + \mathbf{W}\boldsymbol{v} = \mathbf{0}$. 
    
    \item \textbf{Spatial block:} $-\boldsymbol{v}\boldsymbol{v}^\top + \mathbf{W}^2 = \mathbf{I}_d$.
    We explicitly compute $\mathbf{W}^2$. Let $k = \|\boldsymbol{\beta}\|^2$ and $c = \frac{w-1}{k}$. Then $\mathbf{W} = \mathbf{I} + c\boldsymbol{\beta}\boldsymbol{\beta}^\top$.
    \[
    \mathbf{W}^2 = (\mathbf{I} + c\boldsymbol{\beta}\boldsymbol{\beta}^\top)^2 = \mathbf{I} + 2c\boldsymbol{\beta}\boldsymbol{\beta}^\top + c^2(\boldsymbol{\beta}\boldsymbol{\beta}^\top)^2.
    \]
    Note that $(\boldsymbol{\beta}\boldsymbol{\beta}^\top)^2 = \boldsymbol{\beta}(\boldsymbol{\beta}^\top\boldsymbol{\beta})\boldsymbol{\beta}^\top = k \boldsymbol{\beta}\boldsymbol{\beta}^\top$. Thus:
    \[
    \mathbf{W}^2 = \mathbf{I} + (2c + c^2 k)\boldsymbol{\beta}\boldsymbol{\beta}^\top.
    \]
    We simplify the scalar coefficient $2c + c^2 k$:
    \[
    2\frac{w-1}{k} + \frac{(w-1)^2}{k^2}k = \frac{w-1}{k} (2 + w - 1) = \frac{w^2 - 1}{k}.
    \]
    Recall $w^2 = \frac{1}{1-k} \Rightarrow w^2(1-k)=1 \Rightarrow w^2-1 = w^2 k$.
    Thus, the coefficient becomes $\frac{w^2 k}{k} = w^2$.
    So, $\mathbf{W}^2 = \mathbf{I} + w^2 \boldsymbol{\beta}\boldsymbol{\beta}^\top$.
    Since $\boldsymbol{v} = -w\boldsymbol{\beta}$, we have $\boldsymbol{v}\boldsymbol{v}^\top = w^2 \boldsymbol{\beta}\boldsymbol{\beta}^\top$.
    Therefore, $-\boldsymbol{v}\boldsymbol{v}^\top + \mathbf{W}^2 = -w^2 \boldsymbol{\beta}\boldsymbol{\beta}^\top + (\mathbf{I} + w^2 \boldsymbol{\beta}\boldsymbol{\beta}^\top) = \mathbf{I}_d$. 
\end{enumerate}

Substituting these verified blocks back, we confirm 
\[
\mathbf{L}^\top \boldsymbol{\eta} \mathbf{L} = \operatorname{diag}(-1, \mathbf{I}_d) = \boldsymbol{\eta}.
\]
This completes the proof.
\end{proof}

\subsection{Connection to Tree Contrastive Loss}

\begin{mymath}
    \textbf{Theorem VII.1}(\textbf{Connection between Structural Entropy and Tree Contrastive Learning})
    \emph{Given an induced weighted graph $\tilde{G} = (V, \tilde{A})$, where $\tilde{A}$ is the virtual adjacency matrix constructed via Eq. (37),
the tree contrastive learning with the loss function of Eq. (36)  minimizes the graph conductance of $\tilde{G}$, and is upper bounded by the $H$-dimensional structural information $\mathcal{H}^H(\tilde{G})$.}
\end{mymath}
\begin{proof}
    We provide a constructive proof demonstrating that under the stated geometric assumptions, the InfoNCE objective creates a graph topology where the "natural" partition exhibits vanishing conductance, thereby aligning with the objective of structural entropy minimization.
    
    We begin by establishing geometric bounds on the embedding distances derived from the InfoNCE optimization properties. The loss function explicitly minimizes $d_\mathbb{L}(\boldsymbol{z}_i, \boldsymbol{z}'_i)$ for positive pairs while maximizing distances for negative pairs. Based on the theorem's assumptions, we derive the following bounds:
    \begin{itemize}
        \item \textbf{Intra-cluster Bound:} Standard InfoNCE treats distinct nodes $i, k \in S$ ($i \neq k$) as negative pairs. However, under the \emph{Connectivity} assumption, there exists a sequence of augmented views linking $i$ to $k$ (e.g., $z_i \to z'_i \approx z'_k \to z_k$). By the triangle inequality of the metric $d_\mathbb{L}$ and Assumption (i), the distance between any intra-cluster pair is bounded by the sum of alignment errors along this path:
        \[
            d_\mathbb{L}(\boldsymbol{z}_i, \boldsymbol{z}_k) \le L \cdot \varepsilon := \delta.
        \]
        This ensures that despite the repulsive term in the loss, intra-cluster nodes remain confined within a radius $\delta$, guaranteeing high edge weights: $\tilde{A}_{ik} \ge \exp(-\delta/t)$.
        
        \item \textbf{Inter-cluster Bound:} For $i \in S$ and $j \notin S$, Assumption (iii) enforces a worst-case separation $M$. This reflects the Uniformity property of contrastive learning, where distinct semantic manifolds are pushed apart. Thus, $\tilde{A}_{ij} \le \exp(-M/t)$.
    \end{itemize}

    Next, we apply these bounds to analyze the conductance $\tilde{\phi}(S)$ of a semantic cluster $S$ on the induced graph $\tilde{G}$. Recall that conductance is defined as the ratio of the cut weight to the volume:
    \[
        \tilde{\phi}(S) = \frac{\text{Cut}(S, S^c)}{\text{Vol}(S)} = \frac{\sum_{i \in S} \sum_{j \notin S} \tilde{A}_{ij}}{\sum_{i \in S} (\sum_{k \in S} \tilde{A}_{ik} + \sum_{j \notin S} \tilde{A}_{ij})}.
    \]
    Substituting the geometric bounds derived in Step 1:
    \begin{itemize}
        \item The numerator is bounded strictly by the separation assumption:
        \[
        \text{Cut}(S, S^c) \le |S| \cdot |S^c| \cdot e^{-M/t}.
        \]
        \item The denominator is lower-bounded by the intra-cluster cohesion (ignoring the vanishing inter-cluster terms):
        \[
        \text{Vol}(S) \ge \sum_{i \in S} \sum_{k \in S} e^{-\delta/t} = |S|^2 \cdot e^{-\delta/t}.
        \]
    \end{itemize}
    Combining these, the conductance is bounded by:
    \begin{equation}
        \tilde{\phi}(S) \le \frac{|S| |S^c| e^{-M/t}}{|S|^2 e^{-\delta/t}} = \frac{|S^c|}{|S|} \exp\left( -\frac{M - \delta}{t} \right).
    \end{equation}
    In the limit of ideal optimization where alignment is tight ($\delta \to 0$) and separation is large ($M \to \infty$), the term $\exp(-(M-\delta)/t)$ vanishes. This proves that minimizing InfoNCE drives $\tilde{\phi}(S) \to 0$.

    Finally, we demonstrate the consistency with Structural Entropy. The $H$-dimensional structural entropy $\mathcal{H}^H(\tilde{G})$ measures the dynamic uncertainty of random walks on the graph given a hierarchy. A fundamental result in structural entropy theory states that $\mathcal{H}^H(\tilde{G})$ is upper-bounded by a function of the local conductances of the partition nodes:
    \[
        \mathcal{H}^H(\tilde{G}) \le C + \sum_{S \in \mathcal{P}} \text{Vol}(S) \cdot \tilde{\phi}(S),
    \]
    where $C$ is a constant. Minimizing the weighted sum of conductances is therefore a sufficient condition to lower the upper bound of structural entropy.
    
    As derived above, minimizing the InfoNCE loss forces $\tilde{\phi}(S) \to 0$ for clusters corresponding to semantic categories. Consequently, the optimization trajectory of InfoNCE effectively minimizes the upper bound of the structural entropy. Thus, the two objectives are geometrically consistent in promoting low-conductance partitions.
    
    This completes the proof.
\end{proof}

\subsection{Augmented Structural Entropy}

\begin{mymath}
\textbf{Theorem VII.2 }(\textbf{Graph Conductance Improvement with Graph Fusion})
\emph{Given a graph $G=(\mathcal{V}, \mathcal{E})$ and a virtual graph $\tilde{G}$ as in Theorem VII.1, the fused graph $G^\gamma$ is derived as in Eq. (39).
    For any nontrivial subset $\mathcal{S} \subset \mathcal{V}$ with $0 < \mathrm{vol}_A(\mathcal{S}) < \mathrm{vol}_A(V)$, 
    denote the conductance of $\mathcal{S}$ under a graph $G$ as
    \[
        \phi_G(\mathcal{S}) = \frac{\mathrm{cut}_G(\mathcal{S}, \mathcal{V} \setminus \mathcal{S})}{\min\{\mathrm{vol}_G(\mathcal{S}),\, \mathrm{vol}_G(\mathcal{V} \setminus \mathcal{S})\}},
    \]
    where $\mathrm{cut}_G(\mathcal{S}, \mathcal{P}) = \sum_{i \in \mathcal{S}, j \in \mathcal{P}} w_{ij}$ and $\mathrm{vol}_G(\mathcal{S}) = \sum_{i \in \mathcal{S}} \sum_j w_{ij}$.
    Suppose that for every $\mathcal{S} \subset \mathcal{V}$,
    \begin{equation}\label{eq:cond_assump}
        \phi_{\tilde{G}}(\mathcal{S}) \leq \phi_G(\mathcal{S}).
    \end{equation}
    Then, the conductance of the fused graph satisfies
    \[
        \phi_{G^\alpha}(\mathcal{S}) \leq \phi_G(\mathcal{S}), \quad \forall \mathcal{S} \subset V.
    \]
    Moreover, if there exists a subset $\mathcal{S}^\star$ such that $\phi_{\tilde{G}}(\mathcal{S}^\star) < \phi_G(\mathcal{S}^\star)$, 
    then $\phi_{G^\alpha}(\mathcal{S}^\star) < \phi_G(\mathcal{S}^\star)$ for all $\alpha \in (0,1]$.}
\end{mymath}
\begin{proof}
    We prove the theorem by analyzing the numerator (cut capacity) and denominator (effective volume) of the conductance ratio separately, utilizing the linearity of the cut and the concavity of the volume metric.
    Let $\bar{\mathcal{S}} = \mathcal{V} \setminus \mathcal{S}$ denote the complement of $\mathcal{S}$.

    First, consider the cut functional. Let $C_G(\mathcal{S}) = \sum_{i \in \mathcal{S}, j \in \bar{\mathcal{S}}} A_{ij}$ denote the cut size in graph $G$.
    Since the fused adjacency $A^\alpha$ is a linear combination of $A$ and $\tilde{A}$, and the cut summation is linear with respect to edge weights, the cut in $G^\alpha$ is given by:
    \[
        C_{G^\alpha}(\mathcal{S}) = (1 - \alpha) C_G(\mathcal{S}) + \alpha C_{\tilde{G}}(\mathcal{S}).
    \]

    Next, we examine the effective volume $d_G(\mathcal{S}) = \min\{\mathrm{vol}_G(\mathcal{S}), \mathrm{vol}_G(\bar{\mathcal{S}})\}$.
    First, note that the volume functional $\mathrm{vol}(\cdot)$ is linear in $A$. Thus:
    \[
        \mathrm{vol}_{G^\alpha}(\mathcal{S}) = (1-\alpha)\mathrm{vol}_G(\mathcal{S}) + \alpha\mathrm{vol}_{\tilde{G}}(\mathcal{S}).
    \]
    Consider the function $f(x, y) = \min(x, y)$. This function is \textit{concave} on the domain $\mathbb{R}_{\ge 0}^2$. 
    By the definition of concavity (a variation of Jensen's inequality for linear combinations), for any non-negative vectors $\mathbf{u}, \mathbf{v} \in \mathbb{R}^2$ and $\lambda \in [0, 1]$, we have $f(\lambda \mathbf{u} + (1-\lambda)\mathbf{v}) \geq \lambda f(\mathbf{u}) + (1-\lambda)f(\mathbf{v})$.
    
    Applying this to the volumes of $\mathcal{S}$ and $\bar{\mathcal{S}}$:
    \begin{align}
        d_{G^\alpha}(\mathcal{S}) 
        &= \min( (1-\alpha)\mathrm{vol}_G(\mathcal{S}) + \alpha\mathrm{vol}_{\tilde{G}}(\mathcal{S}), \nonumber \\
        & (1-\alpha)\mathrm{vol}_G(\bar{\mathcal{S}}) + \alpha\mathrm{vol}_{\tilde{G}}(\bar{\mathcal{S}}) ) \nonumber \\
        &\geq (1-\alpha) \min\{\mathrm{vol}_G(\mathcal{S}), \mathrm{vol}_G(\bar{\mathcal{S}})\} \\
        &+ \alpha \min\{\mathrm{vol}_{\tilde{G}}(\mathcal{S}), \mathrm{vol}_{\tilde{G}}(\bar{\mathcal{S}})\} \nonumber \\
        &= (1-\alpha) d_G(\mathcal{S}) + \alpha d_{\tilde{G}}(\mathcal{S}). \label{eq:volume_concavity}
    \end{align}
    This establishes that the effective volume of the fused graph is bounded below by the weighted average of the individual volumes.

    Combining the linearity of the numerator and this lower bound of the denominator, the conductance of the fused graph satisfies: $\phi_{G^\alpha}(\mathcal{S}) = \frac{C_{G^\alpha}(\mathcal{S})}{d_{G^\alpha}(\mathcal{S})}$.
    Using the lower bound for the denominator from (10), we have:
    \begin{equation}\label{eq:phi_upper_bound}
        \phi_{G^\alpha}(\mathcal{S}) \leq \frac{(1-\alpha)C_G(\mathcal{S}) + \alpha C_{\tilde{G}}(\mathcal{S})}{(1-\alpha)d_G(\mathcal{S}) + \alpha d_{\tilde{G}}(\mathcal{S})}.
    \end{equation}
    The right-hand side represents the \textit{mediant} (weighted average) of the individual conductances $\phi_G(\mathcal{S}) = \frac{C_G(\mathcal{S})}{d_G(\mathcal{S})}$ and $\phi_{\tilde{G}}(\mathcal{S}) = \frac{C_{\tilde{G}}(\mathcal{S})}{d_{\tilde{G}}(\mathcal{S})}$.
    We invoke the standard algebraic property of the mediant: for $a, b, c, d > 0$, if $\frac{c}{d} \leq \frac{a}{b}$, then $\frac{a+c}{b+d} \leq \frac{a}{b}$.
    
    Setting $a = (1-\alpha)C_G(\mathcal{S})$, $b = (1-\alpha)d_G(\mathcal{S})$ and $c = \alpha C_{\tilde{G}}(\mathcal{S})$, $d = \alpha d_{\tilde{G}}(\mathcal{S})$, the assumption $\phi_{\tilde{G}}(\mathcal{S}) \leq \phi_G(\mathcal{S})$ implies $\frac{c}{d} \leq \frac{a}{b}$. Therefore:
    \[
        \frac{(1-\alpha)C_G(\mathcal{S}) + \alpha C_{\tilde{G}}(\mathcal{S})}{(1-\alpha)d_G(\mathcal{S}) + \alpha d_{\tilde{G}}(\mathcal{S})} \leq \phi_G(\mathcal{S}).
    \]
    Combined with (11), we obtain $\phi_{G^\alpha}(\mathcal{S}) \leq \phi_G(\mathcal{S})$.

    Finally, consider the case of strictly better semantic conductance: $\phi_{\tilde{G}}(\mathcal{S}) < \phi_G(\mathcal{S})$.
    The strict improvement in $\phi_{G^\alpha}(\mathcal{S})$ arises from two independent sources:
    \begin{enumerate}
        \item \textbf{Strict Mediant:} Since $\phi_{\tilde{G}}(\mathcal{S}) < \phi_G(\mathcal{S})$ and $\alpha > 0$, the weighted average in the RHS of (11) is strictly less than the maximum:
        \[
             \frac{(1-\alpha)C_G(\mathcal{S}) + \alpha C_{\tilde{G}}(\mathcal{S})}{(1-\alpha)d_G(\mathcal{S}) + \alpha d_{\tilde{G}}(\mathcal{S})} < \phi_G(\mathcal{S}).
        \]
        \item \textbf{Volume Concavity:} Even if the mediant equality held, any strict inequality in the volume bound (10) (i.e., $d_{G^\alpha}(\mathcal{S}) > (1-\alpha)d_G(\mathcal{S}) + \alpha d_{\tilde{G}}(\mathcal{S})$) would strictly increase the denominator, further reducing $\phi_{G^\alpha}(\mathcal{S})$.
    \end{enumerate}
    Since the first source (Strict Mediant) guarantees a strict inequality regardless of the volume behavior, we conclude:
    \[
        \phi_{G^\alpha}(\mathcal{S}) \leq \text{Mediant} < \phi_G(\mathcal{S}).
    \]
    Thus, $\phi_{G^\alpha}(\mathcal{S}) < \phi_G(\mathcal{S})$ holds strictly.
\end{proof}
}

\subsection{Minority Cluster of Imbalanced Graphs}
\label{proof.imbalanced}

We demonstrate that partitioning a minority cluster decreases the structural entropy of the partitioning tree, given the graph's imbalance.
In other words,  our approach is easier to detect minority clusters.

\vspace{0.05in}
\begin{mymath}
\noindent \textbf{Theorem IV.3} \textbf{(Identifiability)}
    \emph{Given a highly internally connected and unbalanced graph \( G = (V, E) \) with \( N \) nodes (\( N \gg 3 \)), where \( V_1 \) and \( V_2 \) are large classes and \( V_\varepsilon \) is a very small class, such that \( |V_1| \gg |V_\varepsilon| \) and \( |V_2| \gg |V_\epsilon| \), as structural information decreasing, the small class \( V_\varepsilon \) will be distinguished from large classes, i.e., 
    \begin{align}
        \mathcal{H}(G, V_2) + \mathcal{H}(G, V_1 + V_\varepsilon) > \sum_{i\in\{1,2,\varepsilon\}}\mathcal{H}(G, V_i).
    \end{align}
    }
    \end{mymath}
\begin{proof}
Without loss of generality, we only consider the case of 2 large groups and 1 small group.
For an imbalanced graph \( G = (V, E) \) with \(N\) nodes (\( N \gg 3 \)), which is divided into 3 classes.

The two groups \(G_1\) and \(G_2\) are considered as two major classes such that  \( |V_1| \gg |V_\varepsilon| \) and  \( |V_1| \gg |V_\varepsilon| \). The number of cut edges between \(G_1\) and \(G_2\) is \(g_\alpha\), and the one between \(G_1\) and \(G_\varepsilon\) is \(g_\varepsilon\). To simplify the notation, we denote 
\begin{align}
    \mathcal{H}_1&:=\mathcal{H}(G, V_2) + \mathcal{H}(G, V_1 + V_\varepsilon)  \\
    \mathcal{H}_2&:=\sum_{i\in\{1,2,\varepsilon\}}\mathcal{H}(G, V_i) 
\end{align}
When $G_\varepsilon$ merges to \(G_1\), the graph structural information is
\begin{align}
    \begin{split}
        \mathcal{H}_1 &= -g_\alpha \log_{2}\left(\frac{V_1+V_\varepsilon}{V}\right) - g_\alpha \log_{2}\left(\frac{V_2}{V}\right) \\
    &\quad - \sum_i d_i \log_{2}\left(\frac{d_i}{V_1+V_\varepsilon}\right) - \sum_j d_j \log_{2}\left(\frac{d_j}{V_2}\right)\\
    &\quad- \sum_k d_k \log_{2}\left(\frac{d_k}{V_1+V_\varepsilon}\right),
    \end{split}
\end{align}
In another case, when these 3 categories are treated as independent clusters, the graph structural information is written as follows,
\begin{align}
    \begin{split}
        \mathcal{H}_2 &= -g_\alpha \log_{2}\left(\frac{V_1}{V}\right) - g_\alpha \log_{2}\left(\frac{V_2}{V}\right) - g_\varepsilon \log_{2}\left(\frac{V_\varepsilon}{V}\right) \\
    &\quad - \sum_i d_i \log_{2}\left(\frac{d_i}{V_1}\right) - \sum_j d_j \log_{2}\left(\frac{d_j}{V_2}\right)\\
    &\quad- \sum_k d_k \log_{2}\left(\frac{d_k}{V_\varepsilon}\right).
    \end{split}
\end{align}
Then, we subtract $\mathcal{H}_2$ from $\mathcal{H}_1$,
\begin{align}
    \mathcal{H}_1 - \mathcal{H}_2 &= -\Bigg( g_\alpha \log_2\left(\frac{V_1 + V_\varepsilon}{V_1}\right) 
    + \sum_i d_i \log_2\left(\frac{V_1}{V_1 + V_\varepsilon}\right) \nonumber \\
    &\quad + \sum_k d_k \log_2\left(\frac{V_\varepsilon}{V_1 + V_\varepsilon}\right) 
    - g_\varepsilon \log_2\left(\frac{V_\varepsilon}{V}\right) \Bigg) \nonumber \\
    &= -\Big( g_\alpha \log_2(V_1 + V_\varepsilon) - g_\alpha \log_2(V_1) \nonumber \\
    &\quad + V_1 \log_2 V_1 - V_1 \log_2(V_1 + V_\varepsilon) \nonumber \\
    &\quad + V_\varepsilon \log_2 V_\varepsilon - V_\varepsilon \log_2(V_1 + V_\varepsilon) \nonumber \\
    &\quad - g_\varepsilon \log_2 V_\varepsilon + g_\varepsilon \log_2 V \Big) \nonumber \\
    &= -\Big[ (g_\alpha - V_1 - V_\varepsilon) \log_2(V_1 + V_\varepsilon) \nonumber \\
    &\quad + (V_1 - g_\alpha) \log_2 V_1 + (V_\varepsilon - g_\varepsilon) \log_2 V_\varepsilon \nonumber \\
    &\quad + g_\varepsilon \log_2 V \Big] \nonumber \\
    &= (V_1 + V_\varepsilon - g_\alpha) \log_2(V_1 + V_\varepsilon) \nonumber \\
    &\quad - (V_1 - g_\alpha) \log_2 V_1 - (V_\varepsilon - g_\varepsilon) \log_2 V_\varepsilon \nonumber \\
    &\quad - g_\varepsilon \log_2 V.
\end{align}
Let \(\frac{V_\varepsilon}{V_1} = \xi_1, V_1 + V_\varepsilon = (\xi_1 + 1)V_1, \frac{V_\varepsilon}{V_2} = \xi_2\),
\begin{align}
    \begin{split}
        \mathcal{H}_1 - \mathcal{H}_2&= (V_1 + V_\varepsilon - g_\alpha) \log_{2} (\xi_1 + 1) V_1 - (V_1 - g_\alpha) \\
    &\cdot\log_{2} V_1 - (V_\varepsilon - g_\varepsilon) \log_{2} (\xi_1 V_1) - g_\varepsilon \log_{2} V \\
    &= (V_1 + V_\varepsilon - g_\alpha) \log_{2} (\xi_1 + 1) + V_\varepsilon \log_{2} V_1\\& - (V_\varepsilon - g_\varepsilon) \log_{2} \xi_1 - (V_\varepsilon - g_\varepsilon) \log_{2} V_1-g_\varepsilon \log_{2} V \\
    &= (V_1 + V_\varepsilon - g_\alpha) \log_{2} (\xi_1 + 1) + g_\varepsilon \log_{2} V_1\\
    & - (V_\varepsilon - g_\varepsilon) \log_{2} \xi_1 - g_\varepsilon \log_{2} V\\
    &> (V_1 + g_\varepsilon - g_\alpha) \log_{2} (\xi_1 + 1) + g_\varepsilon \log_{2} V_1\\
    &- g_\varepsilon \log_{2} V
    \end{split}
\end{align}
We have the following inequality,
\begin{align}
    \begin{split}
        \mathcal{H}_1 - \mathcal{H}_2&= (V_1- g_\alpha) \log_{2} (\xi_1 + 1) - g_\varepsilon \log_{2}\frac{1+\xi_1+\frac{V_2}{V_1}}{1+\xi_1}\\
    &=(V_1- g_\alpha) \log_{2} (\xi_1 + 1) - g_\varepsilon \log_{2}(1+\frac{V_2}{V_1+V_\varepsilon})\\
    &>V_1log_{2}(1+\xi_1)-g_\alpha log_{2}(1+\frac{\overline{V_1}}{V_1})\\
  & = log_{2}(\frac{(1+\xi_1)^{g_\alpha}}{(1+\frac{\overline{V_1}}{V_1})^{g_\alpha}}) + log_{2} (1+\xi_1)^{V_1-g_\alpha} \\ 
   &=  log_{2}((\frac{V_1+V_\epsilon}{V_1+V_\epsilon+V_2})^{g_\alpha}  (1+\xi_1)^{V_1-g_\alpha}).
    \end{split}
\end{align}
Since
\begin{align}
 \frac{V}{V_1} = \frac{V_1 + V_2 + V_\varepsilon}{V_1} = 1 + \varepsilon_1 + \frac{V_2}{V_1},V_1\gg g_\alpha,
\end{align}
we have
\begin{align}
 \mathcal{H}_1 > \mathcal{H}_2
 \end{align}
\end{proof}

\subsection{Geometric Centroid in Manifolds}
\label{proof.geoeq}

We show that the arithmetic mean is the geometric centroid in Lorentz model, and is equivalent to gyro-midpoint in Poincar\'{e} ball model.

 \vspace{0.05in}
 \begin{mymath}
\noindent \textbf{Theorem VI.2} \textbf{(Geometric Centroid).} 
\emph{In hyperbolic space $\mathbb L^{\kappa, d_{\mathcal T}}$,
for any set of $\{\boldsymbol z^{h}_i\}$,
the arithmetic mean of
$\boldsymbol z^{h-1}_j =
\frac{1}{\sqrt{-\kappa}} \sum_{i=1}^n \frac{ c_{ij}}{\lvert \lVert \sum_{l=1}^n c_{lj} \boldsymbol z^{h}_l  \rVert_\mathbb{L} \rvert} \boldsymbol z^{h}_i $
is the manifold $\boldsymbol z^{h-1}_j \in \mathbb L^{\kappa, d_{\mathcal T}}$,
and is the close-form solution of the geometric centroid specified in the minimization of 
$z^{h-1}_j = \arg_{z_j^{h-1}} \min\sum_{i=1}^{N} c_{ij} d^2_L \left( z^{h-1}_j, z^h_i \right).$
}
\end{mymath}
\begin{proof}
Since $d^2_\mathbb{L}(\boldsymbol z^h_i, \boldsymbol z^{h-1}_j) = \lVert \boldsymbol z^h_i- \boldsymbol z^{h-1}_j\rVert^2_\mathbb{L}=\lVert \boldsymbol z^h_i\rVert^2_\mathbb{L} + \lVert \boldsymbol z^{h-1}_j\rVert^2_\mathbb{L} - 2\langle \boldsymbol z^h_i, \boldsymbol z^{h-1}_j 
 \rangle_\mathbb{L} = 2/\kappa - 2\langle \boldsymbol z^h_i, \boldsymbol z^{h-1}_j 
 \rangle_\mathbb{L}$, minimizing $z^{h-1}_j=\sum^N_{i=1}c_{ij}d^2_L(\left( z^{h-1}_j, z^h_i \right))$ is equivalent to maximizing $\sum_i c_{ij} \langle \boldsymbol z^h_i, \boldsymbol z^{h-1}_j 
 \rangle_\mathbb{L} =  \langle \sum_i c_{ij} \boldsymbol z^h_i, \boldsymbol z^{h-1}_j \rangle_\mathbb{L}$. To maximize the Lorentz inner product, $\boldsymbol z^{h-1}_j$ must be $\eta \sum_i c_{ij} \boldsymbol z^h_i$ for some positive constant $\eta$. Then the inner product will be
 \begin{align}
     \langle \boldsymbol z^{h-1}_j, \boldsymbol z^{h-1}_j \rangle_\mathbb{L} = \eta^2 \langle \sum_i c_{ij} \boldsymbol z^h_i, \sum_i c_{ij} \boldsymbol z^h_i \rangle_\mathbb{L} = \frac{1}{\kappa}.
 \end{align}
 So $\eta$ will be $\frac{1}{\sqrt{-\kappa}\lvert \lVert \sum_i c_{ij} \boldsymbol z^h_i \rVert_\mathbb{L} \rvert}$, the proof is completed.
\end{proof}

\paragraph{Remark}
In fact, the geometric centroid in Theorem 5.2 is also equivalent to the gyro-midpoint in Poincar\'{e} ball model of hyperbolic space.
\begin{proof}
    Let $\boldsymbol z^h_i = [t^h_i, (\boldsymbol s^h_i)^T]^T$, substitute into Theorem 52, we have
    \begin{align}
        \begin{split}
            \boldsymbol z^{h-1}_j &= \frac{1}{\sqrt{-\kappa}} \sum_{i=1}^n \frac{ c_{ij}}{\lvert \lVert \sum_{l=1}^n c_{lj} \boldsymbol z^{h}_l  \rVert_\mathbb{L} \rvert} \boldsymbol z^{h}_i \\
        &=\frac{1}{\sqrt{-\kappa}}\frac{[\sum_i c_{ij}t^h_i, \sum_i c_{ij}\boldsymbol s^h_i]^T}{\sqrt{(\sum_i c_{ij}t^h_i)^2 - \lVert \sum_i c_{ij}\boldsymbol s^h_i\rVert^2_2}}.
        \label{eq.expand}
        \end{split}
    \end{align}
    Recall the stereographic projection $\mathcal{S}: \mathbb{L}^{\kappa, d_\mathcal{T}} \rightarrow \mathbb{B}^{\kappa, d_\mathcal{T}}$:
    \begin{align}
        \mathcal{S}([t^h_i, (\boldsymbol s^h_i)^T]) = \frac{\boldsymbol s^h_i}{1 + \sqrt{-\kappa}t^h_i}.
    \end{align}
    Then we apply $\mathcal{S}$ to both sides of Eq. (\ref{eq.expand}) to obtain
    \begin{align}
        \begin{split}
            \boldsymbol b^{h-1}_j &= \frac{1}{\sqrt{-\kappa}}\frac{\sum_i c_{ij}\boldsymbol s^h_i}{\sqrt{(\sum_i c_{ij}t^h_i)^2 - \lVert \sum_i c_{ij}\boldsymbol s^h_i\rVert^2_2} + \sum_i c_{ij}t^h_i} \\
        &= \frac{1}{\sqrt{-\kappa}}\frac{\frac{\sum_i c_{ij}\boldsymbol s^h_i}{\sum_i c_{ij}t^h_i}}{1 + \sqrt{1 - \frac{\lVert \sum_i c_{ij}\boldsymbol s^h_i\rVert^2_2}{(\sum_i c_{ij}t^h_i)^2}}} \\
        &= \frac{\bar{\boldsymbol b}^h}{1 + \sqrt{1 + \kappa \lVert \bar{\boldsymbol b}^h \rVert^2_2}},
        \end{split}
    \end{align}
    where $\boldsymbol b^{h-1}_j = \mathcal{S}(\boldsymbol z^{h-1}_j)$, and $\bar{\boldsymbol b}^h = \frac{1}{\sqrt{-\kappa}}\frac{\sum_i c_{ij}\boldsymbol s^h_i}{\sum_i c_{ij}t^h_i}$.
    Let $\boldsymbol b^h_i = \mathcal{S}(\boldsymbol z^h_i)$, then 
    \begin{align}
        \bar{\boldsymbol b}^h = 2\frac{\sum_i c_{ij}
        \frac{\boldsymbol b^h_i}{1 + \kappa \lVert \boldsymbol b^h_i \rVert^2_2}}{\sum_i c_{ij}\frac{1 - \kappa \lVert \boldsymbol b^h_i \rVert^2_2}{1 + \kappa \lVert \boldsymbol b^h_i \rVert^2_2}}
        =\frac{\sum_i c_{ij} \lambda^\kappa_{\boldsymbol b^h_i} \boldsymbol b^h_i }{\sum_i c_{ij}(\lambda^\kappa_{\boldsymbol b^h_i}-1)},
    \end{align}
    which is the gyro-midpoint in Poincar\'{e} ball model. The proof is completed.
\end{proof}

\subsection{On Structural Information Minimization}
\label{proof.decrease}

We show the partitioning tree of minimal structural entropy without the constraint on the tree height.

\vspace{0.05in}
\begin{mymath}
\noindent\textbf{Theorem IV.4} \textbf{(Monotonic Decrease with Height).}
    \emph{As the dimensionality \( H \) of a partitioning tree increases, its structural entropy \( \mathcal{H}^\mathcal T(G; H) \) decreases monotonically, i.e., \( \mathcal{H}^\mathcal T(G; H+1) < \mathcal{H}^\mathcal T(G; H)  \).}
    \end{mymath}
\begin{proof}
    Consider a network \( G \) with volume \( V(G) \) and an initial \( K \)-dimensional partition tree \( C_K \), constructed using an algorithm that minimizes the \( K \)-dimensional structural information. For each layer \( i \) in the partition tree, let \(\alpha_i\) denote the partition of the graph at that level, \( V_{\alpha_i} \) the volume of node \(\alpha_i\), and \( g_{\alpha_i} \) the number of cut edges associated with node \(\alpha_i\). The structural entropy of the \( K \)-dimensional partition tree is given by
\[
\mathcal{H}(C_K) = -\frac{1}{V(G)} \sum_{i=1}^{K} g_{\alpha_i} \log_2 \frac{V_{\alpha_i}}{V_{\alpha_m}},
\]
where \( V_{\alpha_m} \) is the total volume of the graph at the root of the tree.

Next, perform a combination operation on the \( K \)-th layer of the tree to construct a \( K+1 \)-dimensional partition tree \( C_{K+1} \). Specifically, merge the child nodes \(\alpha_1, \alpha_2, \dots, \alpha_t\) into a new node \(\alpha'_1\) and the remaining child nodes \(\alpha_{t+1}, \dots, \alpha_k\) into a new node \(\alpha'_2\). After this operation, the structural entropy of \( C_{K+1} \) is expressed as
\begin{align}
\mathcal{H}(C_{K+1}) &= -\frac{1}{V(G)} \Bigg( \sum_{i=1}^{t} g_{\alpha_i} \log_2 \frac{V_{\alpha_i}}{V_{\alpha'_1}} + \sum_{i=t+1}^{k} g_{\alpha_i} \log_2 \frac{V_{\alpha_i}}{V_{\alpha'_2}} \notag \\
&\quad + g_{\alpha'_1} \log_2 \frac{V_{\alpha'_1}}{V_{\alpha_m}} + g_{\alpha'_2} \log_2 \frac{V_{\alpha'_2}}{V_{\alpha_m}} \Bigg),
\end{align}
where \( g_{\alpha'_1} = \sum_{i=1}^{t} g_{\alpha_i} \) and \( g_{\alpha'_2} = \sum_{i=t+1}^{k} g_{\alpha_i} \).

To compare the structural entropy of the two partition trees, consider the difference \( \Delta \mathcal{H} = \mathcal{H}(C_{K+1}) - \mathcal{H}(C_K) \). Substituting the expressions for \( \mathcal{H}(C_K) \) and \( \mathcal{H}(C_{K+1}) \), we have
\begin{align}
\Delta \mathcal{H} &= -\frac{1}{V(G)} \Bigg( \sum_{i=1}^{t} g_{\alpha_i} \log_2 \frac{V_{\alpha_i}}{V_{\alpha'_1}} + \sum_{i=t+1}^{k} g_{\alpha_i} \log_2 \frac{V_{\alpha_i}}{V_{\alpha'_2}} \notag \\
&\quad + g_{\alpha'_1} \log_2 \frac{V_{\alpha'_1}}{V_{\alpha_m}} + g_{\alpha'_2} \log_2 \frac{V_{\alpha'_2}}{V_{\alpha_m}} \Bigg).
\end{align}

Note that, \( \sum_{i=1}^{t} g_{\alpha_i} + \sum_{i=t+1}^{k} g_{\alpha_i} = g_{\alpha'_1} + g_{\alpha'_2} \), and using the fact that \( \frac{V_{\alpha'_1}}{V_{\alpha_m}} < 1 \) and \( \frac{V_{\alpha'_2}}{V_{\alpha_m}} < 1 \), it follows that the logarithmic terms are negative. Consequently, \( \Delta \mathcal{H} < 0 \), implying that
\[
\mathcal{H}(C_{K+1}) < \mathcal{H}(C_K).
\]

By induction, the structural entropy of the partition tree decreases monotonically as the dimensionality of the tree increases. 
\end{proof}

\begin{algorithm}[t]
    \caption{Training \texttt{ASIL}}
    \label{alg:lsenet}
    \renewcommand{\algorithmicrequire}{\textbf{Input:}}
    \renewcommand{\algorithmicensure}{\textbf{Output:}}
    \begin{algorithmic}[1]
        \REQUIRE A weighted graph $G=(\mathcal{V}, \mathcal{E}, \mathbf{X})$ and the height of  deep partitioning tree $H$, fusion parameter $\gamma \in (0,1]$, nearset neighbors number $k$.
        %, Training iterations $L$
         \ENSURE The partitioning tree $\mathcal{T}$, and tree node embeddings $\{\mathbf{Z}_\gamma^h\}_{h=1,...,H}$ at all the levels.
        %\FOR{$epoch=1$ to $L$}
          \WHILE{not converged}
                    
        \STATE Obtain initial  embeddings $\mathbf{Z}^H = \operatorname{LConv}(\mathbf{X}, \mathbf{A})$;
        \STATE Infer virtual graph  $\tilde{\mathbf{A}}_{ij} = \exp(-d_{\mathbb{L}}(L(\boldsymbol{z}_i), L(\boldsymbol{z}_j))/\tau)$ with  Lorentz boost $L$ in Eq. 32;
          \STATE Sparsify the virtual graph with $k$-nearest neighbors;
        \STATE Conduct graph fusion by $\mathbf{A}^\gamma = \mathbf{A} + \gamma \tilde{\mathbf{A}}$;
        \STATE Obtain leaf node embeddings $\mathbf{Z}_\gamma^H = \operatorname{LConv}(\mathbf{X}, \mathbf{A}^\gamma)$;
        \STATE Compute the level-wise assignment and parent node embeddings $\{\mathbf{Z}_\gamma^h\}_{h=1,...,H-1}$ recursively with  Eqs. 27, 29 and 31;
        \STATE Compute the overall objective $\mathcal{H}^{\mathcal{T}_\text{net}}(G^\gamma;\mathbf{Z};\mathbf{\Theta})$;
        \STATE Update the parameters with gradient-descent optimizer;
        \ENDWHILE
        \STATE Construct the optimal partitioning tree of  root node $\lambda$ with a queue $\mathcal{Q}$;
        \WHILE{$\mathcal{Q}$ is not empty}
            \STATE Get first item $\alpha$ in $\mathcal{Q}$;
            \STATE Let $h=\alpha.h+1$ and search subsets $P$ from $\mathbf{S}^h$ defined with level-wise assignments in Eq. 8;
            \STATE Create nodes from $P$ and put into the queue $\mathcal{Q}$;
            \STATE Add theses nodes into $\alpha$'s children list;
        \ENDWHILE
        \STATE Return the partitioning tree $\mathcal{T}:=\lambda$.
    \end{algorithmic}
\end{algorithm}

\begin{algorithm}[t]
    \caption{\texttt{LSEnet} function}
    \label{alg:lsenet}
    \renewcommand{\algorithmicrequire}{\textbf{Input:}}
    \renewcommand{\algorithmicensure}{\textbf{Output:}}
    \begin{algorithmic}[1]
        \REQUIRE A weighted graph $G=(\mathcal{V}, \mathcal{E}, \mathbf{X})$, Height of partitioning tree $H$
       \ENSURE Tree nodes embeddings $\{\mathbf{Z}^h\}_{h=1,...,H}$ at all the levels;
       % \FOR{$epoch=1$ to $L$}
            \STATE Obtain leaf node embeddings $\mathbf{Z}^H = \operatorname{LConv}(\mathbf{X}, \mathbf{A})$;
            \FOR{$h=H-1$ to $1$}
                \STATE Compute $\mathbf{Z}^h=\operatorname{LAgg}(\mathbf C^{h+1}, \mathbf Z^{h+1})\in L^{\kappa,d_\mathcal{t}}$;
                \STATE Compute $\mathbf{A}^h=(\mathbf C^{h+1})^\top \mathbf A^{h+1}\mathbf C^{h+1}$;
                \STATE Compute the query $\mathbf Q$ and key $\mathbf K$;
                \STATE Compute the following 
                \begin{align}
                \mathbf{C}^h=\sigma((\operatorname{LAtt}(\mathbf Q, \mathbf K)\odot \mathbf A^{h+1})\operatorname{MLP}(Z^{h+1})), \nonumber
                \end{align}
               \STATE Compute $\mathbf{S}^h=\prod^{h+1}_{k=H+1}\mathbf C^k$ for $h=1, \cdots, H$.
            \ENDFOR
        % \STATE Compute the objective of DSI in $\mathcal{H}^{\mathcal{T}}(G; h) = -\frac{1}{V} \sum_{k=1}^{N_h} \left( V^h_k-\sum_{(i,j) \in \varepsilon} S^h_{ik} S^h_{jk} w_{ij} \right) \log_2 \left( \frac{V^h_k}{V^{h-1}_{k^-}} \right).$
        % \STATE Optimize parameters via Adam.
        % \ENDFOR
        % \STATE Create a root node $\lambda$ and a queue $\mathcal{Q}$.
        % \WHILE{$\mathcal{Q}$ is not empty}
        %     \STATE Get first item $\alpha$ in $\mathcal{Q}$.
        %     \STATE Let $h=\alpha.h+1$ and search subsets $P$ from $\mathbf{S}^h$.
        %     \STATE Create nodes from $P$ and put into the queue $\mathcal{Q}$.
        %     \STATE Add theses nodes into $\alpha$'s children list.
        % \ENDWHILE
        % \STATE Return the partitioning tree $\mathcal{T}:=\lambda$.
    \end{algorithmic}
\end{algorithm}

\section{Algorithms}

We give the pseudocode description of algorithms including training \texttt{ASIL}, decoding the partitioning tree and  hyperbolic node clustering with the partitioning tree.

\subsection{Training \texttt{ASIL}}
Algorithm 1 outlines the training process for the \texttt{ASIL} model.
We invoke Algorithm  \ref{alg:lsenet} aims to learn a partitioning tree structure and corresponding node embeddings from a weighted graph. The algorithm begins by obtaining the embeddings of the leaf nodes through a graph convolution operation. Then, starting from the leaf nodes, it iteratively computes embeddings, adjacency matrices, and structural matrices for each level of the tree, working upwards from the leaves to the root. At each level, it also computes the similarity matrix \( \mathbf{S}^h \) for subsequent optimization.
The training process uses the Adam optimizer to minimize the objective function based on the learned embeddings. 
After training, we obtain the refined deep partitioning tree in the hyperbolic space. 
The algorithm builds the partitioning tree using a breadth-first search starting from the root. New child nodes are created based on the similarity matrices and added to the queue for further processing. 
Finally, the refined partitioning tree is returned, representing the graph's hierarchical structure and providing meaningful node embeddings.

\begin{algorithm}[t]
    \caption{Decoding a partitioning tree from hyperbolic embeddings in BFS manner.}
    \label{alg:decode}
    \begin{algorithmic}[1]
        \REQUIRE Hyperbolic embeddings $\mathbf{Z}^h=\{ \mathbf{z}^h_1, \mathbf{z}^h_2, ..., \mathbf{z}^h_N\}$ for $h=1,...,H$; The height of tree $H$; The matrix $\mathbf{S}^h$ computed $\mathbf{S}^h=\prod^{h+1}_{k=H+1}C^k$ for $h=1, \cdots, H$; List of node attributions $node\_attr$, including $node\_set$, $children$, $height$, $coordinates$; Abstract class $\operatorname{Node}$;
        
        \STATE \textbf{Function} ConstructTree($\mathbf{Z}$, $H$, $node\_attr$)
            \STATE $h = 0$
            \STATE First create a root node.
            \STATE $root = \operatorname{Node}(node\_attr[h])$
            \STATE Create a simple queue.
            \STATE $que = \operatorname{Queue}()$
            \STATE $que.\operatorname{put}(root)$
            
            \WHILE{$que$ is not empty}
                \STATE $node = que.\operatorname{get}()$
                \STATE $node\_set = node.node\_set$
                \STATE $k = node.height + 1$
                
                \IF{$h == H$}
                    \FOR{$i$ in $node\_set$}
                        \STATE $child = \operatorname{Node}(node\_attr[h][i])$
                        \STATE $child.coordinates = \mathbf{Z}^h_i$
                        \STATE $node.children.\operatorname{append}(child)$
                    \ENDFOR
                \ELSE
                    \FOR{$k=1$ to $N_h$}
                        \STATE $L\_child$ = [$i$ in where $S^h_{ik}==1$]
                        
                        \IF{$\operatorname{len}(L\_child) > 0$}
                            \STATE $child=\operatorname{Node}(node\_attr[h][L\_child])$
                            \STATE $child.coordinates = \mathbf{Z}^h_{[L\_child]}$
                            \STATE $node.children.\operatorname{append}(child)$
                            \STATE $que.\operatorname{put}(child)$
                        \ENDIF
                    \ENDFOR
                \ENDIF
            \ENDWHILE
            \STATE \textbf{return} $root$
        \ENSURE \textbf{Function} ConstructTree($\mathbf{Z}$, $H$, $node\_attr$)
    \end{algorithmic}
\end{algorithm}

\subsection{Decoding algorithm}
\label{append.alg}
The Algorithm \ref{alg:decode} is the decoding algorithm to recover a partitioning tree from hyperbolic embeddings and level-wise assignment matrices. we perform this process in a Breadth First Search manner. 
As we create a tree node object, we put it into a simple queue. 
Then we search the next level of the tree nodes and create children nodes according to the level-wise assignment matrices and hyperbolic embeddings. Finally, we add these children nodes to the first item in the queue and put them into the queue. 
For convenience, we can add some attributes into the tree node objects, such as node height, node number, and so on.

\begin{algorithm}[t]
    \caption{Obtaining objective cluster numbers from a partitioning tree.}
    \label{alg:cluster}
    \renewcommand{\algorithmicrequire}{\textbf{Input:}}
    \begin{algorithmic}[1]
        \REQUIRE A partitioning tree $\mathcal{T}$; Tree node embeddings $\mathbf{Z}$ in hyperbolic space; Objective cluster numbers $K$.
        \STATE Let $\lambda$ be the root of the tree.
        \FOR{$u$ in $\mathcal{T}.nodes$ / $\lambda$}
            \STATE Compute distance to $\lambda$, i.e., $d_\mathbb{L}(\boldsymbol{z}_\lambda, \boldsymbol{z}_u)$.
        \ENDFOR
        \STATE Sorted non-root tree nodes by distance to $\lambda$ in ascending order, output a sorted list $L$.
        \STATE Let $h=1$ and count the number $M$ of nodes at height $h$.
        \WHILE{$M > K$}
            \STATE Merge two nodes $u$ and $v$ that are farthest away from root $\lambda$.
            \STATE Compute midpoint $p$ of $u$ and $v$, and compute $d_\mathbb{L}(\boldsymbol{z}_\lambda, \boldsymbol{z}_p)$.
            \STATE Add $p$ into $L$ and sort the list again in ascending order.
            \STATE $M=M-1$.
        \ENDWHILE
        \WHILE{$M < K$}
            \FOR{$v$ in $L$}
                \STATE Let $h=h+1$.
                \STATE Search children nodes of $v$ in $h$-level as sub-level list $S$ and count the number of them as $m$.
                \STATE $M=M+ m - 1$.
                \IF{$M > K$}
                    \STATE Perform merge operation to $S$ the same as it to $L$.
                    \STATE Delete $v$ from $L$, and add $S$ to $L$.
                    \STATE Break the for-loop.
                \ELSIF{$M = K$}
                    \STATE Delete $v$ from $L$, and add $S$ to $L$.
                    \STATE Break the for-loop.
                \ELSE
                    \STATE Delete $v$ from $L$, and add $S$ to $L$.
                \ENDIF
            \ENDFOR
        \ENDWHILE
        \STATE Result set $R=\{\}$
        \FOR{$i=0$ to $K-1$}
            \STATE Get graph node subset $Q$ from $L[i]$.
            \STATE Assign each element of $Q$ a clustering category $i$.
            \STATE Add results into $R$.
        \ENDFOR
        
        \textbf{Output:} $R$
    \end{algorithmic}
    
\end{algorithm}

\subsection{Hyperbolic Node Clustering with Partitioning Tree}
In Algorithm \ref{alg:cluster}, we give the approach to obtain node clusters of  a predefined cluster number from a partitioning tree $\mathcal{T}$. The key idea of this algorithm is that we first search the first level of the tree if the number of nodes is more than the predefined numbers, we merge the nodes that are farthest away from the root node, and if the number of nodes is less than the predefined numbers, we search the next level of the tree, split the nodes that closest to the root. As we perform this iteratively, we will finally get the ideal clustering results. The way we merge or split node sets is that: the node far away from the root tends to contain fewer points, and vice versa.

\section{Structural Entropy}

\subsection{Partitioning Tree of Graph}

% The one-dimensional structural entropy is defined as follows,
% \label{hse}
%     \begin{align}
%         \label{eq.one.si}
%             \mathcal{H}^{1}(G)=-\sum_{v\in \mathcal{V}}\frac{d_{v}}{\operatorname{Vol}(G)}\log_2\frac{d_{v}}{\operatorname{Vol}(G)},
%     \end{align}
% yielding as a constant for certain given graph.
The $H$-dimensional structural information is accompanied with a partitioning tree $\mathcal{T}$ of height $H$.
We consider a graph $G=(\mathcal{V},\mathcal{E})$ with weight function $w$,
and the weighted degree of node $v$ is written as $d_v=\sum\nolimits_{u\in \mathcal N(v)}w(v,u)$, the summation of the edge weights surrounding it. 
$\mathcal N(v)$ denotes the neighborhood of $v$.
For an arbitrary node set $\mathcal{U}  \subset \mathcal{V}$, its volume $\operatorname{Vol}(\mathcal{U})$ is defined by summing of weighted degrees of the nodes in $\mathcal{U}$, and $\operatorname{Vol}(G)$ denotes the volume of the graph.

In a partitioning tree $\mathcal{T}$ with root node $\lambda$, 
each tree node $\alpha$ is associated with a subset of $\mathcal V$, referred to as module $T_\alpha \subset \mathcal{V}$.\footnote{The vertex of partitioning tree is termed as ``tree node'' to distinguish from the node in graphs.}
and its immediate predecessor is written as $\alpha^-$. 
The leaf node of the tree corresponds to the node of graph $G$.
Accordingly, the module of the leaf node is a singleton, while $T_\lambda$ is the node set $\mathcal V$.
% is considered as the node set of $G$.
The structural information assigned to each non-root node $\alpha$ is defined as 
\label{hse}
    \begin{align}
        \label{eq.node.si}
            \mathcal{H}^{\mathcal{T}}(G;\alpha)=-\frac{g_\alpha}{\operatorname{Vol}(G)}\log_2\frac{V_\alpha}{V_{\alpha^-}},
    \end{align}
where the scalar $g_\alpha$ is the total weights  of graph edges with exactly one endpoint in the $T_\alpha$, and $V_\alpha$ is the volume of the module $T_\alpha$.
The $H$-dimensional structural information of $G$ with respect to the partitioning tree $\mathcal{T}$ is written as follows,
\label{hse}
    \begin{align}
        \label{eq.tree.si}
            \mathcal{H}^{\mathcal{T}}(G) = \sum\nolimits_{\alpha \in \mathcal{T}, \alpha \neq \lambda }\mathcal{H}^{\mathcal{T}}(G;\alpha).
    \end{align}
Traversing all possible partitioning trees of $G$ with height $H$,  the \emph{$H$-dimensional structural entropy} of $G$ is defined as
    \begin{align}
        \mathcal{H}^{H}(G) = \min_\mathcal{T} \mathcal{H}^{\mathcal{T}}(G), \quad
    \label{eq.opt_tree}
        \mathcal{T}^*=\arg_{\mathcal{T}} \min \mathcal{H}^{\mathcal{T}}(G),
    \end{align}
where  $\mathcal{T}^*$ is the optimal partitioning tree of $G$ that minimizes the uncertainty of graph, 
and encodes the self-organization of node clusters in the tree structure.

\begin{algorithm}[t]
\caption{Coding tree with height $k$ via structural entropy minimization}
\label{code:coding_tree} 
\textbf{Input:} a graph $G=(\mathcal{V}, \mathcal{E})$, a positive integer $k>1$\\
\textbf{Output:} a coding tree $T$ with height $k$

\begin{algorithmic}[1]
\STATE Generate a coding tree $T$ with a root node $v_r$ and all nodes in $\mathcal{V}$ as leaf nodes;
\STATE // Stage 1: Bottom to top construction;
\WHILE{$|v_r.children|>2$} {
  \STATE Select $v_i$ and $v_j$ from $v_r.children$, conditioned on \\
  $argmax_{(v_i, v_j)}\{\mathcal{H}^T(G) - \mathcal{H}^{T_{\text{MERGE}(v_i, v_j)}}(G)\}$;
  \STATE $\text{MERGE}(v_i, v_j)$;
}
\ENDWHILE
\STATE // Stage 2: Compress $T$ to the certain height $k$;
\WHILE{$\text{Height}(T)>k$} {
  \STATE Select $v_i$ from $T$, conditioned on \\
  $argmin_{v_i}\{\mathcal{H}^{T_{\text{REMOVE}(v_i)}}(G) - \mathcal{H}^T(G)|$\\
  \qquad \qquad \,\, $v_i\neq v_r \,\&\, v_i\notin \mathcal{V}\}$;
  \STATE $\text{REMOVE}(v_i)$;
}
\ENDWHILE
\STATE // Stage 3: Fill $T$ to avoid cross-layer links;
\FOR{$v_i\in T$} {
  \IF{$|\text{Height}(v_i.parent)-\text{Height}(v_i)|>1$} 
    \STATE $\text{FILL}(v_i, v_i.parent)$;
  \ENDIF
}
\ENDFOR
\STATE return $T$;
\end{algorithmic} 
\end{algorithm}

\subsection{Heuristic Algorithm of Constructing Partitioning Tree}

Given a tree height, \cite{li2016structural} propose a heuristic greedy algorithm to construct the partitioning tree by structural information minimization.
% Orthogonal to our study, the original minimization of structural entropy guides the construction of the partitioning tree with a given fixed height through a heuristic greedy algorithm, which directs hierarchical clustering pooling. 
In Algo. \ref{code:coding_tree}, the construction procedure is divided into  three stages.
\emph{In the first stage}, the full-height binary coding tree is generated from to the top, two child nodes of the root are merged in each iteration to form a new partition, aiming to maximize the reduction in structural entropy. 
\emph{In the second stage}, to achieve the desired level of graph coarsening, the tree is compressed by removing nodes. 
In each step, an inner node is selected for removal such that the structural entropy of the tree is minimized. 
At the end of this phase, a coding tree of height $k$ is obtained. 
% However, cross-layer links may result in missing nodes in the hierarchical pooling process. 
% To resolve this, 
\emph{The third phase} ensures the completeness of information transmission between layers without disrupting the structural entropy of the coding tree. Finally, the coding tree $T$ for the graph $G$ is obtained, where $T = (\mathcal{V}^T, \mathcal{E}^T)$, with $\mathcal{V}^T = (\mathcal{V}^T_0, \dots, \mathcal{V}^T_k)$ and $\mathcal{V}^T_0 = \mathcal{V}$.

{
\subsection{Comparison to Other Graph Entropys}
Although DSI or SI are still entropies that describe graph complexity, but they are essentially different from classical graph entropy such as Shannon entropy and Von Neumann entropy.

\textbf{Shannon entropy}~\cite{coverelementsa}
  \begin{itemize}
    \item \textit{Definition:} $-\sum_i p_i \log p_i$, where $p_i=\frac{d_i}{\sum_{j=1}^{N}d_j}$, $d_i$ is the degree of node $i$.
    \item \textit{Input:} A probability distribution over nodes or edges (e.g., degree distribution).
    \item \textit{Output:} A scalar measuring statistical uncertainty.
    \item \textit{Properties:} It exhibits the spread of local node attributes.
    \item \textit{Relation to clustering:} cannot be served as  clustering objective.
  \end{itemize}

\textbf{Von Neumann entropy}~\cite{10.4018/jats.2009071005}
  \begin{itemize}
    \item \textit{Definition:} $-\sum_i \lambda_i \log \lambda_i$, where $\{\lambda_i\}$ are eigenvalues of the normalized Laplacian.
    \item \textit{Input:} Eigenvalues of the normalized Laplacian.
    \item \textit{Output:} A scalar reflecting global smoothness of graph signal.
    \item \textit{Properties:} It shows the structure of global connectivity (e.g. the size of bottlenecks).
    \item \textit{Relation to clustering:} Spectral gaps may suggest cluster boundary.
  \end{itemize}

\textbf{Structural entropy}~\cite{li2016structural}
  \begin{itemize}
    \item \textit{Definition:} $-\sum\nolimits_{\alpha \in \mathcal{T}, \alpha \neq \lambda }\frac{g_\alpha}{\operatorname{Vol}(G)}\log_2\frac{V_\alpha}{V_{\alpha^-}}$.
    \item \textit{Input:} The graph $G = (V, E)$ and a partitioning tree $\mathcal{T}$.
    \item \textit{Output:} A scalar describe the self-organization of graph, and an optimal partitioning tree.
    \item \textit{Properties:} It measures the partitioning uncertainty, shows the structure of local and global connectivity. The dimension of structural entropy reflects the height of clustering hierarchy.
    \item \textit{Relation to clustering:} It naturally discovers the community structure in graph. Clustering result can be decoded from the optimal partitioning tree $\mathcal{T}^*$ that minimizes structural information. 
  \end{itemize}
}

\section{Riemannian Geometry}

In this section, we first introduce the basic notations in Riemannian geometry (e.g., manifold, curvature, geodesic, midpoint, exponential and  logarithmic map),  then elaborate on two popular model of hyperbolic space, Lorentz model and Poincar\'{e} ball model.
Furthermore, we specify the alignment between tree and hyperbolic space.

\subsection{Important Concepts}

\subsubsection*{Riemannian Manifold}
Riemannian manifold is a real and smooth manifold $\mathcal{M}$ equipped with Riemannian metric tensor $g_{\boldsymbol x}$ on the tangent space $\mathcal{T}_{\boldsymbol{x}}\mathcal{M}$ at point $\boldsymbol{x}$.
A Riemannian metric assigns to each $\boldsymbol{x} \in \mathcal{M}$ a positive-definite inner product $g_{\boldsymbol{x}}: \mathcal{T}_{\boldsymbol{x}}\mathcal{M} \times \mathcal{T}_{\boldsymbol{x}}\mathcal{M} \rightarrow \mathbb{R}$, which induces a norm defined as $\lVert \cdot \rVert: \boldsymbol{v} \in \mathcal{T}_{\boldsymbol{x}}\mathcal{M} \mapsto \sqrt{g_{\boldsymbol{x}}(\boldsymbol{v}, \boldsymbol{v})} \in \mathbb{R}$.

\subsubsection*{Curvature}
In Riemannian geometry, curvature quantifies how a smooth manifold bends from flatness. A manifold \( \mathcal{M} \) is called constant curvature space (CCS) if its scalar curvature is consistent at every point. Here are three common spaces/models for different constant curvature spaces: the hyperbolic space with $\kappa < 0$, the spherical model with $\kappa > 0$, and the Euclidean space with $\kappa=0$.

\subsubsection*{Geodesic}
A manifold is geodesically complete if the maximal domain for any geodesic is \( \mathbb{R} \), and a geodesic is the shortest curve between two points $\boldsymbol{x,y}$ in manifold. In Riemannian manifold \( (\mathcal{M}, g) \), the geodesic induces a distance
\begin{align}
d(\boldsymbol{x,y}) = \inf\{L(\gamma) \mid & \gamma \text{ is a piecewise smooth curve}\},
\end{align}
the completeness of \( d(\cdot, \cdot) \) implies that every Cauchy sequence converges within the space.

\subsubsection*{Exponential Map}
An exponential map at point $\boldsymbol x \in \mathcal{M}$ is denoted as $\operatorname{Exp}_{\boldsymbol x}(\cdot): \boldsymbol u \in \mathcal{T}_{\boldsymbol{x}}\mathcal{M} \mapsto \operatorname{Exp}_{\boldsymbol x}(\boldsymbol u) \in \mathcal{M}$. It transforms $\boldsymbol x$ along the geodesic starting at $\boldsymbol x$ in the direction of $\boldsymbol u$.

\subsubsection*{Logarithmic Map}
The logarithmic map is the inverse of the exponential map. Given two points \(\boldsymbol x, y \in \mathcal{M}\), the logarithmic map \(\text{Log}_x(y)\) finds the direction and length of the geodesic from \(x\) to \(y\) and represents it as a vector in the tangent space at \(x\).

\subsection{Midpoint in the Manifold}
\label{hyperbolic.midpoint}

Let $(\mathbb{M}, g)$ be a Riemannian manifold, and $\boldsymbol{x}_1, \boldsymbol{x}_2, ..., \boldsymbol{x}_n$ are points on the manifold. The Fr\'{e}chet variance at point $\boldsymbol{\mu}\in \mathbb{M}$ of these points are given by
\begin{align}
    \Psi(\boldsymbol{\mu}) = \sum_i d^2(\boldsymbol{\mu}, \boldsymbol{x}_i).
\end{align}
%If for some $\boldsymbol{\mu}$s locally minimize the variance, they are called Karcher means. 
If there is a point $\boldsymbol{p}$ locally minimizes the variance, it is called Fr\'{e}chet mean. Generally, if we assign each $\boldsymbol{x}_i$ a weight $w_i$, the Fr\'{e}chet mean can be formulated as
\begin{align}
    \boldsymbol{p} = \mathop{\arg\min}_{\boldsymbol{\mu}\in \mathbb{M}} \sum_i w_i d^2(\boldsymbol{\mu}, \boldsymbol{x}_i).
\end{align}
Note that, $d$ is the canonical distance in the manifold. 

\subsection{Lorentz Model}

The $d$-dimensional Lorentz model equipped with constant curvature $\kappa < 0$ and Riemannian metric tensor $\mathbf{R} =diag(-1,1,...,1)$, is defined in $(d+1)$-dimensional Minkowski space whose origin is $(\sqrt{-1/\kappa}, 0,...,0)$. Formally, the model is expressed as:
\begin{align}
    \mathbb{L}^d_\kappa=\{\boldsymbol{x}\in \mathbb{R}^{d+1} \lvert \langle \boldsymbol{x}, \boldsymbol{x} \rangle_\mathbb{L}=\frac{1}{\kappa}\},
\end{align}
where the Lorentz inner product is defined as:
\begin{align}
    \langle \boldsymbol{x}, \boldsymbol{y} \rangle_\mathbb{L}=-x_0y_0+\sum_{i=1}^dx_iy_i=\boldsymbol{x}^T\mathbf{R}\boldsymbol{y}
\end{align}
The distance between two points is given by
\begin{align}
    d_\mathbb{L}(\boldsymbol{x}, \boldsymbol{y}) = \sqrt{-\kappa} \operatorname{cosh}^{-1}(-\langle \boldsymbol{x}, \boldsymbol{y} \rangle_\mathbb{L}).
\end{align}
The tangent space at $\boldsymbol{x} \in \mathbb{L}^d_\kappa$ is the set of $\boldsymbol{y} \in \mathbb{L}^d_\kappa$ such that orthogonal to $\boldsymbol{x}$ w.r.t. Lorentzian inner product, denotes as:
\begin{align}
\mathcal{T}_{\boldsymbol{x}} \mathbb{L}^d_\kappa=\{ \boldsymbol{y} \in \mathbb{R}^{d+1} \lvert \langle \boldsymbol{y}, \boldsymbol{x}\rangle_\mathbb{L}=0 \}.
\end{align}
The exponential and logarithmic maps at $\boldsymbol x$ are defined as 
\begin{align}
    \operatorname{Exp}_{\boldsymbol x}^\kappa(\boldsymbol u) &= \cosh(\sqrt{-\kappa}\lVert \boldsymbol u\rVert_\mathbb{L})\boldsymbol x + \sinh(\sqrt{-\kappa}\lVert \boldsymbol u\rVert_\mathbb{L})\frac{\boldsymbol u}{\sqrt{-\kappa}\lVert \boldsymbol u\rVert_\mathbb{L}}   \label{expmap} \\
    \operatorname{Log}_{\boldsymbol{x}}^\kappa(\boldsymbol{y})&=\frac{\cosh^{-1}(\kappa\langle \boldsymbol x, \boldsymbol y
    \rangle_\mathbb{L})}{\sqrt{(\kappa\langle \boldsymbol x, \boldsymbol y
    \rangle_\mathbb{L})^2-1}}(\boldsymbol y - \kappa\langle \boldsymbol x, \boldsymbol y
    \rangle_\mathbb{L}\boldsymbol x).
\end{align}

\subsection{Poincar\'{e} Ball Model}

The $d$-dimensional Poincar\'{e} ball model with constant negative curvature $\kappa$, is defined within a $d$-dimensional hypersphere, formally denoted as $\mathbb{B}^d_\kappa = \{\boldsymbol{x} \in \mathbb{R}^d \lvert \lVert \boldsymbol{x} \rVert^2=-\frac{1}{\kappa}\}$. The Riemannian metric tensor at $\boldsymbol{x}$ is 
$g_{\boldsymbol x}^\kappa=(\lambda_{\boldsymbol{x}}^\kappa)^2g_\mathbb{E}=\frac{4}{(1+\kappa\lVert \boldsymbol{x} \rVert^2)^2}g_\mathbb{E}$, 
where $g_\mathbb{E}=\mathbf{I}$ is the Euclidean metric. The distance function is given by
\begin{align}
     d_\mathbb{B}^\kappa(\boldsymbol{x}, \boldsymbol{y})=\frac{2}{\sqrt{-\kappa}}\tanh^{-1}(\lVert (-\boldsymbol{x}) \oplus_\kappa \boldsymbol{y} \rVert),
\end{align}
where $\oplus_\kappa$ is the M{\"o}bius addition
\begin{align}
    \boldsymbol{x} \oplus_\kappa \boldsymbol{y} = 
    \frac{(1-2\kappa\boldsymbol{x}^\top\boldsymbol{y}-\kappa\lVert \boldsymbol{y} \rVert^2)\boldsymbol{x}+(1+\kappa\lVert \boldsymbol{x} \rVert^2)\boldsymbol{y}}{1-2\kappa\boldsymbol{x}^\top\boldsymbol{y}+\kappa^2\lVert \boldsymbol{x} \rVert^2\lVert \boldsymbol{y} \rVert^2}.
\end{align}
The exponential and logarithmic maps of Pincar\'{e} ball model are defined as 
\begin{align}
    \operatorname{Exp}_{\boldsymbol{x}}^\kappa(\boldsymbol{v})&=\boldsymbol{x} \oplus_\kappa(\frac{1}{\sqrt{-\kappa}}\tanh(\sqrt{-\kappa}\frac{\lambda^\kappa_{\boldsymbol{x}}\lVert \boldsymbol{v} \rVert}{2})\frac{\boldsymbol{v}}{\lVert \boldsymbol{v} \rVert}) \\
    \operatorname{Log}_{\boldsymbol{x}}^\kappa(\boldsymbol{y})&=\frac{2}{\sqrt{-\kappa}\lambda^\kappa_{\boldsymbol{x}}}
    \tanh^{-1}(\sqrt{-\kappa}\lVert (-\boldsymbol{x}) \oplus_\kappa \boldsymbol{y} \rVert)\frac{(-\boldsymbol{x}) \oplus_\kappa \boldsymbol{y}}{\lVert (-\boldsymbol{x}) \oplus_\kappa \boldsymbol{y} \rVert}.
\end{align}

\subsection{Tree and Hyperbolic Space}
\label{treehyp}
We denote the embedding of node $i$ of graph $G$ in $\mathbb{H}$ is $\phi_i$. The following definitions and theorem are from \cite{sarkar2012low}.
\begin{definition}[Delaunay Graph]
    Given a set of vertices in $\mathbb{H}$ their Delaunay graph is one where a pair of vertices are neighbors if their Voronoi cells intersect.
\end{definition}

\begin{definition}[Delaunay Embedding of Graphs]
    Given a graph $G$, its Delaunay embedding in $\mathbb{H}$ is an embedding of the vertices such that their Delaunay graph is $G$.
\end{definition}

\begin{definition}[$\beta$ separated cones]
    Suppose cones $C(\overset{\longrightarrow}{\phi_j\phi_i}, \alpha)$ and $C(\overset{\longrightarrow}{\phi_j\phi_x}, \gamma)$ are adjacent with the same root $\varphi j$. Then the cones are $\beta$ separated if the two cones are an angle $2\beta$ apart. That is, for arbitrary points $p \in C(\overset{\longrightarrow}{\phi_j\phi_i}, \alpha)$ and $q \in C (\overset{\longrightarrow}{\phi_j\phi_x}, \gamma)$, the angle $\angle p\phi_jq > 2\beta$.
\end{definition}

\begin{lemma}
    If cones $C(\overset{\longrightarrow}{\phi_j\phi_i}, \alpha)$ and $C (\overset{\longrightarrow}{\phi_j\phi_x}, \gamma)$ are $\beta$ separated and $\phi_r \in C(\overset{\longrightarrow}{\phi_j\phi_i}, \alpha)$ and $\phi_s \in C (\overset{\longrightarrow}{\phi_j\phi_x}, \gamma)$ then there is a constant $\nu$ depending only on $\beta$ such that $|\phi_r\phi_j |_{\mathbb{H}} + |\phi_s\phi_j|_{\mathbb{H}} > |\phi_r\phi_s|_\mathbb{H} > |\phi_r\phi_j |_\mathbb{H} + |\phi_s\phi_j |_{\mathbb{H}} - \nu$.
\end{lemma}

\begin{theorem}[\cite{sarkar2012low}]
    If all edges of $\mathcal{T}$ are scaled by a constant factor $\tau \geq \eta_{\max}$ such that each edge is longer than $\nu\frac{(1+\epsilon)}{\epsilon}$ and the Delaunay embedding of $\mathcal{T}$ is $\beta$ separated, then the distortion over all vertex pairs is bounded by $1+\epsilon$.
\end{theorem}

Following the above Theorem, we know that a tree can be embedded into hyperbolic space with an arbitrarily low distortion.

\section{Empirical Details}

In this section, we provide a  detailed description of datasets and baselines. The implementation note summarizes the model configuration and further information on reproducibility. Codes are available at \url{https://github.com/RiemannGraph/DSE_clustering}

\subsection{Dataset Description}
\label{append.database}
We evaluate our model on a variety of datasets, i.e., KarateClub, FootBall, Cora, Citeseer, Amazon-Photo (AMAP), and a larger Computers and Pubmed. All of them are publicly available. We give the statistics of the datasets in Table 1 as follows.

\begin{table}[h]
  \centering
  \caption{The statistics of the datasets.}
  \begin{tabular}{|c|c|c|c|c|}
    \hline
    Datasets & \# Nodes & \# Features & \# Edges & \# Classes \\
    \hline
    KarateClub & 34 & 34 & 156 & 4 \\
    FootBall & 115 & 115 & 1226 & 12 \\
    Cora & 2708 & 1433 & 5278 & 7 \\
    Citeseer & 3327 & 3703 & 4552 & 6 \\
    Pubmed & 19717 & 500 & 44324 & 3 \\
    AMAP & 7650 & 745 & 119081 & 8 \\
    AMAC & 13752 & 767 & 245861 & 10\\
    \hline
  \end{tabular}
\end{table}

\begin{figure}[t]
\centering
% 第一行，三个子图
\subfigure[Cora]{
\includegraphics[width=0.3\linewidth]{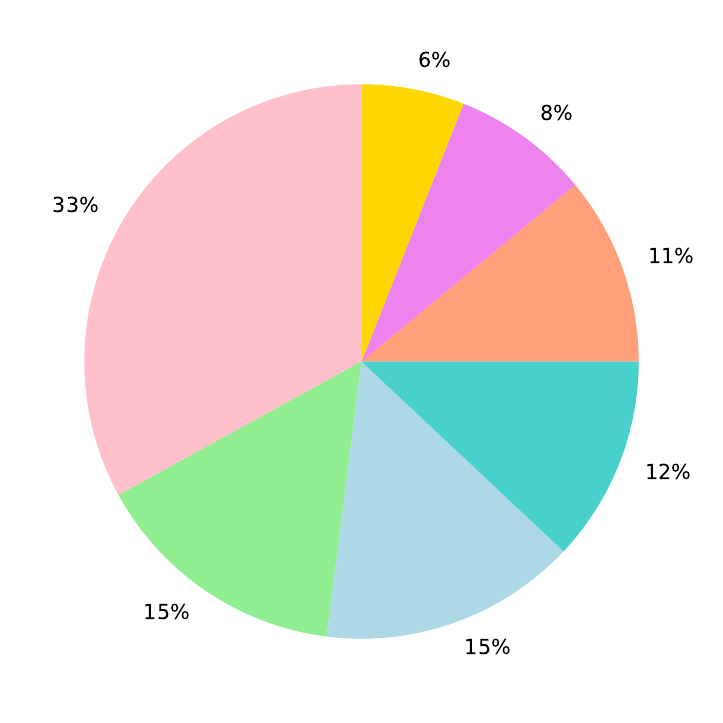}}
\hfill
\subfigure[Photo]{
\includegraphics[width=0.3\linewidth]{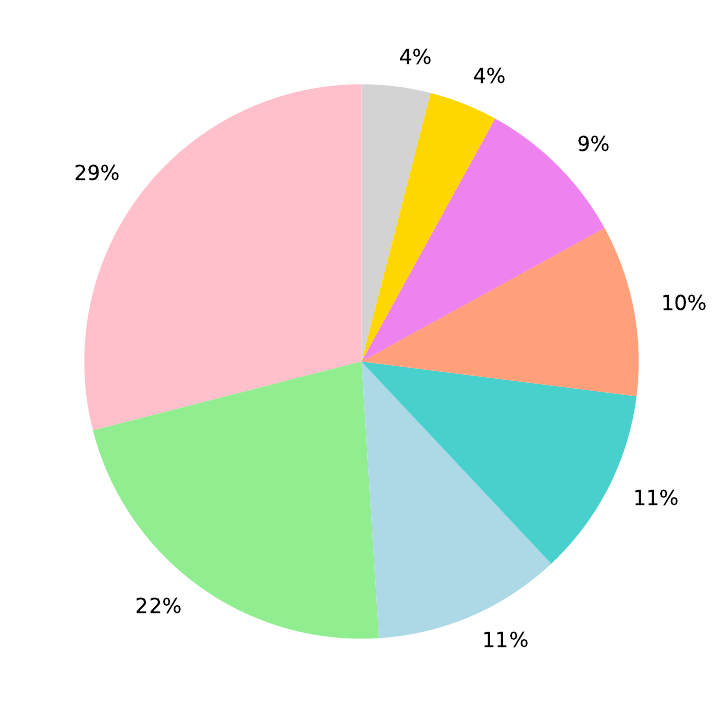}}
\hfill
\subfigure[Pubmed]{
\includegraphics[width=0.3\linewidth]{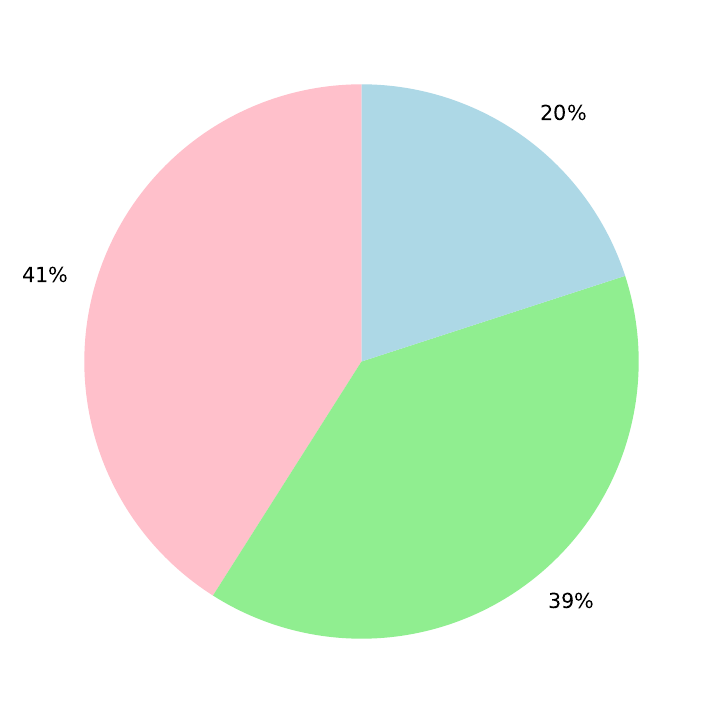}}

\vspace{0.2in} % 两行之间增加一些垂直空间

% 第二行，两个子图
\subfigure[Citeseer]{
\includegraphics[width=0.45\linewidth]{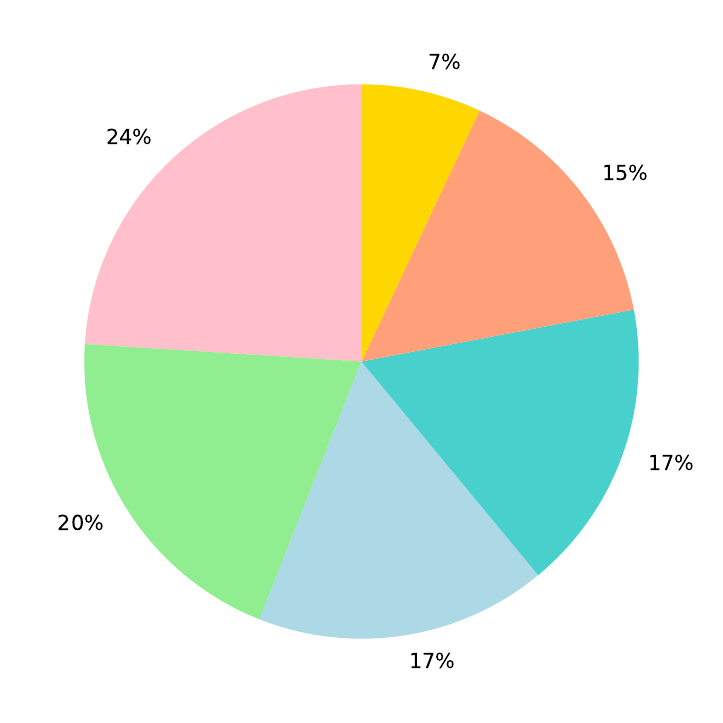}}
\hfill
\subfigure[Computer]{
\includegraphics[width=0.45\linewidth]{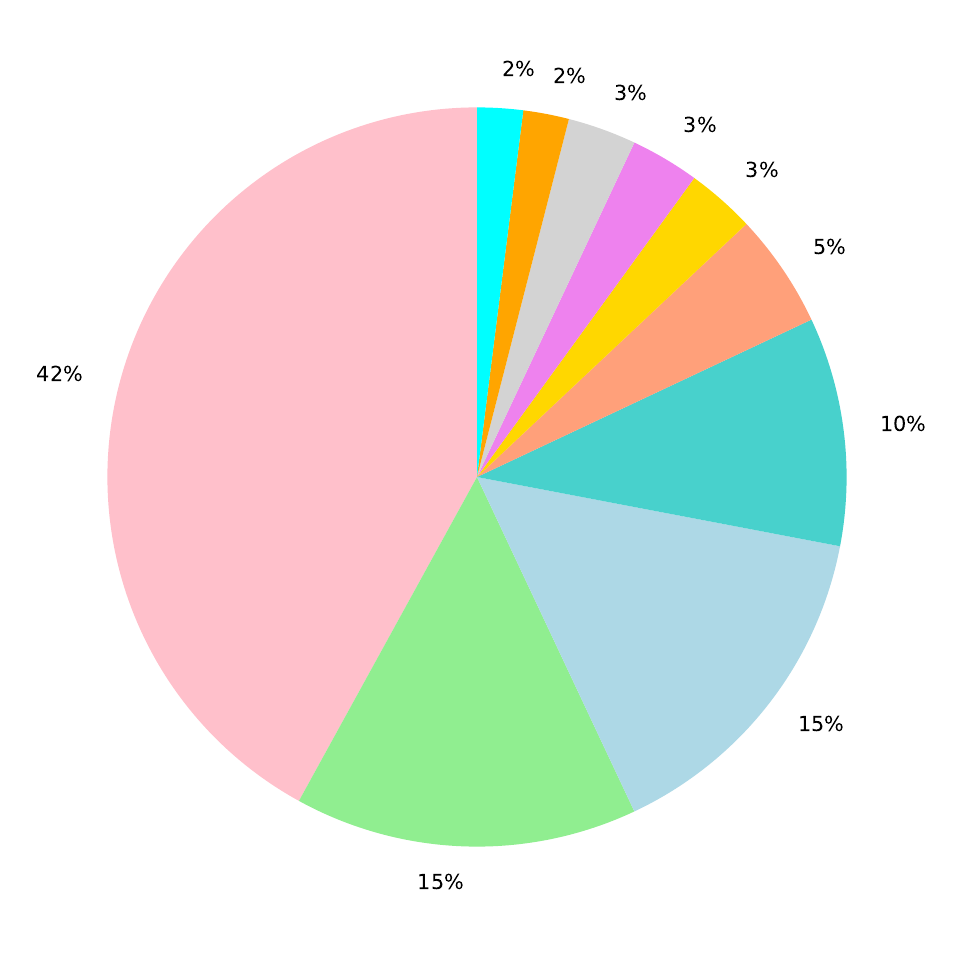}}

\caption{Comparison of clustering performance on different datasets}
\label{fig:datasets_comparison}
\end{figure}

{
\subsection{Baselines \& Configurations}
In this paper, we compare with deep graph clustering methods and self-supervised GNNs introduced as follows,
\begin{itemize} 
  \item[\textbullet] \textbf{RGC} \cite{liu2023reinforcement} enables the deep graph clustering algorithms to work without the guidance of the predefined cluster number. We set actor-critic with 2-layer GCN. Search range $K \in [2, 20]$. Epochs = 300. Code: \url{https://github.com/Liu-Chengran/RGC}

  \item[\textbullet] \textbf{DinkNet} \cite{liu2023dink} optimizes the clustering distribution via the designed dilation and shrink loss functions in an adversarial manner. We set mini-batch strategy (batch size = 1024), auxiliary dimension $P = 512$. $K$ provided during training. Code: \url{https://github.com/keliangliang/DinkNet}

  \item[\textbullet] \textbf{Congregate} \cite{sun2023contrastive} approaches geometric graph clustering with the re-weighting contrastive loss in the proposed heterogeneous curvature space. We set curvature $\kappa$ searched over \{-1, 0, 1\}. Embedding dim = 512 Clustering: $K$-means. Code: \url{https://github.com/RiemannGraph/CONGREGATE}

  \item[\textbullet] \textbf{GC-Flow} \cite{wang2023gc} models the representation space by using Gaussian mixture, leading to an anticipated high quality of node clustering. We set 4 coupling layers, 512-dim hidden states. lr = $5\times10^{-4}$; epochs = 200. $K$ known during inference. Code: \url{https://github.com/Tingting-Wang/GC-Flow}

  \item[\textbullet] \textbf{S$^3$GC} \cite{devvrit2022s3gc} introduces a salable contrastive loss in which the nodes are clustered by the idea of walktrap, i.e., random walk trends to trap in a cluster. We set smoothing $\alpha = 0.2$, propagation steps $T = 5$. MLP projector (2 layers, ReLU). Clustering: $K$-means with ground-truth $K$. Code: \url{https://github.com/flyingdoog/S3GC}

  \item[\textbullet] \textbf{MCGC} \cite{pan2021multi} learns a new consensus graph by exploring the holistic information among attributes and graphs rather than the initial graph. We set each view processed by 2-layer GCN. Final clustering via $K$-means with true $K$. Code: \url{https://github.com/panzhihua2020/MCGC}

  \item[\textbullet] \textbf{DCRN} \cite{liu2021deep} designs the dual correlation reduction strategy to alleviate the representation collapse problem. We set embedding dim = 512; hyperparameters $\alpha = 0.1$, $\beta = 0.01$; epochs = 100. Note: OOM on \textit{Computer} dataset (24GB GPU). Code: \url{https://github.com/Liu-Dongxu/DCRN}

  \item[\textbullet] \textbf{Sublime} \cite{liu2022towards} guides structure optimization by maximizing the agreement between the learned structure and a self-enhanced learning target with contrastive learning. We set encoder with 2-layer GCN (512 hidden dimension). Default augmentations (edge dropping = 0.2). Clustering by $K$-means with ground-truth $K$. Code: \url{https://github.com/Liu-Chengran/Sublime}

  \item[\textbullet] \textbf{gCooL} \cite{DBLP:conf/www/LiJT22} clusters nodes with a refined modularity and jointly trains the cluster centroids with a bi-level contrastive loss. We set encoder with 2-layer GAT (8 heads, 512-dim embedding). Temperature = 0.2. Clustering: $K$-means with known $K$. Code: \url{https://github.com/baoyujing/gCooL}

  \item[\textbullet] \textbf{FT-VGAE} \cite{DBLP:conf/ijcai/MrabahBK22} makes effort to eliminate the feature twist issue in the autoencoder clustering architecture by a solid three-step method.

  \item[\textbullet] \textbf{ProGCL} \cite{xia2021progcl} devises two schemes (ProGCL-weight and ProGCLmix) for further improvement of negatives-based GCL methods.

  \item[\textbullet] \textbf{MVGRL} \cite{HassaniA20} contrasts across the multiple views, augmented from the original graph to learn discriminative encodings. We set encoder-decoder: 2-layer GCN (512 hidden dimension). Clustering: $K$-means. lr = $1\times10^{-3}$; epochs = 200. Code: \url{https://github.com/NajouaE/FT-VGAE}
  \item[\textbullet] \textbf{VGAE} \cite{kipf2016variational} makes use of latent variables and is capable of learning interpretable latent representations for undirected graphs. We set encoder with 2-layer GCN (32-dim hidden), 512-dim latent space. Clustering via $K$-means with true $K$. Optimizer: Adam; lr = $1\times10^{-3}$; epochs = 200. Code: \url{https://github.com/tkipf/gae}
  \item[\textbullet] \textbf{ARGA} \cite{pan2018adversarially} regulates the graph autoencoder with a novel adversarial mechanism to learn informative node embeddings. We set the same encoder as VGAE~\cite{kipf2016variational}. $K$-means used for clustering. Optimizer: Adam; lr = $1\times10^{-3}$; epochs = 200. Code: \url{https://github.com/Ruiqi-Huang/ARGA}

  \item[\textbullet] \textbf{GRACE} \cite{yang2017graph} has multiples nonlinear layers of deep denoise autoencoder with an embedding loss, and is devised to learn the intrinsic distributed representation of sparse noisy contents. We set encoder: 2-layer GCN; projection: 512-dim. Temperature $\tau = 0.2$. Clustering: $K$-means with known $K$. Optimizer: Adam; lr = $1\times10^{-3}$. Code: \url{https://github.com/CRIPAC-DIG/GRACE}

  \item[\textbullet] \textbf{DEC} \cite{xie2016unsupervised} uses stochastic gradient descent (SGD) via backpropagation on a clustering objective to learn the mapping. We followed the original protocol: pre-train a stacked autoencoder with the same architecture as the encoder, then fine-tune the whole network with the clustering loss. The number of clusters $K$ was set to the ground-truth value. Optimizer: Adam; learning rate: $1\times10^{-3}$; batch size: 256. Code: \url{https://github.com/piiswrong/dec}

  \item[\textbullet] \textbf{GCSEE} \cite{ding2023graph} focuses on mining different types of structure information simultaneously to enhance the structure embedding as a self-supervised method. We set GCN backbone (3 layers, 512 hidden). True $K$ used in clustering module. Code: \url{https://github.com/DingSheng/GCSEE}

  \item[\textbullet] \textbf{FastDGC} \cite{ding2024towards} introduces the dynamic graph weight updating strategy to optimize the adjacency matrix. We set efficient graph autoencoder with 3 propagation steps and 2-layer encoder. Clustering via $K$-means with ground-truth $K$. Code: \url{https://github.com/DingSheng/FastDGC}

  \item[\textbullet] \textbf{MAGI} \cite{liu2024revisiting} leverages modularity maximization as a contrastive pretext task to effectively uncover the underlying information of communities in graphs, and avoid the problem of semantic drift. We set edge dropping = 0.3, modularity temperature = 0.5, embedding dim is 512. Clustering: $K$-means. Code: \url{https://github.com/Liu-Chengran/MAGI}

  \item[\textbullet] \textbf{DGCluster} \cite{bhowmick2024dgcluster} uses pairwise (soft) memberships between nodes to solve the graph clustering problem via modularity maximization. We set encoder: 2-layer GNN (512 hidden dimension). No post-hoc clustering needed. Code: \url{https://github.com/abhowmick97/DGCluster}
\end{itemize}
}

\subsection{Implementation Notes}
\label{append.note}

\texttt{ASIL} is implemented upon PyTorch 2.0\footnote{https://pytorch.org/}, Geoopt\footnote{https://geoopt.readthedocs.io/en/latest/index.html}, PyG\footnote{https://pytorch-geometric.readthedocs.io/en/latest/} and NetworkX\footnote{https://networkx.org/}.
%to implement our proposed model.
%\paragraph{hyperparameter settings}
The dimension of structural information is a hyperparameter, which is equal to the height of partitioning tree.
The hyperbolic partitioning tree is learned in a $2-$dimensional hyperbolic space of Lorentz model by default.
The hyperparameter $N_1$ of network architecture is set as $10$. The augmentation ratio $\gamma$ is $0.01$. 
In \texttt{LSEnet}, we stack the Lorentz convolution once.
In Lorentz tree contrastive learning, the number of nearset neighbors in Lorentz Boost Construction is $8$.
For the training phase, we use Adam  and set the learning rate to $0.003$.

\subsection{Isometry Augmentation}
Given a graph with an adjacency matrix $\mathbf{A}\in \mathbb{R}^{N\times N}$ and a similarity matrix $\mathbf{S}\in \mathbb{R}^{N\times N}$ induced by the attributes $\mathbf {X}\in \mathbb{R}^{N\times d}$, 
the  adjacency matrix of augmented graph  $\hat{\mathbf{A}}\in \mathbb{R}^{N\times N}$ considering the isometry in the  $L$ convolution is the \textbf{closed-form} minimizer of the following optimization 
\begin{align}
\min \  \| f(\mathbf A, \mathbf S, \mathbf X;L)\mathbf G^\top_\theta-\hat{\mathbf{A}}^L\mathbf X\|^2,
\end{align}
subject to the symmetry matrix cone of  $\hat{\mathbf{A}}$, 
where $\mathbf{G}_\theta \in \mathbb{R}^{d\times d}$  denotes the Givens rotation, and  
$f(\mathbf{A}, \mathbf{S}, \mathbf {X};L) \in \mathbb{R}^{N\times d}$ is the aggregator of structural and attribute representations.

The construction above enjoys the existence of closed-form solution.
In particular, the augmented graph is expressed as follows,
\begin{align}
\hat{\mathbf A}=\mathbf P\operatorname{diag}(\lambda_1^{\frac{1}{L}}, \cdots, \lambda_d^{\frac{1}{L}})\mathbf P^\top,
    \label{eq.augmented}
\end{align} 
where $\mathbf P$ and $\lambda$'s are derived as the eigenvectors and eigenvalues of the following matrix
    \begin{align}
                \resizebox{0.89\hsize}{!}{$
    \label{eq. AL}
        \hat{\mathbf{A}}^L=\frac{1}{2}[\mathbf{U}^{-1}(\mathbf{X}\mathbf{B}^T+\mathbf{B}\mathbf{X}^T) + (\mathbf{X}\mathbf{B}^T+\mathbf{B}\mathbf{X}^T)\mathbf{U}^{-1}].
        $}
    \end{align}
We have $\mathbf{B}=f(\mathbf{A}, \mathbf{S}, \mathbf{X};L)\mathbf{G}_\theta^T$, and $\mathbf{U}=\mathbf{X}\mathbf{X}^T + \omega\mathbf{I}$ with penalty coefficient $\omega$.

\begin{table}[t]
  \centering
  \caption{Additional results of different representation space of partitioning tree of \texttt{ASIL}.}
    \vspace{-0.05in}
  \label{tab-space}
    \resizebox{\linewidth}{!}{
     \begin{tabular}{ c |ccc|ccc}
      \hline
         & \multicolumn{3}{c|}{\textbf{AMAP}}  & \multicolumn{3}{c}{\textbf{Computer}} \\
        &ACC & NMI  & ARI & ACC & NMI  & ARI \\
      \hline
     \textbf{$\mathbb{H}^{-2}$}
       & \textbf{79.96}  & 74.25  & \textbf{70.01}  & 46.23
       & 53.52 & 43.25 \\
    \hline
     \textbf{$\mathbb{H}^{-1}$}
       & 78.68  & \textbf{74.54}  & 69.24  & \textbf{47.34}
       & \textbf{55.93} & \textbf{43.67} \\
    \hline
    \textbf{$\mathbb{E}^0$}
       & 65.20  & 60.14  & 60.51  & 43.51
       & 39.88 & 38.52 \\
    \hline
     \textbf{$\mathbb{S}^1$}
       & 62.41  & 62.38  & 56.52  & 40.19
       & 36.71 & 35.25 \\
    \hline
     \textbf{$\mathbb{S}^2$}
       & 62.04  & 60.21  & 54.68  & 40.54
       & 37.77 & 34.62 \\
    \hline
    \end{tabular}
}
  \vspace{0.5in}
\end{table}

\begin{figure}[t]
\centering 
\includegraphics[width=1.05\linewidth]{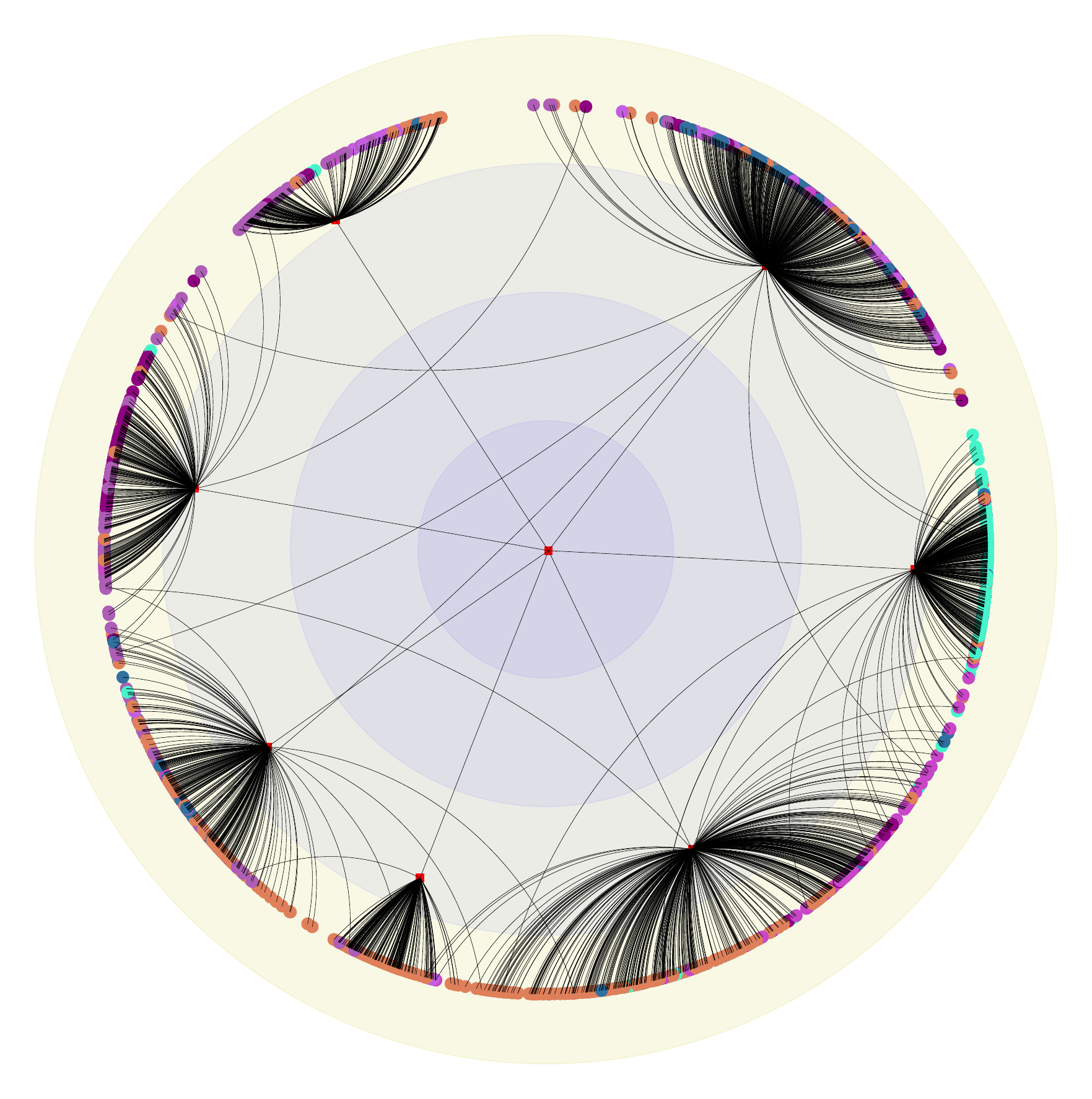}
\caption{Visualization of hyperbolic partitioning trees of Cora.}
\label{Cora.tree}
\end{figure}

\section{Additional Results}
In this section, we provide supplementary experimental results and details that were not fully presented in the main experimental section due to space limitations.

\subsection{Geometric Ablation}
To further validate the impact of different underlying manifolds on the partition tree embedding, and to maintain generality, we further classify Hyperbolic and Hyperspherical spaces based on their curvature. Additionally, we include the AMAP and Computer datasets for further verification, as shown in Table III. Here, $\mathbb{H}$ represents Hyperbolic space, $\mathbb{E}$ represents Euclidean space, and $\mathbb{S}$ represents Hyperspherical space, with the curvature values indicated in the upper-right corner.

\subsection{Visualization}
In addition to the Karate and Football datasets mentioned in the main text, we also visualize the ASIL hyperbolic partition tree on the classic graph clustering Cora dataset, as shown in Fig. 2. The 2D visualization and projection methods are consistent with those used in the main text, with different ground truth label clusters represented in distinct colors.

\vspace{1in}

% \bibliography{icml2024}
% \bibliographystyle{IEEEtran}

\end{document}